\documentclass{article} % For LaTeX2e

\usepackage[final]{colm2026_conference}

\usepackage{amsmath,amsfonts,bm}

\def\eqref#1{equation~\ref{#1}}
\def\1{\bm{1}}

\DeclareMathAlphabet{\mathsfit}{\encodingdefault}{\sfdefault}{m}{sl}
\SetMathAlphabet{\mathsfit}{bold}{\encodingdefault}{\sfdefault}{bx}{n}

\usepackage{amssymb}
\usepackage{algorithm}
\usepackage{algorithmic}
\usepackage{booktabs}
\usepackage{colortbl}
\usepackage{fancyvrb}
\usepackage{float}
\usepackage{graphicx}
\usepackage{lineno}
\usepackage{listings}
\usepackage{makecell}
\usepackage{microtype}
\usepackage{multirow}
\usepackage{multicol}
\usepackage{pgfmath}
\usepackage{pifont}
\usepackage{tabularx}
\usepackage{tcolorbox}
\usepackage{url}
\usepackage{wrapfig}
\usepackage{xcolor}
\usepackage{xspace}

\usepackage[toc,page]{appendix}
\usepackage{titletoc}

\definecolor{mycitecolor}{HTML}{6C94F8}
\usepackage[colorlinks=true, citecolor=mycitecolor, linkcolor=mycitecolor,
urlcolor=mycitecolor]{hyperref}
\setcitestyle{authoryear,round,citesep={;},aysep={,},yysep={;}}

\newcommand{\cmark}{\textcolor{green!60!black}{\ding{51}}}
\newcommand{\xmark}{\textcolor{red!80!black}{\ding{55}}}

\definecolor{table1_lightgreen}{HTML}{E5F7E4}
\definecolor{table1_mediumgreen}{HTML}{D0F0CF}
\definecolor{table1_deepgreen}{HTML}{9FE09A}

\usepackage[utf8]{inputenc}
\usepackage{array}

\usepackage{enumitem}
\setlist[itemize]{itemsep=1pt, leftmargin=*}

\usepackage{colortbl}

\definecolor{MyDarkGreen}{RGB}{45,155,45}
\definecolor{MyDarkRed}{rgb}{0.8,0.02,0.02}
\definecolor{MyRed}{rgb}{0.8,0.0,0.0}
\definecolor{MyGold}{rgb}{0.75,0.6,0.12}
\definecolor{MyDarkgray}{rgb}{0.66, 0.66, 0.66}
\definecolor{lightblue}{RGB}{195, 230, 252}
\definecolor{lightorange}{RGB}{255, 218, 185}
\definecolor{lightpurple}{RGB}{229, 198, 253}
\definecolor{lightgreen}{RGB}{197, 232, 201}
\definecolor{ours_green}{RGB}{100, 180, 96}
\definecolor{reason}{RGB}{55, 130, 240}
\definecolor{vision}{RGB}{243, 145, 29}
\definecolor{answer}{RGB}{111,0,255}

\newcommand{\ours}{\textsc{\textcolor{ours_green}{CU}\textcolor{reason}{R}\textcolor{vision}{V}}\xspace}

\newcommand{\ourswicon}{%
  \texorpdfstring{%
    \includegraphics[height=3.0ex]{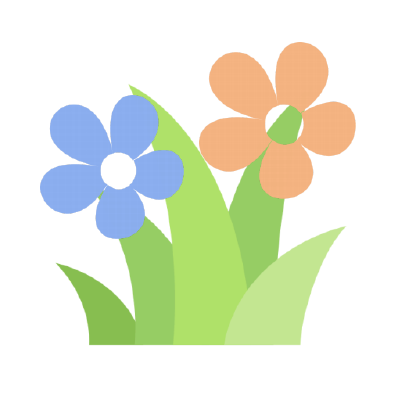}\,\textsc{\textcolor{ours_green}{CU}\textcolor{reason}{R}\textcolor{vision}{V}}%
  }{%
    \textsc{CURV}%
  }\xspace%
}

\newcommand{\oursdata}{\textsc{CCQA}\xspace}

\newcommand{\modeA}{\textcolor{answer}{\textbf{A}}\xspace}
\newcommand{\modeR}{\textcolor{reason}{\textbf{R}}\xspace}
\newcommand{\modeV}{\textcolor{vision}{\textbf{V}}\xspace}

\newcommand{\modeRA}{\textcolor{reason}{\textbf{R}}\textcolor{answer}{\textbf{A}}\xspace}
\newcommand{\modeVA}{\textcolor{vision}{\textbf{V}}\textcolor{answer}{\textbf{A}}\xspace}
\newcommand{\modeRV}{\textcolor{reason}{\textbf{R}}\textcolor{vision}{\textbf{V}}\xspace}
\newcommand{\modeRVA}{\textcolor{reason}{\textbf{R}}\textcolor{vision}{\textbf{V}}\textcolor{answer}{\textbf{A}}\xspace}

\definecolor{xuehang_color}{rgb}{1.0, 0.4, 0.2}

\NewDocumentCommand{\qingyun}
{ mO{} }{\textcolor{cyan}{\textsuperscript{\textit{Qingyun}}\textsf{\textbf{\small[#1]}}}}

\NewDocumentCommand{\heng}
{ mO{} }{\textcolor{red}{\textsuperscript{\textit{Heng}}\textsf{\textbf{\small[#1]}}}}

\title{\ourswicon: Enhancing \underline{C}hart \underline{U}nderstanding Through Curriculum \underline{V}isual Grounded \underline{R}easoning}

\makeatletter
\newcommand{\blfootnote}[1]{%
  \begingroup
  \renewcommand{\thefootnote}{}%
  \renewcommand{\@makefntext}[1]{\noindent##1}%
  \footnote{#1}%
  \addtocounter{footnote}{-1}%
  \endgroup
}
\makeatother

\author{
\textbf{Xuehang Guo}\textsuperscript{1},\quad
\textbf{Pingyue Zhang}\textsuperscript{2},\quad
\textbf{Ruiyi Zhang}\textsuperscript{3},\quad
\textbf{Zhenhailong Wang}\textsuperscript{4},\quad
\textbf{Hanrui Lyu}\textsuperscript{2},\\[0.5ex]
\textbf{Heng Ji}\textsuperscript{4},\quad
\textbf{Tong Sun}\textsuperscript{3},\quad
\textbf{Qingyun Wang}\textsuperscript{1},\quad
\textbf{Manling Li}\textsuperscript{2}\\[1.2ex]
\textsuperscript{1}William \& Mary\qquad
\textsuperscript{2}Northwestern University\qquad
\textsuperscript{3}Adobe\qquad
\textsuperscript{4}UIUC
}

\newcommand{\Fref}[1]{Fig.~\ref{#1}}
\newcommand{\Tref}[1]{Tab.~\ref{#1}}
\newcommand{\Sref}[1]{\S~\ref{#1}}
\newcommand{\Eref}[1]{Eq.~\ref{#1}}

\begin{document}

\ifcolmsubmission
\linenumbers
\fi

\maketitle
\blfootnote{\textbf{Correspondence to:} Manling Li (\url{manling.li@northwestern.edu}), Qingyun Wang (\url{qwang16@wm.edu}), Xuehang Guo (\url{xguo15@wm.edu})}

\begin{abstract}
Chart question answering (CQA) requires multimodal large language models (MLLMs) to integrate visual comprehension with logical reasoning, yet current models struggle with accurate visual grounding and coherent reasoning chains. 
While extrinsic chain-of-thought prompting and visual cues significantly improve performance, current MLLMs lack intrinsic visual grounded reasoning capabilities, leading to inaccurate perception and reasoning disconnected from visual evidence.
To address these limitations, we propose \ours, a curriculum learning framework that develops intrinsic visual reasoning capabilities by reformulating CQA as multi-step visual grounded reasoning, where each step coordinates logical reasoning with dynamic visual grounding through spatial attention concentration.
To assist model learning, we further introduce \oursdata, a three-level curriculum dataset with scalable synthetic generation across diverse chart types and reasoning patterns. Our curriculum systematically progresses from basic single-operation reasoning to complex multi-chart compositional tasks.
Experiments demonstrate that \ours achieves up to $\uparrow20.50\%$ improvements over baselines and is generalizable to real-world benchmarks (up to $\uparrow12.30\%$)
and out-of-domain multimodal reasoning tasks (up to $\uparrow10.20\%$), validating the effectiveness of internalizing visual reasoning with dynamic grounding for enhanced chart understanding capabilities. Code is available at: \url{https://xhguo7.github.io/CURV/}.
\end{abstract}

%%%%%%  Abstract - Old Versions  %%%%%%
% Chart question answering (CCQA) requires multimodal large language models (MLLMs) to integrate visual comprehension with logical reasoning, yet current models struggle with accurate visual grounding and coherent reasoning chains. Inspired by how humans decompose complex chart understanding through systematic visual attention shifts and step-by-step reasoning, we aim to endow MLLMs with enhanced intrinsic visual reasoning capabilities by simulating human reasoning process through surpervised finetuning. We propose \ours, a curriculum learning framework that develops intrinsic grounded visual reasoning capabilities by reformulating CQA as multi-turn visual reasoning, where each step coordinates logical reasoning with dynamic visual grounding. To assist model learning, we further introduce \oursdata, a three-level curriculum dataset with scalable synthetic generation across diverse chart types and reasoning patterns. Our curriculum systematically progresses from basic single-operation reasoning to complex multi-chart compositional tasks. Experiments demonstrate that \ours achieves up to $10.79\%$ accuracy improvements over baselines and strong generalization to real-world benchmarks (up to $3.50\%$ improvements) and out-of-domain multimodal reasoning tasks (more than $1.30\%$ accuracy gains), validating the effectiveness of simulating human-like decomposed reasoning for enhanced multimodal understanding.

% \input{sections/xuehang_outline}
% \input{sections/thx!}

\begin{figure}[htbp]
    \vspace{0pt}

    \small
    \centering
    \includegraphics[width=1.0\textwidth]{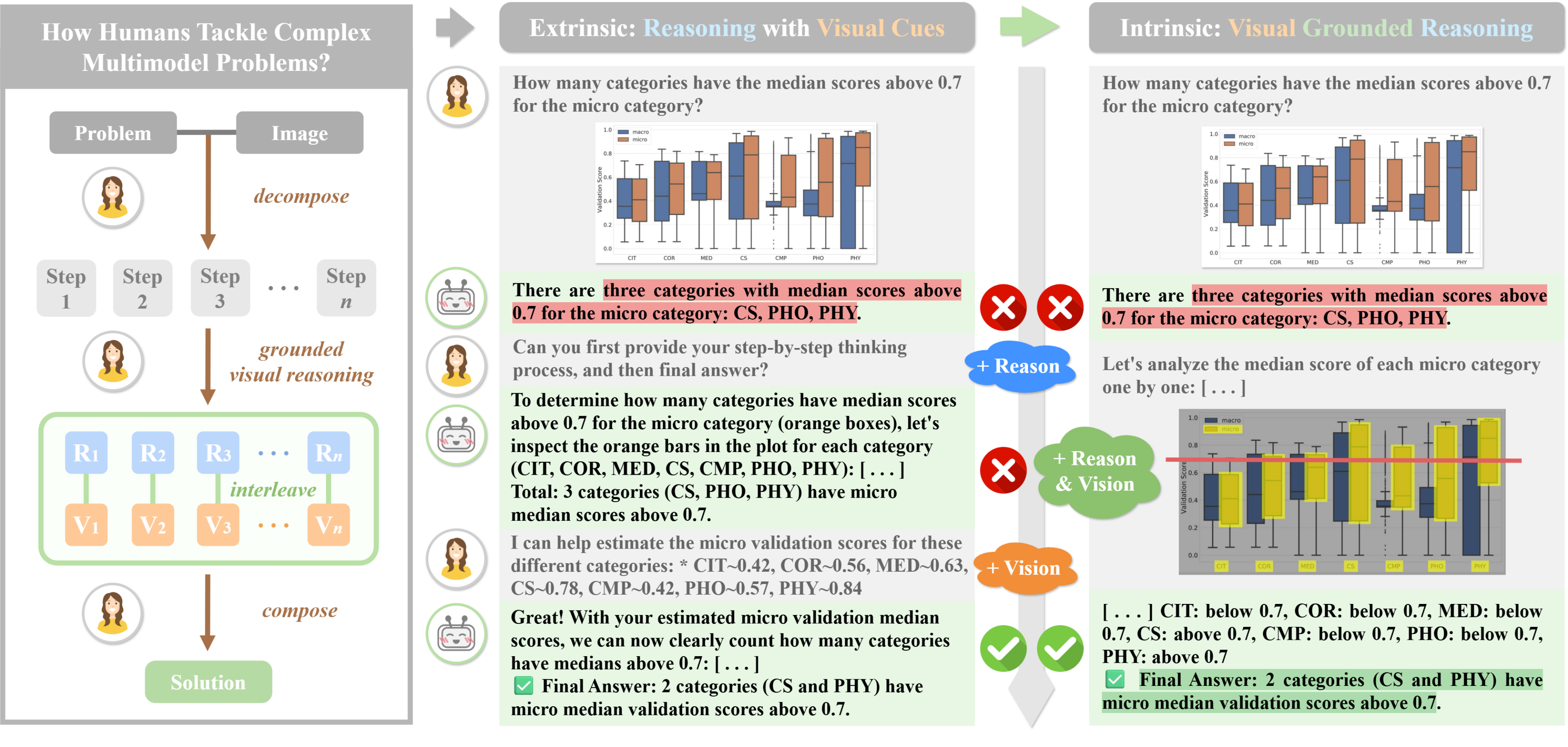}
    
    \vspace{-6pt}
    
    \caption{\textbf{From Extrinsic Assistance to Intrinsic Visual Grounded Reasoning.} Inspired by human ways of thinking, we present \ours to internalize extrinsic \textbf{\textcolor{reason}{CoT prompting}} and \textbf{\textcolor{vision}{visual guidance}} to intrinsic capabilities, enabling models to perform \textbf{\textcolor{ours_green}{visual grounded reasoning}} through dynamically shifting focuses across targeted image regions.}
    \label{fig:cover}

    \vspace{0pt}
\end{figure}

\section{Introduction}
\label{sec:intro}

\textit{How do humans tackle multimodal problems?} Inspired by cognitive theories \citep{Baddeley1974WorkingMemory,johnson1983mentalmodels,Barsalou2008GroundedCA,grant2003eye}, humans \textbf{decompose} complex tasks into stepwise reasoning chains, \textbf{interleave} each step with dynamic visual grounding, and \textbf{compose} these grounded steps into a coherent solution (\Fref{fig:cover}).
Chain-of-thought (CoT) reasoning has demonstrated its effectiveness in decomposing problems into stepwise inferences \citep{xu-etal-2024-faithful,zhang-etal-2025-cot}.
This ability becomes more critical in multimodal reasoning, where multimodal large language models (MLLMs) are expected to integrate visual and textual information \citep{fan-etal-2025-unveiling,zhang-etal-2025-improve} while visual perception errors contribute to the majority of multimodal reasoning failures \citep{wang2025papo}.
Without external support such as explicit CoT prompting or visual cues (\Fref{fig:cover}), MLLMs struggle with accurate visual grounded reasoning \citep{Wang2025VisuallyDescriptive,Wang2025VGR,Wang2025TreeBench}.

%%% 2. Challenges
This challenge is particularly evident in chart question answering (CQA), where models need to faithfully interpret complex geometric structures, spatial relationships, and quantitative patterns to derive correct answers.
As a result, CQA requires models to accurately perceive visual details, perform step-by-step reasoning over interconnected components, and dynamically shift focus across different chart regions \citep{chen2025mintcot}.
However, existing MLLMs exhibit several fundamental limitations in this setting (\S\ref{sec:preliminary} \& \ref{appendix:preliminary}):
\textbf{(1) Decomposition:} \textit{They struggle to decompose complex problems into coherent chains of reasoning, often producing inconsistent or logically flawed intermediate steps (\Fref{fig:appendix_preliminary_evidence_1_charxiv}-\ref{fig:appendix_preliminary_evidence_2_chartmuseum})};
\textbf{(2) Interleaved Visual Reasoning:} \textit{They show limitations in accurately grounding individual reasoning steps in the visual input, such as misreading chart values or attending to correct regions (\Fref{fig:appendix_preliminary_evidence_1_charxiv})};
and \textbf{(3) Composition:} \textit{They exhibit difficulties in integrating logical reasoning with visual grounding across multiple steps into a coherent, interleaved chain, leading to a disconnect between what is perceived, reasoned, and concluded (\Fref{fig:appendix_preliminary_evidence_2_chartmuseum})}.
Collectively, these limitations lead to inaccurate perception and reasoning that is disconnected from the visual evidence, ultimately causing errors even when the necessary information is present.

To address these limitations, we propose \textbf{\ours}, a curriculum learning (CL) framework that develops intrinsic visual grounded reasoning capabilities in MLLMs.
Our approach reformulates CQA as multi-step reasoning processes where each step couples logical reasoning with dynamic visual grounding.
Instead of relying on extrinsic assistance, \ours enables models to internalize the ability through dynamically focusing on relevant chart regions while maintaining coherent reasoning chains across steps.
Curriculum learning progresses from single-operation reasoning to complex multi-operation compositions, allowing models to gradually develop both visual perception accuracy and reasoning sophistication.
Our main contributions include:
\begin{itemize}[topsep=-4pt, itemsep=1pt, parsep=1pt]
    \item We propose \textbf{\ours}, a curriculum learning framework that develops MLLMs' intrinsic visual reasoning capabilities by progressively shifting visual attention along reasoning, transitioning from basic single-operation tasks to complex nested reasoning (\S\ref{sec:methodology}).
    \item We introduce a scalable synthetic CQA data generation method (\S\ref{sec:dataset_construction}) that systematically increases task complexity via nested reasoning chains, enabling efficient data creation.
    \item We present \textit{\underline{C}urriculum \underline{C}hart \underline{Q}uestion \underline{A}nswering} (\textbf{\oursdata}) (\S\ref{sec:dataset_construction}) that supports curriculum learning across four different generation modes, three visual grounding strategies, and multi-level metrics for comprehensive performance evaluation (\S\ref{sec:preliminary} \& \ref{sec:experiments}).
    \item Our experiments demonstrate that CoT reasoning with visual grounding provides models with step-by-step alignment between visual perception and logical reasoning, leading to notable improvements across different chart types, complexity, and domains (\S\ref{sec:experiments}).
\end{itemize}

\section{What Prohibits MLLMs From Chart Understanding Success?}
\label{sec:preliminary}

\begin{wraptable}[11]{r}{0.5\textwidth}
\vspace{0pt}

\centering
\small

\renewcommand{\arraystretch}{1.5}
\resizebox{0.5\textwidth}{!}{%
\begin{tabular}{cccccc}
\toprule
\multirow{2}{*}{\textbf{Mode}} & \multicolumn{3}{c}{\textbf{Error Analysis (\textit{count})}} & \multirow{2}{*}{\textbf{Acc (\%)}} & \multirow{2}{*}{\textbf{$\Delta_{acc}$ (\%)}} \\
\cmidrule(lr){2-4}
 & \cellcolor{lightorange}\textbf{Vision} & \cellcolor{lightblue}\textbf{Reasoning} & \cellcolor{lightpurple}\textbf{Answer} & & \\
\midrule
\modeA & -- & -- & \cellcolor{red!57}\textbf{34} & 43.33 & -- \\
\modeVA & \cellcolor{red!0}0 & \cellcolor{red!39}\textbf{18} & \cellcolor{red!10}5 & 61.67 & \cellcolor{table1_mediumgreen}\textbf{$\uparrow 18.33$} \\
\modeRA & \cellcolor{red!24}\textbf{17} & \cellcolor{red!15}6 & \cellcolor{red!10}5 & 53.33 & \cellcolor{table1_lightgreen}\textbf{$\uparrow 10.00$} \\
\modeRVA & \cellcolor{red!0}0 & \cellcolor{red!10}5 & \cellcolor{red!6}3 & \textbf{86.67} & \cellcolor{table1_deepgreen}{$\uparrow \mathbf{43.33}$} \\
\bottomrule
\end{tabular}%
}

\vspace{-6pt}

% \caption{\footnotesize \textbf{CQA Preliminary.} Preliminary results on GPT-4o using accuracy (\%) and relative improvements $\Delta_{acc}$ (\%). Error analysis counts the number of cases.}
\caption{\footnotesize \textbf{CQA Preliminary.} Preliminary results on GPT-4o using accuracy (\%) and relative improvements $\Delta_{acc}$ (\%).}
\label{tab:preliminary_gpt4o_conditions}

\end{wraptable}

\setlength{\intextsep}{0pt}

\subsection{Preliminary Analysis of CQA Failures}
\label{subsec: motivation 1 - find current weakness}

\textit{What are the bottlenecks that hinder MLLMs from correctly understanding chart images?}

We evaluate GPT-4o \citep{openai2024gpt4o} on 60 CQA samples randomly selected from CharXiv with four modes: \textbf{(1) Answer only (\modeA):} MLLM directly generates the answer;
\textbf{(2) Vision + Answer (\modeVA):} MLLM is provided with human-annotated visual information to give the answer;
\textbf{(3) Reason + Answer (\modeRA):} MLLM is prompted to first generate CoT reasoning, followed by the final answer;
\textbf{(4) Reason + Vision + Answer (\modeRVA):} MLLM is provided with the same visual information as in \textbf{\modeVA}, prompted to first generate CoT reasoning and then the final answer.
Surprisingly (Tab.~\ref{tab:preliminary_gpt4o_conditions}), GPT-4o achieves significantly higher scores ($\uparrow 10.00\%$) when prompted to generate reasoning without visual cues (\textbf{\modeRA}). Moreover, combining perception with reasoning (\textbf{\modeRVA}) yields the best performance ($\uparrow 43.33\%$).
These highlight \textit{MLLMs' lack of logical decomposition and visual reasoning capabilities}.

\subsection{Enhancing Chart Reasoning Through Dynamic Visual Grounding}
\label{subsec: motivation 2 - validate our method}

% Preliminary Test - Applied

\begin{wrapfigure}[17]{r}{0.5\textwidth}
    \small
    \centering
    \vspace{-3pt}
    \includegraphics[width=0.5\textwidth]{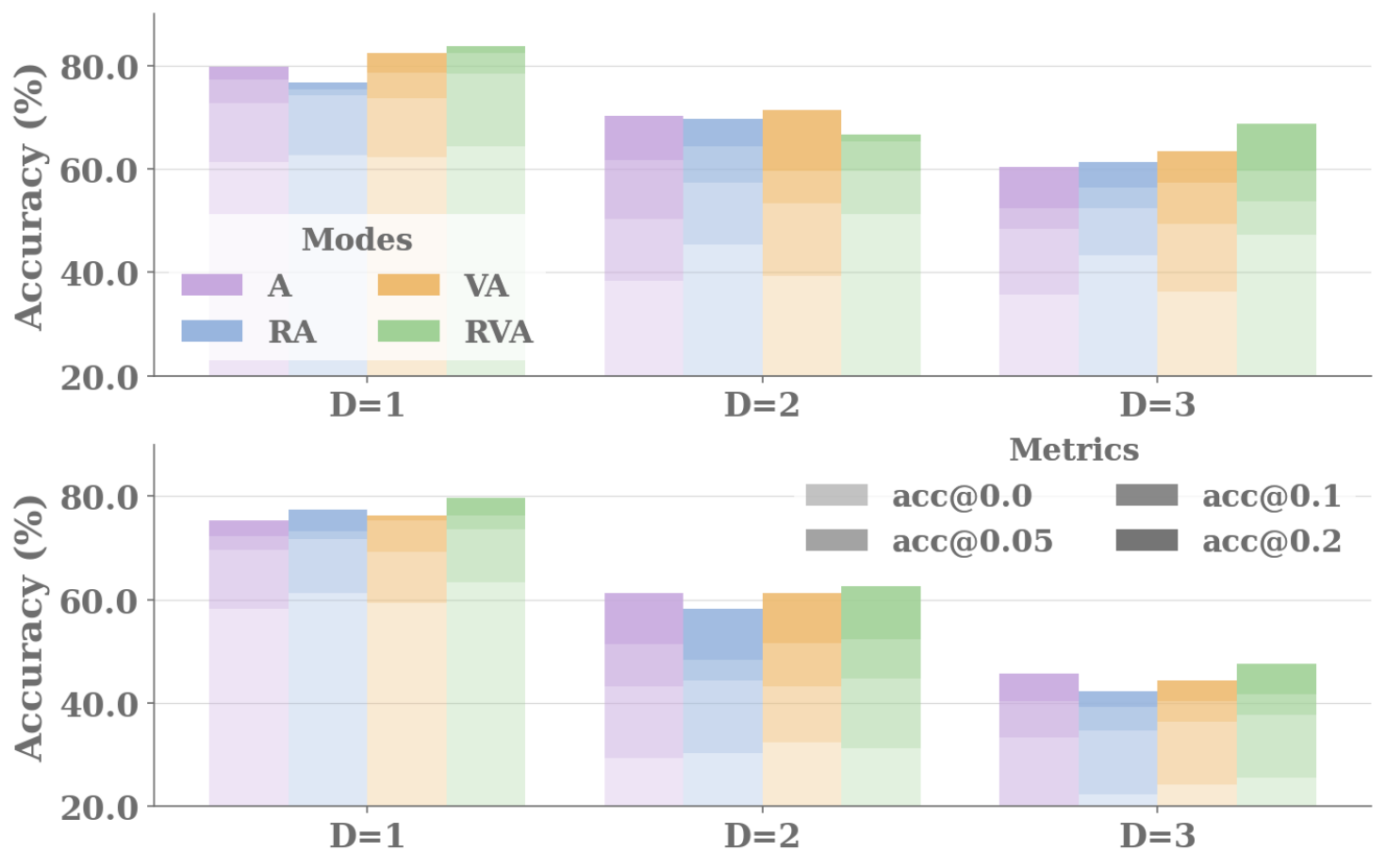}
    \vspace{-18pt}
    \caption{\textbf{Performance Across Reasoning Depths and Modes.}
    We evaluate GPT-4o (\textit{upper}) and Qwen2.5-VL-7B (\textit{lower}) on 1,800 randomly selected samples from \oursdata (evenly distributed across $D_i, 1 \leq i \leq 3$).}
    \label{fig:preliminary_model_comparison}
\end{wrapfigure}

\setlength{\intextsep}{0pt}

\textit{How to improve MLLMs' intrinsic visual reasoning capabilities?}

Motivated by the effectiveness of extrinsic CoT prompting and visual guidance (\Tref{tab:preliminary_gpt4o_conditions}), we aim to internalize these capabilities within MLLMs (\Fref{fig:cover}).
To concretize this approach, we expand our preliminary exploration to visual reasoning with grounded focuses (Tab.~\ref{fig:preliminary_model_comparison}) using \oursdata (\S\ref{sec:dataset_construction}) and two models: Qwen2.5-VL-7B \citep{qwen2.5_report} and GPT-4o \citep{openai2024gpt4o}.
Similarly, we examine model performance under four modes using ground-truth reasoning and visual grounding.
%
%
%
%%%%%%%%%%%%%%%%%%%%%%%%%%%%%%
%        Motivation
%%%%%%%%%%%%%%%%%%%%%%%%%%%%%%
Despite increased curriculum difficulty ($1 \leq D \leq 3$), both models achieve higher performance when equipped with either reasoning or visual assistance, with performance further improved when both are combined.
Consistently, extrinsic assistance enhances chart understanding through structured reasoning and visual grounding, motivating our core hypothesis: \textit{MLLMs can internalize these capabilities through visual grounded reasoning} (\S\ref{sec:methodology} \& \ref{sec:dataset_construction}).

%%%%%%%%%%%%%%%%%%%%%%%%%%%%%%
%        Methodology
%%%%%%%%%%%%%%%%%%%%%%%%%%%%%%

\section{\textbf{\ours}: Chart Reasoning with Dynamic Visual Grounding}
\label{sec:methodology}

\subsection{Problem Formulation}
\label{subsec:problem_formulation}

\textbf{Problem Definition.} Given a chart image $\mathcal{I} \in \mathbb{R}^{H \times W \times C}$ and a question $Q$, the goal of CQA is to generate the answer $A$. However, current MLLMs directly learn the mapping:

\vspace{-5.6mm}
\begin{equation}
f_{\theta}: (\mathcal{I}, Q) \rightarrow A
\end{equation}
\vspace{-6.6mm}

Not only does this direct mapping approach lack an intermediate reasoning structure that enables accurate visual perception and robust visual understanding, but it also fails to effectively and dynamically ground reasoning chains in visual space.

\textbf{Our Approach.} We propose to decompose this problem into a multi-step reasoning process with dynamic visual grounding. Specifically, we reformulate the CQA task as:

\vspace{-4.6mm}
\begin{equation}
f_{\theta}: (\mathcal{I}, Q) \rightarrow \{(R_1, V_1), (V_2, V_2), \ldots, (R_T, V_T)\} \rightarrow A
\label{equation:ours}
\end{equation}
\vspace{-5.6mm}

where
$R_t$ represents the $t$-th reasoning step in natural language,
$V_t$ denotes visual regions grounding $R_t$ in the visual space,
$T$ is the total number of reasoning steps to reach $A$,
and the sequence $\{(R_t, V_t)\}_{t=1}^T$ forms a structured progressive visual reasoning chain.

\begin{figure}[!t]
    \vspace{-16pt}

    \small
    \centering
    \includegraphics[width=1.0\textwidth]{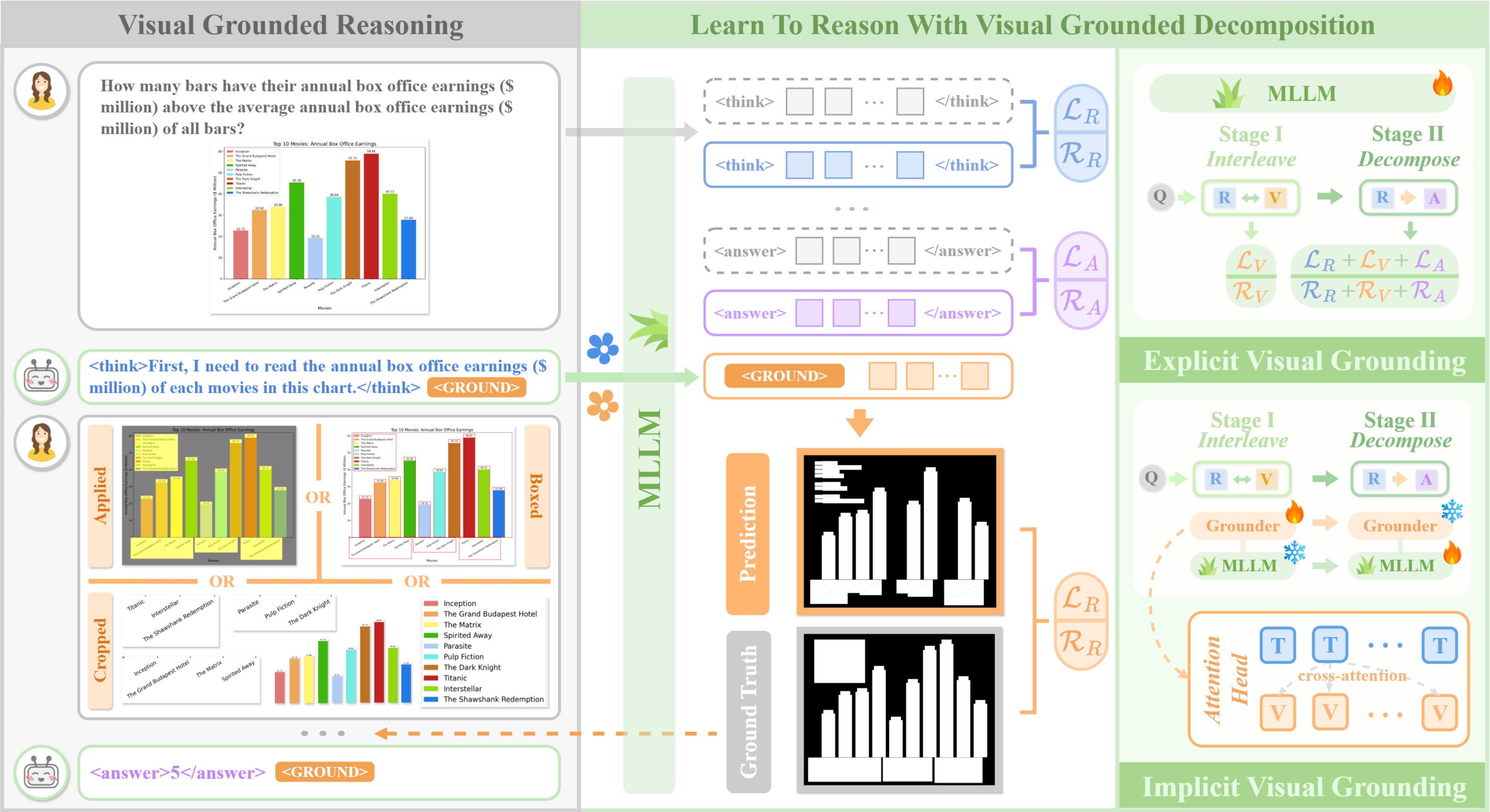}
    
    \vspace{-6pt}
    
    \caption{\textbf{\ours Overview.} Different visual grounding (\S\ref{appendix: 3 visual grounding strategies}) and training (\S\ref{appendix:sec:in_depth_analysis}) approaches to develop intrinsic \textbf{\textcolor{vision}{visual}} \textbf{\textcolor{reason}{reasoning}} capabilities via \textbf{\textcolor{ours_green}{multi-step visual grounded reasoning}}.}
    \label{fig:methodology}

    \vspace{-3pt}

\end{figure}

\subsection{Multi-Step Reasoning With Dynamic Visual Grounding}
\label{subsec:multi_turn_framework}

We design a two-stage multi-step curriculum learning framework (\ours) that enables MLLMs to develop intrinsic progressive reasoning capabilities with dynamic visual grounding, moving beyond extrinsic assistance toward self-contained visual reasoning (\Fref{fig:cover}). To further strengthen this process, we implement \textit{explicit} and \textit{implicit} visual grounding (\S\ref{appendix:subsec:explicit_vs_implicit}).

\textbf{Stage I: Visual Grounding (\modeRV).}
In the first stage, the model learns to establish multimodal correspondences between reasoning flows and associated visual focuses.
Specifically, given the chart image $\mathcal{I}$ and question $Q$, at time $t$, the model learns to predict the visual focus $V_t$ that grounds $R_t$ onto $\mathcal{I}$:

\vspace{-7.6mm}
\begin{equation}
V_t = f_{\theta}^{(S1)}\!\left(\mathcal{I},\, Q,\, \{R_{t'}, V_{t'}\}_{t'=1}^{t-1},\, R_t \right)
\label{equation:stage1_generate}
\end{equation}
\vspace{-4.6mm}

\noindent where prior grounding pairs $\{(R_{t'}, V_{t'})\}_{t'=1}^{t-1}$ provide context for consistent grounding across steps.
The training objective supervises the predicted visual focuses:

\vspace{-4.6mm}
\begin{equation}
\mathcal{L}^{(S1)} = \sum_{t=1}^{T} \mathcal{L}_V(V_t,\, V^*_t)
\label{equation:stage1_loss}
\end{equation}
\vspace{-4.6mm}

\noindent where $\mathcal{L}_V(V_t, V^*_t)$ denotes the grounding loss between the predicted visual focus $V_t$ and the ground-truth $V^*_t$ at step $t$.

\textbf{Stage II: Interleaved Visual Reasoning (\modeRVA).}
Building upon the visual grounding capability acquired in \textit{Stage I}, the model learns to \textit{proactively leverage} visual grounding as feedback during reasoning.
Concretely, at each step $t$, the visual focus $V_t$ produced alongside reasoning $R_t$ is applied to the input chart image $\mathcal{I}$ to construct the grounded visual state $\mathcal{I}_t$ via one of the visual grounding methods (\Fref{fig:methodology}).
This grounded image $\mathcal{I}_t$ is then provided as additional visual input when generating the \textit{next} reasoning step, enabling the model to dynamically shift its visual attention in alignment with the evolving logical reasoning:

\vspace{-4.6mm}
\begin{equation}
(R_t, V_t) = f_{\theta}^{(S2)}\!\left(\mathcal{I},\, Q,\, \left\{R_{t'},\, V_{t'} \!\rightarrow\! \mathcal{I}_{t'}\right\}_{t'=1}^{t-1}\right)
\label{equation:stage2_interleave}
\end{equation}
\vspace{-5.6mm}

where $V_{t'} \rightarrow \mathcal{I}_{t'}$ denotes the new state mapped into $\mathcal{I}_{t'}$.
\textit{Stage II}'s training objective is:

\vspace{-5.0mm}
\begin{equation}
\mathcal{L}^{(S2)} = \sum_{t=1}^{T} \lambda_R\mathcal{L}_R(R_t,\, R^*_t) + \lambda_V\sum_{t=1}^{T} \mathcal{L}_V(V_t,\, V^*_t) + \lambda_A\mathcal{L}_A(A,\, A^*)
\label{equation:stage2_loss}
\end{equation}
\vspace{-4.0mm}

\noindent where $\mathcal{L}_R(R_t, R^*_t)$ is the reasoning supervision loss at step $t$, $\mathcal{L}_V(V_t, V^*_t)$ is the grounding loss at step $t$, $\mathcal{L}_A(A, A^*)$ is the final answer prediction loss, and $\lambda_R$, $\lambda_V$, and $\lambda_A$ balance the visual grounded reasoning objective against answer supervision.
As such, \textit{Stage II} develops the model's intrinsic progressive visual reasoning capabilities, enabling it to:
(1) \textit{dynamically focus} on relevant chart regions by updating visual grounding at each step,
(2) \textit{ground logical reasoning} in concrete visual evidence, and
(3) \textit{maintain cross-step coherence} through joint conditioning on both the textual reasoning history and the evolving visual states.

\begin{figure}[H]
    \centering

    \vspace{16pt}
    \includegraphics[width=1.0\textwidth]{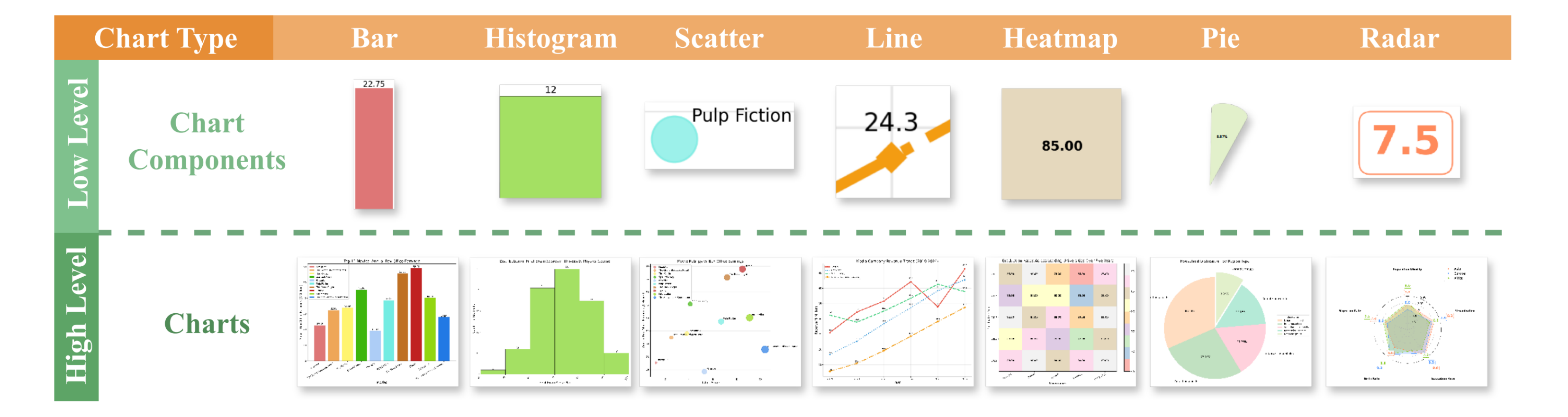}
    
    \vspace{-10pt}
    
    \caption{\textbf{From Low-Level Chart Components To High-Level Charts.} We decompose all types of charts into low-level components to endow MLLMs with both foundational chart understanding abilities and adaptive generalizabilities to high-level complexities.}
    \label{fig:meta_learning}

    \vspace{-16pt}
    
\end{figure}

\subsection{Curriculum Visual Grounded Reasoning}
\label{subsec:curriculum_learning}

\textbf{Visual Grounding.}
In addition to \textit{explicit} and \textit{implicit} visual reasoning (\S\ref{appendix:subsec:explicit_vs_implicit}), we propose three visual grounding strategies (\S\ref{appendix: 3 visual grounding strategies}), including \textcolor{vision}{\textbf{applied}}, \textcolor{vision}{\textbf{boxed}}, and \textcolor{vision}{\textbf{cropped}}, to dynamically shift visual focuses during reasoning (\Fref{fig:methodology}).

\textbf{Reasoning Depth.}
\label{subsubsec:concept_reasoning_depth}
To formalize reasoning complexity in CQA, we introduce two distinct but complementary concepts that characterize the reasoning process:
\begin{itemize}[topsep=-4pt, itemsep=1pt, parsep=1pt]
    \item \textbf{Number of Reasoning Steps ($T$):} The total number of CoT reasoning steps $\{R_t\}_{t=1}^T$ a model goes through to reach the final answer $A$. Multiple reasoning steps may operate at the same logical complexity level (\textit{i.e.}, \textit{reasoning depth tier} below) while contributing different pieces of information toward the solution.
    \item \textbf{Tier of Reasoning Depth ($D$):} The maximum number of nested logical functions required to solve the task, corresponding to the deepest level of functional composition in the reasoning chain. Formally, for a question requiring nested functions $f_1(f_2(...(f_D(x))))$, the reasoning depth is $D$. This metric captures the inherent logical complexity of a problem, independent of how many intermediate steps a model uses to reach $A$.
\end{itemize}
\vspace{5pt}

\textbf{Curriculum Learning.}
Charts are a unique form of data organized through structured relationships among fundamental meta-elements (Fig.~\ref{fig:meta_learning}).
Through two-stage \textbf{\textit{training curriculum}}, we incorporate \textbf{\textit{chart curriculum}} to guide the model from understanding low-level visual components to high-level chart structures (Fig.~\ref{fig:meta_learning}) across diverse chart types.
With gradually increased task difficulty along both reasoning and visual dimensions, we define data curriculum consisting of three \textit{levels} (\S\ref{subsec:curriculum_construction}) across five fine-grained \textit{tiers} (\S\ref{appendix:five fine-grained curriculum tiers}).

\textbf{\ours Variants.}
\ours accommodates multiple design choices along two orthogonal dimensions:
First, it supports three visual grounding strategies (\S\ref{appendix: 3 visual grounding strategies}) that control the representation and granularity of visual grounding.
We further implement \textit{explicit} visual grounding to develop MLLMs' intrinsic visual grounded reasoning capabilities (\S\ref{subsec:multi_turn_framework}), while performing ablations of \textit{implicit} grounding (\S\ref{appendix:subsec:explicit_vs_implicit}) to validate the critical role of visual grounding in \ours as intermediate visual-reasoning alignment guidance rather than an ultimate learning objective (\S\ref{subsec:ablation_results}).
Second, \ours is compatible with different training paradigms, including \textit{SFT}, \textit{reinforcement learning (RL)}, and \textit{their combination}, enabling flexible instantiations under varying optimization objectives and resource constraints (\S\ref{subsec:main_results} \& \ref{appendix:subsec:rl_vs_sft}).

\section{\textbf{\oursdata}: Curriculum Chart Question Answering}
\label{sec:dataset_construction}

\subsection{Dataset Construction Principles}
\label{subsec:dataset_principles}

\begin{figure}[!t]
    \centering

    \vspace{-16pt}
    \includegraphics[width=1.0\textwidth]{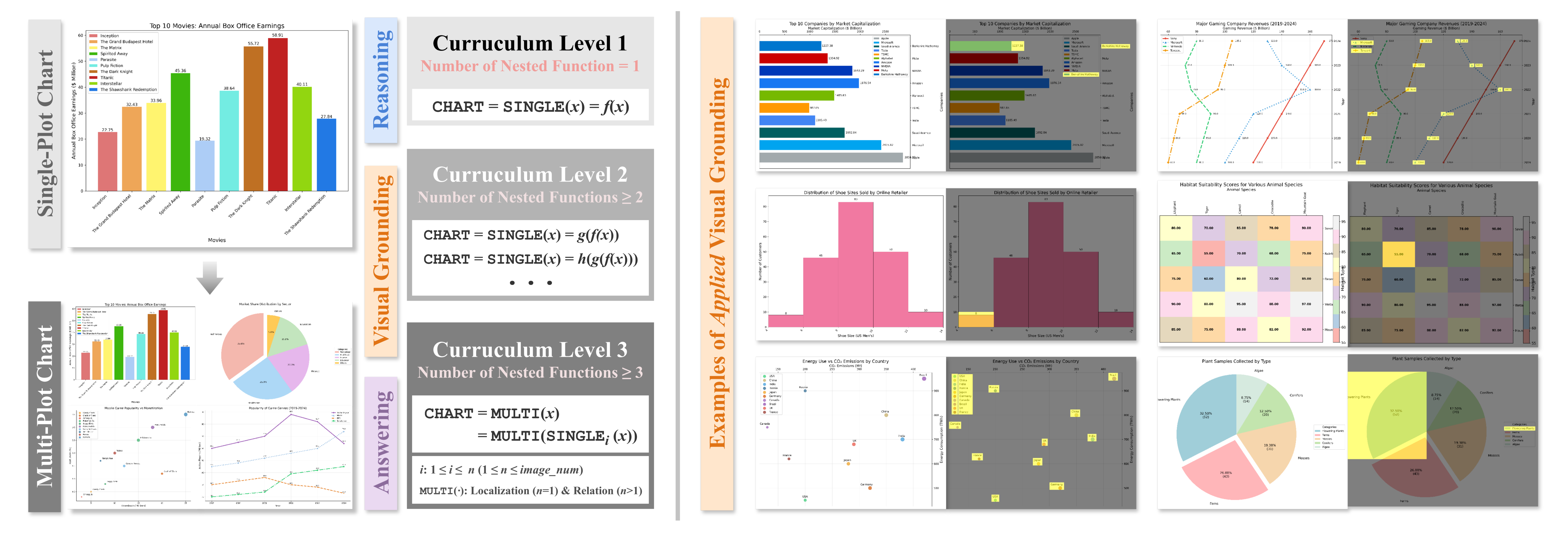}
    
    \vspace{-6pt}
    
    \caption{\textbf{Multi-Level Curriculum Construction.} We construct \oursdata through reasoning decomposition, interleaving visual reasoning, and reasoning chain composition. Examples of \textcolor{vision}{\textbf{applied}} visual grounding is shown on the right. More examples can be found in \Sref{appendix: 3 visual grounding strategies}.}
    \label{fig:dataset_construction}

    \vspace{0pt}
    
\end{figure}

Supporting \ours (\S\ref{sec:methodology}), we introduce \textit{Curriculum Chart Question Answering} (\textbf{\oursdata}), a curriculum learning dataset to progressively develop visual reasoning capabilities. Our dataset construction presents three core principles with rigorous quality control (\S\ref{appendix:subsec:dataset_construction_quality_control}):

\textbf{Progressive Complexity.}
We implement three-level curriculum through systematic variation in reasoning depth ($D$), chart complexity,
% (\textit{i.e.,} number of elements and subplots), 
and operation sophistication (\Fref{fig:dataset_construction}).
% (\textit{i.e.,} basic to nested).

\textbf{Interleaved Visual Reasoning.}
Each reasoning step $R_t$ is paired with corresponding ground-truth $V_t^*$ and binary masks $M_t^*$, enabling direct alignment of visual grounding.

\textbf{Template-Based Accuracy.}
We employ synthetic templates (\S\ref{appendix:subsec:data_structure}) to ensure data accuracy and systematic coverage of reasoning patterns,
% while maintaining structural consistency across difficulty levels.
effectively supporting a stable progression of curriculum.
As shown below, all chart-specific features are replaced by plotting data:

\vspace{-1.6mm}
\begin{center}
\tcbox[colback=gray!20, colframe=white, boxrule=0pt, left=1mm, right=1mm, top=1mm, bottom=1mm]{%
  \parbox{0.95\linewidth}{\small
    \centering
    \textbf{\textsc{Question:}} What is the $<\texttt{y\_axis\_title}>$ of the $<\texttt{object\_singular}>$?
  }%
}
\end{center}

\subsection{Curriculum Chart Question Answering}
\label{subsec:dataset construction}

% Pie chart for chart type distribution

\begin{wrapfigure}[8]{r}{0.4\textwidth}
    \vspace{-28pt}
    \centering
    \includegraphics[width=0.36\textwidth]{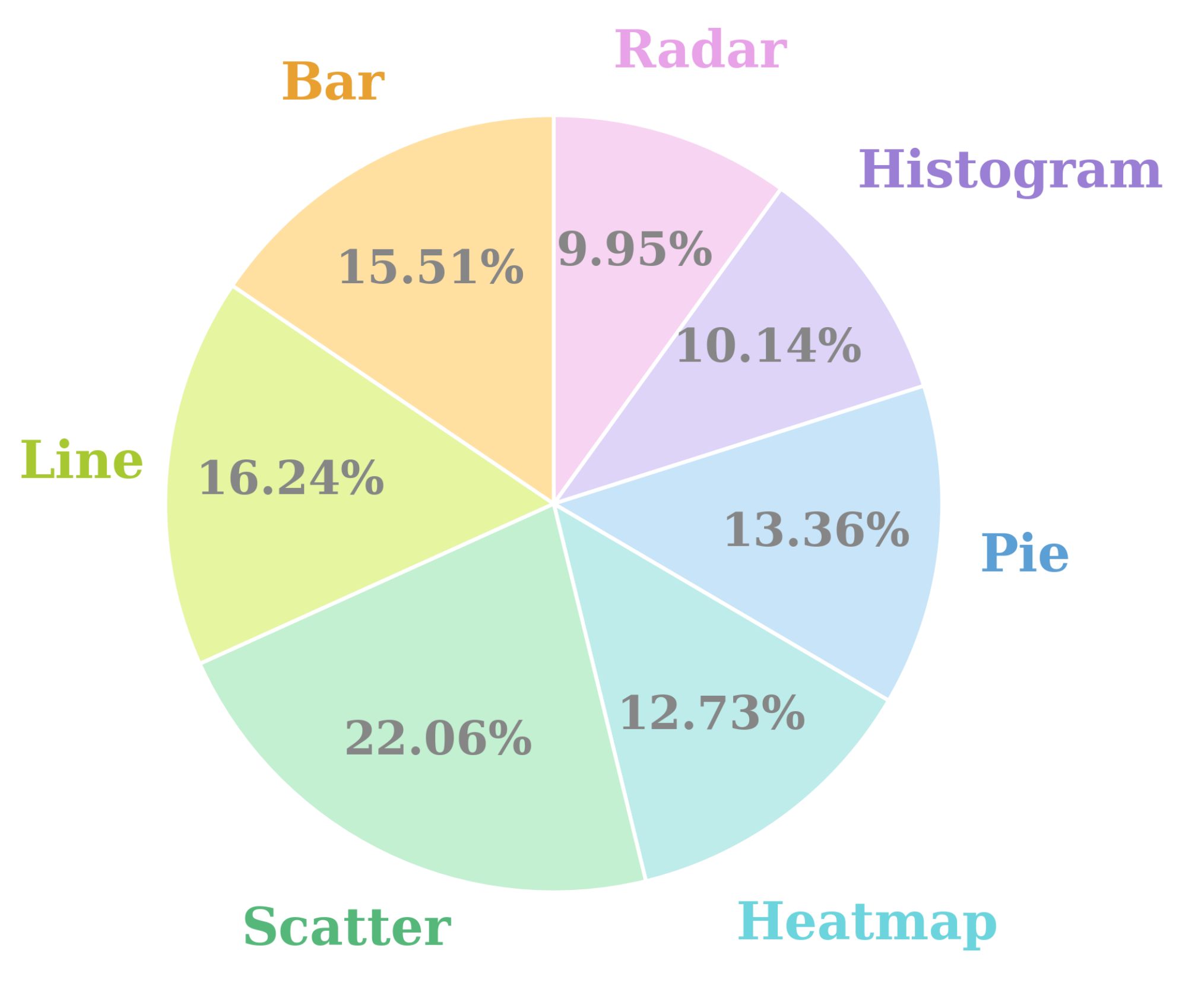}
    
    \vspace{-12pt}
    
    \caption{\textbf{Chart Type Distribution}
    % We employ seven chart types to construct \oursdata.
    }
    \label{fig:chart_type_distribution}
\end{wrapfigure}

\setlength{\intextsep}{0pt}

\textbf{Chart Types.}
We include 7 chart types to endow \oursdata with high visual diversity (\Fref{fig:chart_type_distribution} \& \Tref{tab:data_construct}): \textit{bar}, \textit{histogram}, \textit{scatter}, \textit{line}, \textit{heatmap}, \textit{pie}, and \textit{radar}.

\textbf{Data Category.}
We define 30 distinct domain categories (\Tref{tab:data_construct}), employing GPT-4o \citep{openai2024gpt4o} to generate plotting data for chart drawing (\Fref{fig:appendix_gpt4o_plotting_data_generation}).

\textbf{QA Types.}
Our curriculum templates (\S\ref{subsec:curriculum_construction}) cover various operations (\Tref{tab:cqa_operators}) in atomic or nested forms across different chart components and subplots.

\textbf{Curriculum with Meta Learning.}
Our dataset construction implements a \textit{meta-learning} paradigm that maximizes visual reasoning generalization while minimizing visual overfitting (\S\ref{appendix:subsec:meta-learning}).
With only 30 unique charts for each chart type (\textit{i.e.,} 7 types $\times$ 30 categories), diverse \textbf{\textit{query-reason-ground-answer}} quadruplets are derived from each image via systematic template instantiation (\S\ref{subsec:curriculum_construction}).
Aiming to foster MLLMs' intrinsic visual grounded reasoning abilities rather than memorizing specific visual appearances, we generate multiple CQA instances $\{(Q_k, \{D_d, B_d^*\}_{d=1}^{D}, A_k)\}_{k=1}^K$ ($K \gg 1$) for each base image $\mathcal{I}_j$ to encourage the model to learn robust visual reasoning skills transferable across diverse chart appearances, data distributions, contexts and domains, as well as task complexities.

\textbf{Generalize To Real-World Charts \& Domains.}
Bridging visual perception and logical reasoning, we construct \oursdata to underline the significance of enhancing MLLMs' understanding of fundamental visual components and spatial features for accurate visual reasoning (\S\ref{subsec:importance_of_foundational_learning}).
\ours finetuned on \oursdata are also applicable to real-world chart understanding and out-of-domain benchmarks (\S\ref{subsec:evaluation_data}), validating not only the effectiveness but also the adaptability and generalizability of our approach (\S\ref{appendix:subsec:preliminary_on_method_motivation} \& \ref{appendix:subsec:meta-learning}).

\section{Experiments}
\label{sec:experiments}

\subsection{Setup}
\label{subsec:exp_setup}
% SFT for Chart Reasoning With Grounding

\textbf{Baseline.}
We use two \textit{close-source} MLLMs, GPT-4o~\citep{openai2024gpt4o} and GPT-4.1-mini~\citep{openai2025-gpt-4.1-mini}, and
seven \textit{open-source} MLLMs, Llama-3.2-Vision~\citep{llama3.2-11b-vision-instruct}, Gemma-3~\citep{gemma3-4b-it}, InternVL3~\citep{internvl3_2025}, and Qwen2.5-VL~\citep{qwen2.5_report} with different sizes, as comparison baselines.

\textbf{Model.}
\label{subsec: 4 generation modes}
We employ five MLLMs as base models, finetuned through \ours for enhanced visual-grounded chart reasoning: Qwen2.5-VL (3B and 7B) and InternVL-3 (1B, 2B, and 8B).

\textbf{Data.}
We split \oursdata into non-overlapped \textit{training} and \textit{test} sets. All models are evaluated on the \textit{test} sets unseen for finetuned models.
Implementation details can be found in \S\ref{appendix:subsec:implementation_details}.

% Main Results
% \cellcolor[RGB]{255,230,207}
% \cellcolor[RGB]{226,211,250}
\definecolor{gpt_mllms}{HTML}{DACFF4} 
\definecolor{baseline}{HTML}{DACFF4} 
\definecolor{finetuned}{HTML}{D4E6D9}

\definecolor{gain}{HTML}{389224}
\definecolor{drop}{HTML}{DC5252}

% Macro: \gaincell{baseline}{ours}{value}
% Computes delta = ours - baseline, maps to alpha, colors the cell
\newcommand{\gaincell}[2]{%
  \pgfmathsetmacro{\mydelta}{#2 - #1}%
  \pgfmathsetmacro{\myalpha}{min(max(abs(\mydelta) / 50 * 100, 1), 100)}%
  \pgfmathparse{\mydelta < 0}%
  \ifnum\pgfmathresult=1
    \edef\myfullcolor{drop!\myalpha}%
  \else
    \edef\myfullcolor{gain!\myalpha}%
  \fi
  \expandafter\cellcolor\expandafter{\myfullcolor}#2%
}

\newcommand{\gaincellbf}[2]{%
  \pgfmathsetmacro{\mydelta}{#2 - #1}%
  \pgfmathsetmacro{\myalpha}{min(max(abs(\mydelta) / 50 * 100, 1), 100)}%
  \pgfmathparse{\mydelta < 0}%
  \ifnum\pgfmathresult=1
    \edef\myfullcolor{drop!\myalpha}%
  \else
    \edef\myfullcolor{gain!\myalpha}%
  \fi
  \expandafter\cellcolor\expandafter{\myfullcolor}\textbf{#2}%
}

\newcommand{\gaincellul}[2]{%
  \pgfmathsetmacro{\mydelta}{#2 - #1}%
  \pgfmathsetmacro{\myalpha}{min(max(abs(\mydelta) / 50 * 100, 1), 100)}%
  \pgfmathparse{\mydelta < 0}%
  \ifnum\pgfmathresult=1
    \edef\myfullcolor{drop!\myalpha}%
  \else
    \edef\myfullcolor{gain!\myalpha}%
  \fi
  \expandafter\cellcolor\expandafter{\myfullcolor}\underline{#2}%
}

\begin{table}[t!]

\vspace{-16pt}

\centering
\resizebox{\textwidth}{!}{%
\begin{tabular}{c c | c c c c c | c c c c c | c c c c c}
\toprule
\multirow{2}{*}{\textbf{Model}} & \multirow{2}{*}{\textbf{Size}} & \multicolumn{5}{c|}{\textbf{Level 1}} & \multicolumn{5}{c|}{\textbf{Level 2}} & \multicolumn{5}{c}{\textbf{Level 3}} \\
\cmidrule(lr){3-7} \cmidrule(lr){8-12} \cmidrule(lr){13-17}
& & $\boldsymbol{\textsc{@}M}$ & $\boldsymbol{\textsc{@}0.0}$ & $\boldsymbol{\textsc{@}0.05}$ & $\boldsymbol{\textsc{@}0.1}$ & $\boldsymbol{\textsc{@}0.2}$ & $\boldsymbol{\textsc{@}M}$ & $\boldsymbol{\textsc{@}0.0}$ & $\boldsymbol{\textsc{@}0.05}$ & $\boldsymbol{\textsc{@}0.1}$ & $\boldsymbol{\textsc{@}0.2}$ & $\boldsymbol{\textsc{@}M}$ & $\boldsymbol{\textsc{@}0.0}$ & $\boldsymbol{\textsc{@}0.05}$ & $\boldsymbol{\textsc{@}0.1}$ & $\boldsymbol{\textsc{@}0.2}$ \\
\midrule
\rowcolor{gpt_mllms}
\multicolumn{17}{c}{\textbf{Close-Source MLLMs}} \\
\midrule
GPT-4o & - & 57.64 & 54.07 & 62.00 & 65.36 & 70.50 & 34.04 & 33.75 & 44.93 & 50.54 & 57.96 & 22.14 & 22.29 & 30.25 & 34.04 & 39.14 \\
GPT-4.1-mini & - & \underline{70.86} & \underline{67.43} & \underline{76.29} & \underline{78.79} & \underline{79.93} & \underline{37.61} & \underline{36.54} & \underline{47.18} & \underline{52.93} & \underline{60.86} & \underline{26.14} & \underline{25.46} & \underline{33.11} & \underline{37.79} & \underline{42.32} \\
\midrule
\rowcolor{baseline}
\multicolumn{17}{c}{\textbf{Open-Source Baselines}} \\
\midrule
Gemma-3 & 4B & 38.21 & 28.64 & 32.79 & 37.00 & 41.29 & 18.07 & 12.86 & 18.82 & 23.64 & 28.29 & 11.43 & 9.64 & 13.14 & 15.54 & 18.96 \\
\addlinespace
Llama-3.2-V & 11B & 44.86 & 38.29 & 40.86 & 42.86 & 47.07 & 23.43 & 18.25 & 22.46 & 25.29 & 29.39 & 16.57 & 14.25 & 16.54 & 18.29 & 20.29 \\
\addlinespace

\multirow{3}{*}{InternVL3} & 1B & 20.54 & 16.38 & 20.53 & 24.40 & 29.68 & 8.51 & 7.44 & 12.61 & 15.37 & 20.65 & 6.76 & 5.87 & 8.31 & 10.09 & 10.99 \\
& 2B & 33.53 & 32.52 & 37.52 & 42.52 & 47.29 & 13.11 & 13.22 & 19.89 & 27.14 & 35.55 & 10.69 & 11.18 & 14.71 & 18.72 & 24.90 \\
& 8B & 46.79 & 44.29 & 52.14 & 57.71 & 63.00 & 25.75 & 25.57 & 35.36 & 41.64 & 50.79 & 17.84 & 17.48 & 24.57 & 28.93 & 34.68 \\
\addlinespace
\multirow{2}{*}{Qwen2.5-VL} & 3B & 45.25 & 43.52 & 51.54 & 56.25 & 61.54 & 22.75 & 22.86 & 31.14 & 37.21 & 45.36 & 16.18 & 16.00 & 21.89 & 25.82 & 31.71 \\
& 7B & \underline{54.21} & \underline{50.79} & \underline{60.43} & \underline{64.64} & \underline{69.29} & \underline{28.68} & \underline{28.82} & \underline{39.93} & \underline{45.54} & \underline{52.61} & \underline{19.01} & \underline{19.38} & \underline{26.71} & \underline{31.50} & \underline{36.93} \\

\midrule

\rowcolor{finetuned}
\multicolumn{17}{c}{\textbf{Ours (Stage II)}} \\

\midrule

\multirow{3}{*}{\shortstack{\textbf{Applied}\\(InternVL3)}} & 1B & \gaincell{20.54}{25.79} & \gaincell{16.38}{21.36} & \gaincell{20.53}{25.29} & \gaincell{24.40}{31.43} & \gaincell{29.68}{37.86} & \gaincell{8.51}{10.57} & \gaincell{7.44}{9.25} & \gaincell{12.61}{12.89} & \gaincell{15.37}{17.18} & \gaincell{20.65}{22.21} & \gaincell{6.76}{7.11} & \gaincell{5.87}{7.36} & \gaincell{8.31}{9.25} & \gaincell{10.09}{11.00} & \gaincell{10.99}{12.93} \\
& 2B & \gaincell{33.53}{42.64} & \gaincell{32.52}{41.79} & \gaincell{37.52}{48.71} & \gaincell{42.52}{55.36} & \gaincell{47.29}{62.14} & \gaincell{13.11}{18.68} & \gaincell{13.22}{19.29} & \gaincell{19.89}{25.79} & \gaincell{27.14}{32.93} & \gaincell{35.55}{41.50} & \gaincell{10.69}{11.39} & \gaincell{11.18}{12.57} & \gaincell{14.71}{16.61} & \gaincell{18.72}{20.75} & \gaincell{24.90}{26.32} \\
& 8B & \gaincell{46.79}{58.86} & \gaincell{44.29}{54.64} & \gaincell{52.14}{66.29} & \gaincell{57.71}{69.14} & \gaincell{63.00}{71.79} & \gaincell{25.75}{34.47} & \gaincell{25.57}{33.90} & \gaincell{35.36}{49.67} & \gaincellul{41.64}{56.47} & \gaincellul{50.79}{63.91} & \gaincell{17.84}{18.87} & \gaincell{17.48}{19.84} & \gaincell{24.57}{27.99} & \gaincell{28.93}{31.97} & \gaincell{34.68}{37.41} \\

\addlinespace

\multirow{2}{*}{\shortstack{\textbf{Applied}\\(Qwen2.5-VL)}} & 3B & \gaincell{45.25}{54.21} & \gaincell{43.52}{51.21} & \gaincell{51.54}{59.50} & \gaincell{56.25}{65.00} & \gaincell{61.54}{68.50} & \gaincell{22.75}{25.86} & \gaincell{22.86}{27.14} & \gaincell{31.14}{39.18} & \gaincell{37.21}{47.25} & \gaincell{45.36}{55.18} & \gaincell{16.18}{16.86} & \gaincell{16.00}{17.32} & \gaincell{21.89}{23.93} & \gaincell{25.82}{28.61} & \gaincell{31.71}{33.11} \\
& 7B & \gaincellul{54.21}{65.79} & \gaincellul{50.79}{59.14} & \gaincellul{60.43}{71.93} & \gaincellul{64.64}{75.29} & \gaincellul{69.29}{78.29} & \gaincellul{28.68}{36.82} & \gaincellul{28.82}{34.79} & \gaincellul{39.93}{50.82} & \gaincell{45.54}{56.18} & \gaincell{52.61}{62.75} & \gaincellul{19.01}{21.04} & \gaincellul{19.38}{21.11} & \gaincellul{26.71}{30.21} & \gaincellul{31.50}{34.25} & \gaincellul{36.93}{39.39} \\
\addlinespace
\multirow{2}{*}{\shortstack{\textbf{Boxed}\\(Qwen2.5-VL)}} & 3B & \gaincell{45.25}{58.07} & \gaincell{43.52}{51.79} & \gaincell{51.54}{61.00} & \gaincell{56.25}{65.71} & \gaincell{61.54}{69.86} & \gaincell{22.75}{25.96} & \gaincell{22.86}{25.32} & \gaincell{31.14}{37.75} & \gaincell{37.21}{45.89} & \gaincell{45.36}{53.86} & \gaincell{16.18}{16.92} & \gaincell{16.00}{16.82} & \gaincell{21.89}{22.89} & \gaincell{25.82}{27.21} & \gaincell{31.71}{33.29} \\
& 7B & \gaincell{54.21}{59.79} & \gaincell{50.79}{52.79} & \gaincellul{60.43}{71.93} & \gaincell{64.64}{74.57} & \gaincell{69.29}{76.64} & \gaincell{28.68}{33.79} & \gaincell{28.82}{30.32} & \gaincell{39.93}{49.75} & \gaincell{45.54}{55.89} & \gaincell{52.61}{62.21} & \gaincell{19.01}{20.14} & \gaincell{19.38}{18.04} & \gaincell{26.71}{28.89} & \gaincell{31.50}{32.32} & \gaincell{36.93}{36.29} \\
\addlinespace
\multirow{2}{*}{\shortstack{\textbf{Cropped}\\(Qwen2.5-VL)}} & 3B & \gaincell{45.25}{49.71} & \gaincell{43.52}{46.79} & \gaincell{51.54}{54.86} & \gaincell{56.25}{60.07} & \gaincell{61.54}{65.21} & \gaincell{22.75}{20.68} & \gaincell{22.86}{21.11} & \gaincell{31.14}{32.61} & \gaincell{37.21}{40.36} & \gaincell{45.36}{50.82} & \gaincell{16.18}{15.00} & \gaincell{16.00}{18.89} & \gaincell{21.89}{20.96} & \gaincell{25.82}{24.71} & \gaincell{31.71}{30.25} \\
& 7B & \gaincell{54.21}{58.93} & \gaincell{50.79}{56.57} & \gaincell{60.43}{67.71} & \gaincell{64.64}{72.93} & \gaincell{69.29}{76.71} & \gaincell{28.68}{30.04} & \gaincell{28.82}{29.71} & \gaincell{39.93}{40.43} & \gaincell{45.54}{46.71} & \gaincell{52.61}{53.79} & \gaincell{19.01}{18.68} & \gaincell{19.38}{18.71} & \gaincell{26.71}{26.79} & \gaincell{31.50}{30.36} & \gaincell{36.93}{35.11} \\

\midrule

\rowcolor{finetuned}
\multicolumn{17}{c}{\textbf{Ours (Stage I + II)}} \\

\midrule

\multirow{3}{*}{\shortstack{\textbf{Applied}\\(InternVL3)}} & 1B & \gaincell{20.54}{28.64} & \gaincell{16.38}{26.79} & \gaincell{20.53}{31.57} & \gaincell{24.40}{36.36} & \gaincell{29.68}{43.79} & \gaincell{8.51}{14.18} & \gaincell{7.44}{12.86} & \gaincell{12.61}{16.25} & \gaincell{15.37}{22.86} & \gaincell{20.65}{27.43} & \gaincell{6.76}{10.39} & \gaincell{5.87}{10.64} & \gaincell{8.31}{13.46} & \gaincell{10.09}{15.79} & \gaincell{10.99}{17.75} \\

& 2B & \gaincell{33.53}{47.86} & \gaincell{32.52}{46.93} & \gaincell{37.52}{54.71} & \gaincell{42.52}{60.57} & \gaincell{47.29}{67.79} & \gaincell{13.11}{23.36} & \gaincell{13.22}{23.61} & \gaincell{19.89}{28.79} & \gaincell{27.14}{36.57} & \gaincell{35.55}{43.36} & \gaincell{10.69}{14.89} & \gaincell{11.18}{15.82} & \gaincell{14.71}{20.29} & \gaincell{18.72}{24.79} & \gaincell{24.90}{31.39} \\

& 8B & \gaincell{46.79}{67.71} & \gaincell{44.29}{62.57} & \gaincell{52.14}{70.57} & \gaincell{57.71}{75.14} & \gaincell{63.00}{77.29} & \gaincell{25.75}{39.68} & \gaincell{25.57}{37.79} & \gaincell{35.36}{52.54} & \gaincell{41.64}{58.71} & \gaincell{50.79}{65.04} & \gaincell{17.84}{24.57} & \gaincell{17.48}{22.96} & \gaincell{24.57}{30.54} & \gaincell{28.93}{34.79} & \gaincell{34.68}{40.32} \\

\addlinespace

\multirow{2}{*}{\shortstack{\textbf{Applied}\\(Qwen2.5-VL)}} & 3B & \gaincell{45.25}{60.64} & \gaincell{43.52}{57.21} & \gaincell{51.54}{63.71} & \gaincell{56.25}{69.57} & \gaincell{61.54}{73.79} & \gaincell{22.75}{30.11} & \gaincell{22.86}{29.61} & \gaincell{31.14}{41.82} & \gaincell{37.21}{48.86} & \gaincell{45.36}{57.64} & \gaincell{16.18}{17.96} & \gaincell{16.00}{18.21} & \gaincell{21.89}{24.25} & \gaincell{25.82}{29.07} & \gaincell{31.71}{34.50} \\

& 7B & \gaincellbf{54.21}{69.86} & \gaincellbf{50.79}{66.79} & \gaincellbf{60.43}{75.29} & \gaincellbf{64.64}{78.50} & \gaincellbf{69.29}{80.43} & \gaincellbf{28.68}{40.21} & \gaincellbf{28.82}{38.64} & \gaincellbf{39.93}{54.00} & \gaincellbf{45.54}{60.04} & \gaincellbf{52.61}{66.89} & \gaincellbf{19.01}{26.11} & \gaincellbf{19.38}{24.25} & \gaincellbf{26.71}{32.21} & \gaincellbf{31.50}{36.96} & \gaincellbf{36.93}{42.54} \\

\addlinespace

\multirow{2}{*}{\shortstack{\textbf{Boxed}\\(Qwen2.5-VL)}} & 3B & \gaincell{45.25}{58.50} & \gaincell{43.52}{54.71} & \gaincell{51.54}{64.21} & \gaincell{56.25}{72.57} & \gaincell{61.54}{74.79} & \gaincell{22.75}{27.64} & \gaincell{22.86}{26.86} & \gaincell{31.14}{38.75} & \gaincell{37.21}{46.86} & \gaincell{45.36}{55.36} & \gaincell{16.18}{17.43} & \gaincell{16.00}{17.54} & \gaincell{21.89}{23.50} & \gaincell{25.82}{27.96} & \gaincell{31.71}{33.46} \\
& 7B & \gaincell{54.21}{63.29} & \gaincell{50.79}{59.57} & \gaincell{60.43}{72.14} & \gaincell{64.64}{75.86} & \gaincell{69.29}{77.00} & \gaincell{28.68}{35.32} & \gaincell{28.82}{33.57} & \gaincell{39.93}{51.54} & \gaincell{45.54}{57.54} & \gaincell{52.61}{64.36} & \gaincell{19.01}{23.57} & \gaincell{19.38}{22.14} & \gaincell{26.71}{31.96} & \gaincell{31.50}{35.04} & \gaincell{36.93}{38.25} \\

\addlinespace

\multirow{2}{*}{\shortstack{\textbf{Cropped}\\(Qwen2.5-VL)}} & 3B & \gaincell{45.25}{53.07} & \gaincell{43.52}{50.79} & \gaincell{51.54}{57.71} & \gaincell{56.25}{63.57} & \gaincell{61.54}{69.50} & \gaincell{22.75}{23.29} & \gaincell{22.86}{23.54} & \gaincell{31.14}{34.29} & \gaincell{37.21}{41.43} & \gaincell{45.36}{51.54} & \gaincell{16.18}{17.14} & \gaincell{16.00}{18.04} & \gaincell{21.89}{22.46} & \gaincell{25.82}{26.11} & \gaincell{31.71}{33.04} \\
& 7B & \gaincell{54.21}{62.14} & \gaincell{50.79}{60.71} & \gaincell{60.43}{69.57} & \gaincell{64.64}{74.71} & \gaincell{69.29}{77.86} & \gaincell{28.68}{32.25} & \gaincell{28.82}{31.82} & \gaincell{39.93}{46.32} & \gaincell{45.54}{50.57} & \gaincell{52.61}{57.36} & \gaincell{19.01}{20.11} & \gaincell{19.38}{20.36} & \gaincell{26.71}{28.46} & \gaincell{31.50}{33.86} & \gaincell{36.93}{37.11} \\

% \addlinespace

\bottomrule
\end{tabular}%
}

\vspace{0pt}

\caption{\textbf{Performance On \oursdata.} We compare performance across different curriculum levels using five accuracy metrics $\boldsymbol{acc\textsc{@}}X$ where $X$ is MLLM ($\boldsymbol{\textsc{@}}M$) or ranges (\Sref{subsec:eval_evaluation_metrics}).}
\label{tab:main_result_on_ours_data}

\vspace{0pt}

\end{table}

\subsection{Evaluation}
\label{subsec:experiment-evaluation}

\textbf{Evaluation Data.}
\label{subsec:evaluation_data}
We evaluate on \oursdata (\S\ref{sec:dataset_construction}), CQA, and out-of-domain benchmarks:

\begin{itemize}[topsep=-2pt, itemsep=1pt, parsep=1pt]
    \item \textbf{\oursdata:} Highlighting the significance of learning chart basics through increased task difficulty, we evaluate models on three \textit{test} sets of \oursdata covering three increased curriculum levels, respectively.
    \item \textbf{Chart Benchmarks:} To assess the adaptability of MLLMs to chart understanding benchmarks, we extend our evaluation to \textit{ChartMuseum} (\textit{visual subset})~\citep{chartmuseum_2025}, \textit{CharXiv} (\textit{reasoning subset})~\citep{charxiv_2024}, \textit{ChartQA}~\citep{chartqa_2022}, and \textit{ChartQAPro}~\citep{chartqapro-2025}.
    \item \textbf{Out-of-Domain Benchmarks:} Generalizing to different domains of multimodal reasoning, we extend our evaluation to multi-discipline multimodal reasoning tasks, including \textit{MathVista}~\citep{mathvista-2023} and \textit{MMMU-Pro}~\citep{mmmu-pro-2025}.
\end{itemize}

\textbf{Evaluation Metrics.}
\label{subsec:eval_evaluation_metrics}
We evaluate different aspects of responses through complementary metrics that take both textual outputs and visual grounding into consideration (\S\ref{appendix:sec:eval_metrics}):

\textbf{(1) \textit{Reasoning}:}
For reasoning evaluation, we employ two complementary approaches:

% Performance on chart & out-of-domain benchmarks

\begin{wraptable}{r}{0.75\textwidth}

\vspace{3pt}

\centering
\resizebox{0.75\textwidth}{!}{%
\begin{tabular}{c c | c c c c | c c}
\toprule
\multirow{2}{*}{\textbf{Model}} & \multirow{2}{*}{\textbf{Size}} & \multicolumn{4}{c|}{\textbf{Chart Benchmarks}} & \multicolumn{2}{c}{\textbf{Out-of-Domain}} \\
\cmidrule(lr){3-6} \cmidrule(lr){7-8}
& & \textbf{ChartQA} & \textbf{ChartQA-Pro} & \textbf{CharXiv} & \textbf{ChartMuseum} & \textbf{MathVista} & \textbf{MMMU-Pro} \\
\midrule
\rowcolor{baseline}
\multicolumn{8}{c}{\textbf{Baselines}} \\
\midrule

\multirow{3}{*}{InternVL3} & 1B & 41.68 & 11.48 & 15.70 & 10.01 & 35.80 & 9.94 \\
& 2B & 66.96 & 20.84 & 24.30 & 15.02 & 55.90 & 16.36 \\
& 8B & \underline{74.56} & \underline{30.56} & \underline{36.20} & \underline{24.42} & \underline{68.80} & 26.99 \\
\addlinespace
\multirow{2}{*}{Qwen2.5-VL} & 3B & 62.32 & 17.02 & 19.50 & 12.21 & 56.10 & 21.45 \\
& 7B & 72.48 & 29.77 & 32.50 & 21.62 & 64.60 & \underline{28.21} \\

\midrule

\rowcolor{finetuned}
\multicolumn{8}{c}{\textbf{Ours (Stage I + II)}} \\

\midrule

\multirow{3}{*}{\shortstack{\textbf{Applied}\\(InternVL3)}} & 1B & \gaincell{41.68}{51.56} & \gaincell{11.48}{15.86} & \gaincell{15.70}{18.90} & \gaincell{10.01}{11.21} & \gaincell{35.80}{45.93} & \gaincell{9.94}{12.95} \\
& 2B & \gaincell{66.96}{69.60} & \gaincell{20.84}{23.97} & \gaincell{24.30}{28.10} & \gaincell{15.02}{16.12} & \gaincell{55.90}{62.78} & \gaincell{16.36}{18.27} \\
& 8B & \gaincellbf{74.56}{78.28} & \gaincell{30.56}{32.39} & \gaincellbf{36.20}{40.20} & \gaincellbf{24.42}{26.23} & \gaincellbf{68.80}{75.37} & \gaincell{26.99}{30.81} \\
\addlinespace
\multirow{2}{*}{\shortstack{\textbf{Applied}\\(Qwen2.5-VL)}} & 3B & \gaincell{62.32}{72.36} & \gaincell{17.02}{24.74} & \gaincell{19.50}{31.80} & \gaincell{12.21}{19.32} & \gaincell{56.10}{66.30} & \gaincell{21.45}{22.77} \\
& 7B & \gaincell{72.48}{75.04} & \gaincellbf{29.77}{32.80} & \gaincell{32.50}{36.70} & \gaincell{21.62}{25.12} & \gaincell{64.60}{72.41} & \gaincellbf{28.21}{32.08} \\

\bottomrule

\end{tabular}%
}

\vspace{-3pt}

\caption{\textbf{Performance on Chart and Out-of-Domain Benchmarks.} MLLM-as-judge $\boldsymbol{acc\textsc{@}MLLM}$ (\Sref{subsec:eval_evaluation_metrics}) on baselines and \ours.}
\label{tab:main_results_out-of-domain}

\vspace{6pt}

\end{wraptable}

\vspace{-0.0em}
\textbf{Micro Evaluation ($\boldsymbol{acc\textsc{@}mic}$):} We evaluate reasoning via a combination of five \textit{micro} metrics: \textsc{ROUGE-L} (\Eref{eq:rouge}), \textsc{BLEU} (\Eref{eq:bleu}), \textsc{METEOR} (\Eref{eq:meteor}), \textsc{BERTScore} (\Eref{eq:bertscore}), and \textsc{Cosine Similarity} (\Eref{eq:cosine}).

\vspace{-0.0em}
\textbf{Macro Evaluation ($\boldsymbol{acc\textsc{@}mac}$):} We employ GPT-4.1-mini as the judge for \textit{macro}-level evaluation by assigning quality scores based on three evaluation criteria (\S\ref{appendix:subsec:eval_metric:reasoning}).

\textbf{(2) \textit{Visual Grounding}:}
We leverage Intersection-over-Union (IoU) variants \textsc{cIoU} (\Eref{eq:ciou}) and \textsc{gIoU} (\Eref{eq:giou}), where \textsc{gIoU} \citep{GIoU2019} is generalized IoUs and \textsc{cIoU} evaluates the cumulative intersection over the cumulative unions \citep{CIoU-DIoU-2019}.

\textbf{(3) \textit{Answer}:}
We define CQA accuracy as the mean \textit{answer} accuracy across all testing samples.

\vspace{-0.3em}
\textbf{MLLM as Judge ($\boldsymbol{acc\textsc{@}MLLM}$):} We employ GPT-4.1-mini~\citep{openai2025-gpt-4.1-mini} as the judge to evaluate answer accuracy through \textit{True-or-False} assessment ($pass\textsc{@}1$). More details in \S\ref{appendix:subsec:eval_metric:answer}.

\vspace{-0.3em}
\textbf{Rule as Judge ($\boldsymbol{acc\textsc{@}range}$):} To mitigate potential biases introduced by\textit{ MLLM-as-judge}~\citep{Dorner2024LimitsTSA,Li2024CalibraEvalCPA}, we introduce rule-based evaluation metrics (Algorithm~\ref{algorithm:rule_based_answer_eval}) to assess answer accuracy through systematic parsing and rubric judgment.
In particular, it incorporates four $range$ criteria that capture different levels of strictness (\S\ref{appendix:subsec:eval_metric:answer}), including the \textit{absolute} accuracy ($acc\textsc{@}0.0$) and three progressively \textit{relaxed} thresholds ($acc\textsc{@}0.05$, $acc\textsc{@}0.1$, $acc\textsc{@}0.2$).

\subsection{Main Results On Chart \& Multimodal Understanding}
\label{subsec:main_results}

\textbf{Performance on \oursdata.}
Compared with the baselines (\Tref{tab:main_result_on_ours_data}), our finetuned models achieve consistently higher accuracy across all six metrics, with absolute gains of up to 14.85\% and 20.92\% for single-stage and two-stage training, respectively.
The highest performance is achieved by \ours{}@Applied (Qwen2.5-VL-7B), with up to 15.65\% higher than its base model.
Compared with \textsc{GPT} models, it achieves up to 12.22\% higher accuracy, demonstrating the effectiveness of \ours in enhancing MLLMs' visual reasoning abilities.

\textbf{Performance on Complex Chart Understanding.}  
Although trained solely on single-plot charts ($1 \leq D < 3$), finetuned models generalize to multi-plot charts ($D \geq 3$).
As shown in \Tref{tab:main_result_on_ours_data}, \ours{}@Applied (Qwen2.5-VL-7B) achieves up to $\uparrow$7.10\% improvements, and \ours{}@Applied (InternVL-8B) also shows $\uparrow$6.73\% across all metrics.

\textbf{Performance on Chart Benchmarks.}  
Aiming for \ours to be not only adaptable across different task complexities but also generalizable to real-world chart comprehension, we extend our evaluation to CQA benchmarks (~\S\ref{subsec:evaluation_data}). Results in \Tref{tab:main_results_out-of-domain} highlight the strong generalizability of \ours, with improvements $\geq \uparrow$1.20\% across four chart benchmarks.

\textbf{Generalizability to Out-of-Domain Multimodal Reasoning.}
The advantage of \ours remains consistent in out-of-domain multimodal reasoning across diverse multimodal reasoning categories (\S\ref{subsec:experiment-evaluation}),
attaining up to $\uparrow$10.20\% accuracy improvements (\Tref{tab:main_results_out-of-domain}).

\textbf{Generalizability to Different Training Paradigms.}
\ours is readily applicable to other training paradigms (\S\ref{appendix:subsec:rl_vs_sft}) to enhance MLLMs' intrinsic visual reasoning capabilities. As shown in Tab.~\ref{tab:rl_main_result_on_ours_data}, applying \ours to \textit{reinforcement learning} yields up to 12.58\% improvement, while combining SFT and RL further increases gains to 17.04\%. Despite even higher performance with RL, its substantially higher computational overhead (Tab.~\ref{tab:grounding_computation_and_config}) contrasts with the more favorable performance-efficiency trade-off of SFT.

% Ablation on curv method

\begin{wrapfigure}[13]{r}{0.56\textwidth}
    \centering

    \vspace{-3pt}
    \includegraphics[width=0.55\textwidth]{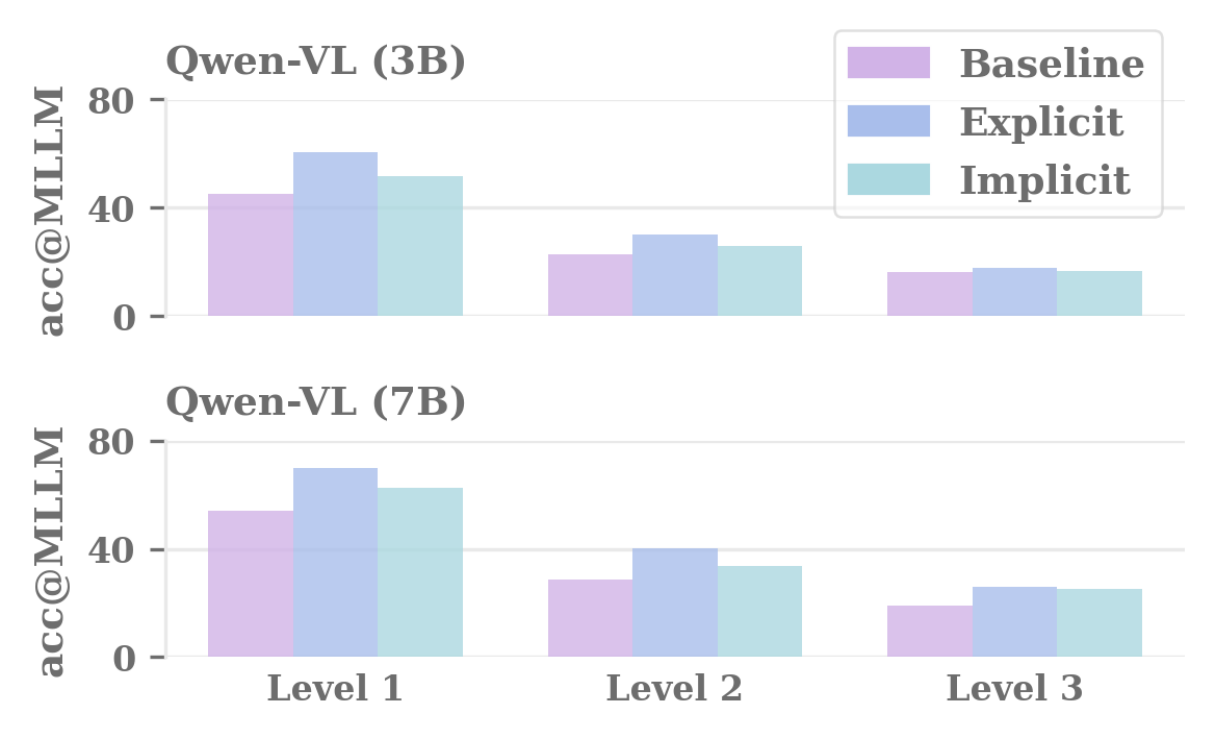}
    \vspace{-10pt}
    \caption{\textbf{Explicit vs. Implicit}}
    \label{fig:ablation_explicit_vs_implicit}
\end{wrapfigure}

\setlength{\intextsep}{0pt}

\subsection{Ablations on Visual Grounded Reasoning}
\label{subsec:ablation_results}

\textbf{Benefits of Explicit over Implicit Visual Grounded Reasoning.} Comparing \ours with \textit{explicit} and \textit{implicit} visual grounded reasoning (\S\ref{appendix:subsec:explicit_vs_implicit}), Fig.~\ref{fig:ablation_explicit_vs_implicit} demonstrates that \textit{explicit} visual grounded reasoning consistently outperforms its \textit{implicit} counterpart ($\uparrow 8.78\%$). This advantage highlights the value of visual grounding as an integral intermediate vision-reasoning bridge, rather than the ultimate learning objective, to effectively develop and enhance MLLMs' intrinsic visual reasoning capabilities.

\textbf{\ours{}$@$\textit{Applied} Presents More Effective Visual Grounding.}
Implementing three visual grounding methods (\S\ref{sec:methodology} \& \S\ref{appendix: 3 visual grounding strategies}), results in \Tref{tab:main_result_on_ours_data} unveils that \textcolor{vision}{\textbf{boxed}} grounding stays less beneficial than directly highlighting regions of focus through \textcolor{vision}{\textbf{applied}} masking, despite its simplicity and straightforwardness.
On the other hand, although restricted by the trade-off between zoom-in resolution and computation overhead (\S\ref{appendix:subsec:computation_overhead}), \textcolor{vision}{\textbf{cropped}} grounding showcases its strengths (\Tref{tab:main_result_on_ours_data}) with up to $\uparrow 7.93\%$ improvements on $acc\textsc{@}MLLM$ and $\uparrow 9.92\%$ gains on $acc\textsc{@}0.0$, despite a sixteen-fold reduction in resolution.

\textbf{Effectiveness of Curriculum Learning.}
Comparing across task difficulty levels (\S\ref{appendix:subsec:ablation_curriculum_learning}), results (\Fref{fig:ablation_curriculum_learning}) show that CL consistently improves performance over standard training, which remains inferior to \ours and exhibits diminished gains as task complexity increases.

% Curriculum Learning Analysis

\begin{wrapfigure}[17]{r}{0.56\textwidth}
    \centering

    \vspace{-0pt}
    \includegraphics[width=0.53\textwidth]{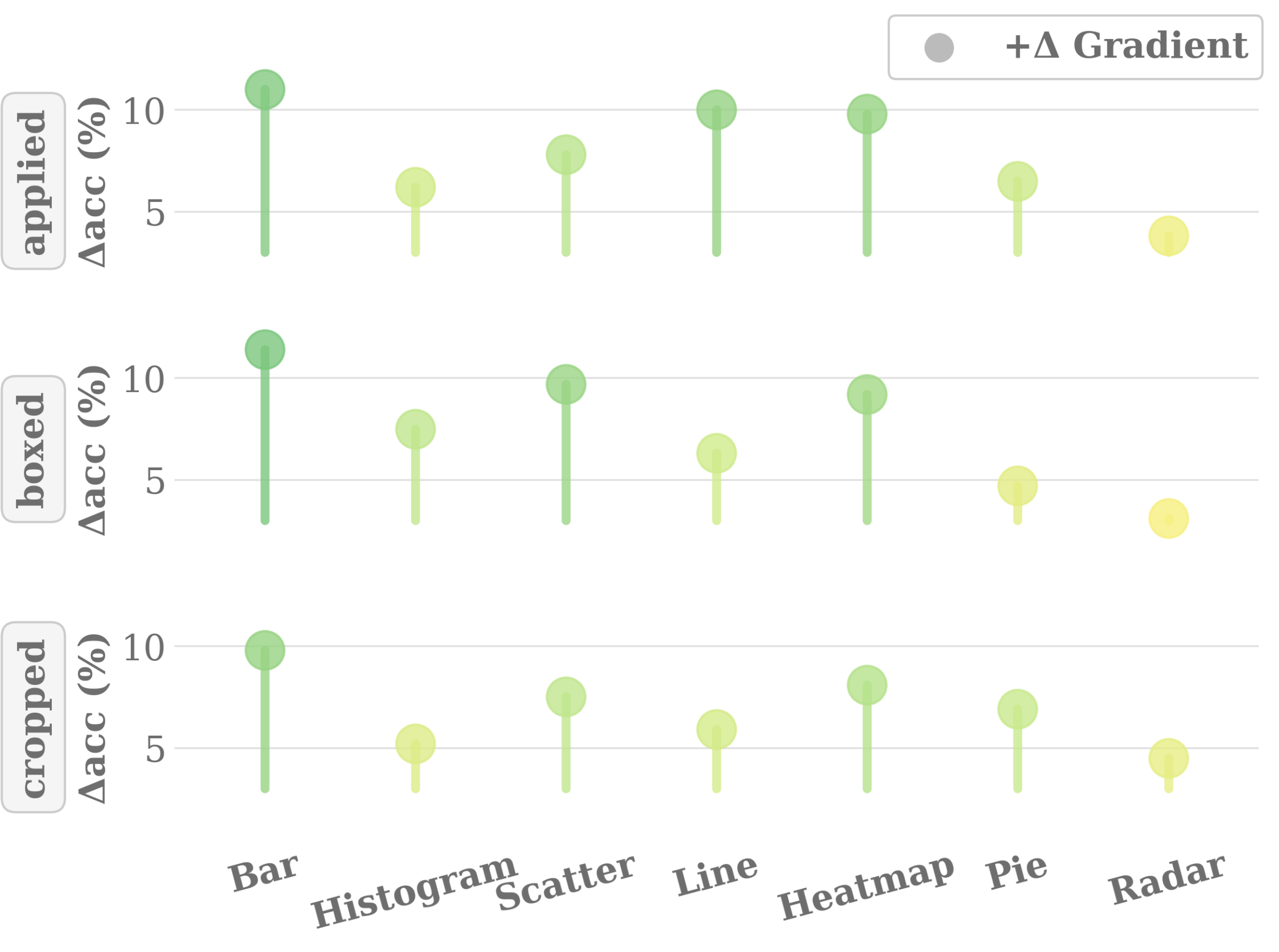}
    \vspace{-12pt}
    
    \caption{\textbf{Ablation on Chart Types.} We compare $\Delta_{acc}$ among chart types.}
    \label{fig:effects_chart_type}
\end{wrapfigure}

\setlength{\intextsep}{0pt}

\textbf{Consistent Gains across Chart Types.}
As shown \Fref{fig:effects_chart_type} (\textit{top}), we compare the accuracy improvements ($\Delta_{acc}$) of \ours{}@\textit{Applied} (Qwen2.5-VL-7B) over its base model across chart types. \textit{Bar} charts benefit the most, followed by \textit{line} plots and \textit{heatmaps}. The remaining chart types also demonstrate positive gains, with \textit{radar} charts contributing the least, reflecting their lower prevalence than other chart types.
Furthermore, the comparison among \ours{}@\textit{Applied}, @\textit{Boxed}, and @\textit{Cropped} (\Fref{fig:effects_chart_type}) shows different distributions across chart types, suggesting different effects across visual grounding strategies (\S\ref{appendix: 3 visual grounding strategies}).

% Curriculum Learning Analysis

\begin{wrapfigure}[14]{r}{0.58\textwidth}

    \vspace{-3pt}

    \small
    \centering
    \includegraphics[width=0.58\textwidth]{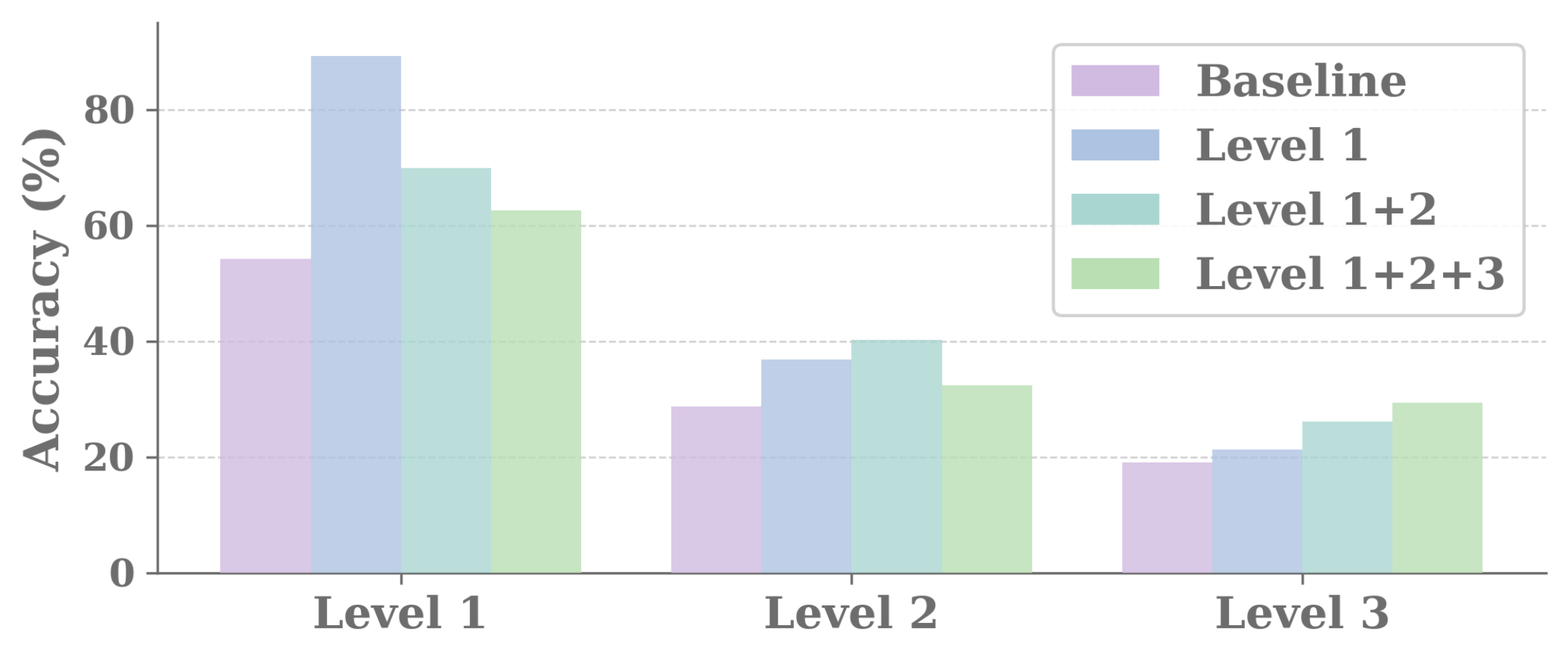}
    \vspace{-20pt}
    \caption{\textbf{Foundational Learning In Chart Understanding.} We compare Qwen2.5-VL-7B with its \textit{applied} model trained on curriculum levels $1$, $1+2$, and $1+2+3$, respectively.}
    \label{fig:complexity_effects}

    \vspace{-18pt}
    
\end{wrapfigure}

\textbf{Foundational Learning Drives More Balanced Gains.}
While all curriculum levels improve overall accuracy, training on \textit{levels} 1+2 yields the most consistent and substantial gains across different levels (\Fref{fig:complexity_effects}). In contrast, relying solely on \textit{level} 3 is less effective, highlighting the significance of foundational learning in establishing adaptive and generalizable visual reasoning capabilities to support more robust chart understanding across varying difficulty levels (\S\ref{subsec:importance_of_foundational_learning}).

\textbf{\ours Enhances Multifaceted Visual Reasoning Abilities.}
\ours notably improves MLLMs' performance, even on the most challenging \textit{level} 3 \textit{multi-chart QA} (Tab.~\ref{tab:main_result_on_ours_data}). Despite the inherent complexity gap between single-chart (\textit{levels} 1-2: \textit{accuracy} up to $80.43\%$) and multi-chart scenarios (\textit{level} 3: \textit{accuracy} $\leq 42.54\%$), \ours narrows this gap by enhancing both \textit{localization} accuracy and \textit{relational} understanding across different chart subplots (\S\ref{subsec:curriculum_construction}).

% Curriculum Level 3 - Analysis

\begin{wrapfigure}[13]{r}{0.58\textwidth}
    \centering
    \small

    \vspace{12pt}
    
    \includegraphics[width=0.58\textwidth]{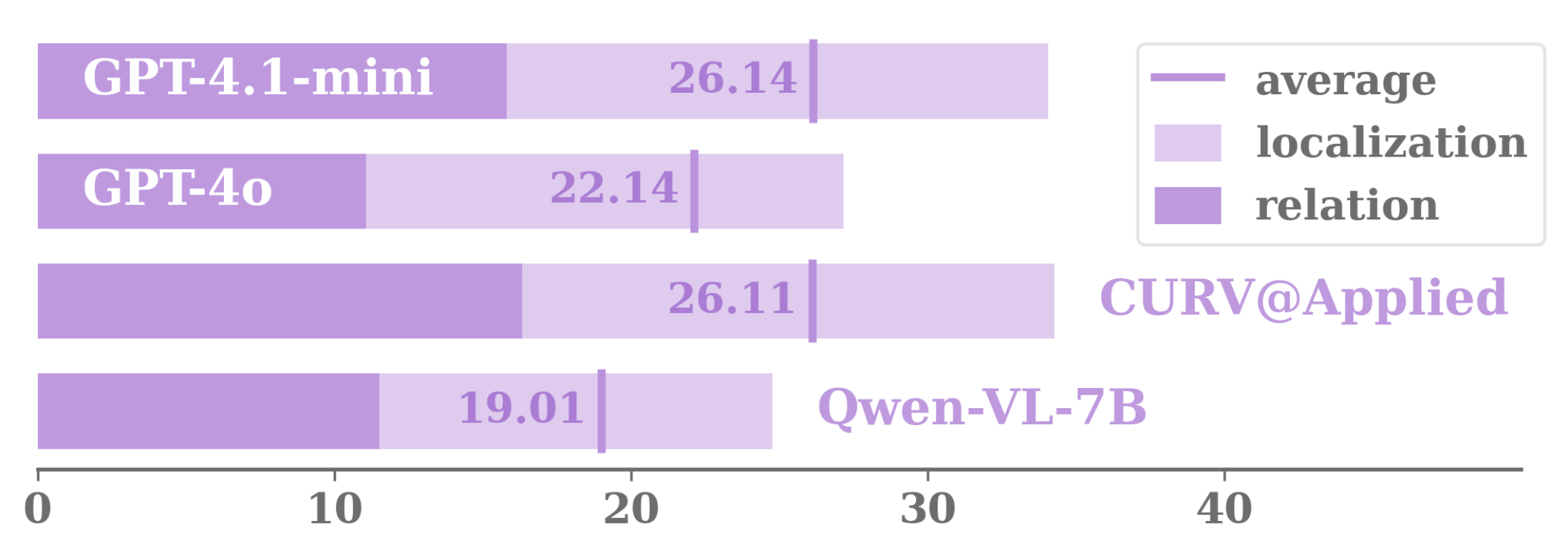}
    \vspace{-18pt}
    \caption{\textbf{Performance on Multi-Chart Understanding.}
    Top-4 accuracy scores on curriculum \textit{level} 3.}
    \label{fig:c3_analysis}
    
\end{wrapfigure}

\setlength{\intextsep}{0pt}

\vspace{-5pt}

The operational breakdown in Fig.~\ref{fig:c3_analysis} shows that \ours leads to clear improvements in \textbf{localization} ($\uparrow 4.68\%$), enabling models to more accurately identify relevant chart subplots within complex layouts. In addition, \ours also effectively improves \textbf{relation} understanding ($\uparrow 2.42\%$), facilitating more reliable cross-chart reasoning over dispersed visual evidence.
Collectively, these gains suggest that \ours strengthens MLLMs' multifaceted visual reasoning abilities to tightly couple visual evidence with logical reasoning. This extends effectively beyond simple CQA settings to more complex multi-chart scenarios and out-of-domain reasoning, indicating that the benefits of \ours arise from improved reasoning grounded in visual structure rather than task-specific adaptation.

\section{Conclusion}

In this work, we present \ours (\S\ref{sec:methodology}), a curriculum learning framework that develops intrinsic visual reasoning capabilities via progressive multi-step visual grounded reasoning.
To support model learning, we construct \oursdata (\S\ref{sec:dataset_construction}) with three progressive curriculum levels. Through systematic experiments (\S\ref{sec:experiments} \& \ref{appendix:sec:in_depth_analysis}), results demonstrate that tightly interleaving reasoning with visual grounding consistently improves performance across curriculum levels and generalizes effectively to chart understanding and out-of-domain multimodal reasoning.
Our work establishes a foundation for developing self-contained visual reasoning capabilities in MLLMs, moving beyond extrinsic assistance toward intrinsic grounded visual reasoning.
% Building upon our model-level enhancement, for future work, we aim to explore agentic chart understanding to better incorporate external knowledge, tools, and collaboration, to complement intrinsic visual reasoning of individual MLLMs.

\clearpage

\section*{Acknowledgments}
The authors acknowledge William \& Mary Research Computing for providing computational resources and/or technical support that have contributed to the results reported within this paper. This work used DeltaAI at NCSA through allocation CIS260012 and CIS230280 from the Advanced Cyberinfrastructure Coordination Ecosystem: Services \& Support (ACCESS) program, which is supported by U.S. National Science Foundation grants \#2138259, \#2138286, \#2138307, \#2137603, and \#2138296. This material is based upon work supported by the Google Cloud Research Credits program with the award GCP19980904.

% \subsection*{Reproducibility Statement}
% We will make the complete source code and curriculum learning datasets public to ensure reproducibility of our work. In this paper, we also elaborate our implementation details, hyperparameter settings, and prompts for LLM-as-judge evaluation to assist the reproduction of our work.

% \section*{Ethics Statement}
% In this work, we introduce a curriculum learning framework with the dataset \oursdata constructed through meta-learning supported CQA creation. Other evaluation benchmarks, including chart understanding and multimodal reasoning, are publicly available and do not contain personally identifiable information or sensitive content. Our methods are designed for research and educational purposes, and we do not foresee direct misuse. With every step being effectively controlled, we positively believe that our work does not violate any ethical standards.

% \clearpage

\bibliography{colm2026_conference}
\bibliographystyle{colm2026_conference}

\clearpage

\appendix

% Table of Contents
\begingroup
\hypersetup{
  linkcolor=mycitecolor
}
\tableofcontents
\endgroup

\clearpage

\section{Preliminary Exploration \& Validation}
\label{appendix:preliminary}

\subsection{Preliminary Exploration On Motivations}
\label{appendix:subsec:preliminary_on_method_motivation}

% Move to Section 3 - Problem Statement
% Preliminary 3 - Error Distribution

\begin{wrapfigure}[19]{r}{0.4\textwidth}
    \vspace{0pt}
    \centering
    \small
    \includegraphics[width=0.4\textwidth]{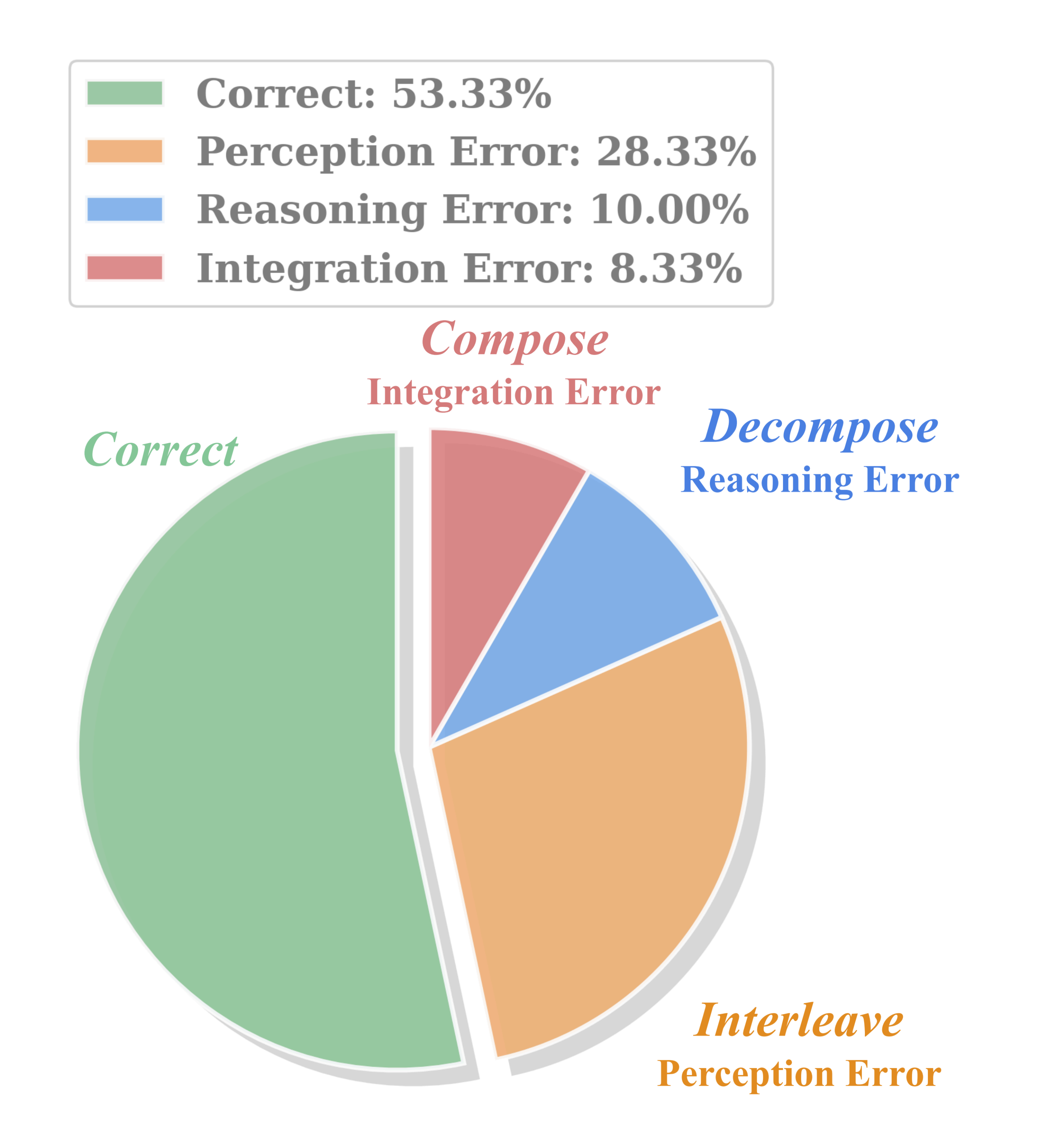}
    
    \vspace{-6pt}
    
    % \caption{\textbf{Error Distribution.} Visualization of error analysis (\Sref{sec:preliminary}).}

    \caption{\textbf{Error Analysis} (\Sref{sec:preliminary})}    
    \label{fig:appendix_preliminary3_error_distribution}
\end{wrapfigure}

Building on the cognitive perspectives that humans solve multimodal problems through \textbf{decomposition}, \textbf{interleaved visual reasoning}, and \textbf{composition} (\Fref{fig:cover}), we conduct preliminary studies to examine where current MLLMs fall short (\S\ref{sec:preliminary}).

Concretely, we evaluate GPT-4o on 60 CharXiv (\citep{charxiv_2024}) samples, categorizing the root causes of CQA failures into three classes: \textit{reasoning errors} (\textbf{decomposition}), \textit{perception errors} (\textbf{interleaved visual reasoning}), and \textit{integration errors} (\textbf{composition}).
As shown in \Fref{fig:appendix_preliminary3_error_distribution}, perception errors ($28.33\%$) emerge as the dominant reason of failures, reflecting the difficulty of accurately grounding reasoning steps in fine-grained chart details.
Beyond perception, models also exhibit weaknesses in decomposition ($10.00\%$), struggling to break down complex problems into coherent chains of reasoning. Additionally, they also show deficiencies in composition ($8.33\%$), failing to integrate grounded visual evidence into a coherent, interleaved chain of thought.
These findings reveal systematic shortcomings in human-inspired reasoning stages, motivating our design of \ours that enhances MLLMs’ intrinsic visual grounded reasoning capabilities by simulating human cognitive process of \textbf{decomposing}, \textbf{interleaving}, and \textbf{composing} toward a solution.

Inspired by recent findings that models can effectively learn from low-level features~\citep{Wang2025VisuallyDescriptive}, we concretize the notion of \textbf{decomposition} in two complementary forms:
(1) \textit{visual decomposition}, where each chart is decomposed into low-level components (\Fref{fig:meta_learning}) to to guide MLLMs’ attention toward fine-grained and informative details; and
(2) \textit{reasoning decomposition}, where each chart understanding problem is decomposed into a structured chain of reasoning steps to help MLLMs enhance their logical reasoning capacities.
Building on these decompositions, we incorporate \textbf{interleaving} insights into our design, enabling reasoning to be interleaved with dynamically shifting visual focuses.
Accordingly, the \textbf{composition} process integrates all intermediate learning in a coherent chain: from \textit{reasoning composition} that consolidates step-wise reasoning into a coherent logical chain, to \textit{visual composition} that progressively aggregates low-level visual interpretations into holistic chart comprehension.

Together, these elements form the foundation of our \textbf{curriculum learning} design (\S\ref{sec:methodology}), which standardizes two dimensions of progression:
(1) \textit{curriculum CQA reasoning difficulty}, controlled by increasing the number of nested functions; and
(2) \textit{curriculum chart visual complexity}, controlled by increasing the number of low-level components, chart types, and chart subplots.
We further support our design with \textbf{meta-learning} (\S\ref{appendix:subsec:meta-learning}) to endow MLLMs with adaptability and generalizability in the face of varying chart types, context domains, and task complexity.

% Moved to section 2: methodology
% \input{figures/fig_meta_learning}

\subsection{Preliminary Exploration on CQA Challenges}
\label{subsec:preliminary_exploration}

\begin{figure}[H]
    \centering
    \includegraphics[width=1.0\textwidth]{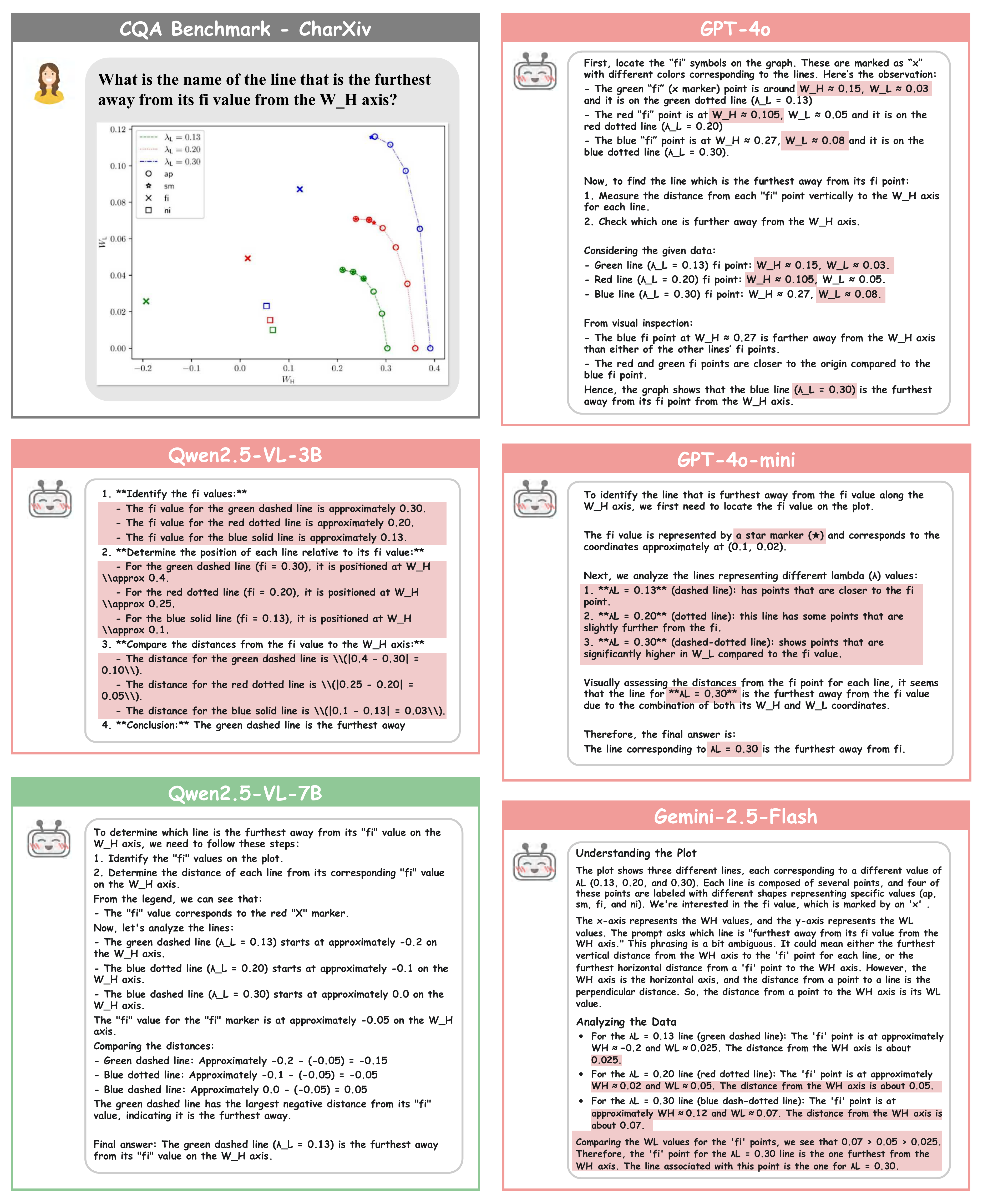}
    \caption{\textbf{Examples of Preliminary Exploration on CharXiv.} We conduct case study on CharXiv samples by prompting the model to generate CoT reasoning and final answer. CoT reasoning prompt is omitted in the input for clarify.}
    \label{fig:appendix_preliminary_evidence_1_charxiv}
\end{figure}

\begin{figure}[H]
    \centering
    \includegraphics[width=0.95\textwidth]{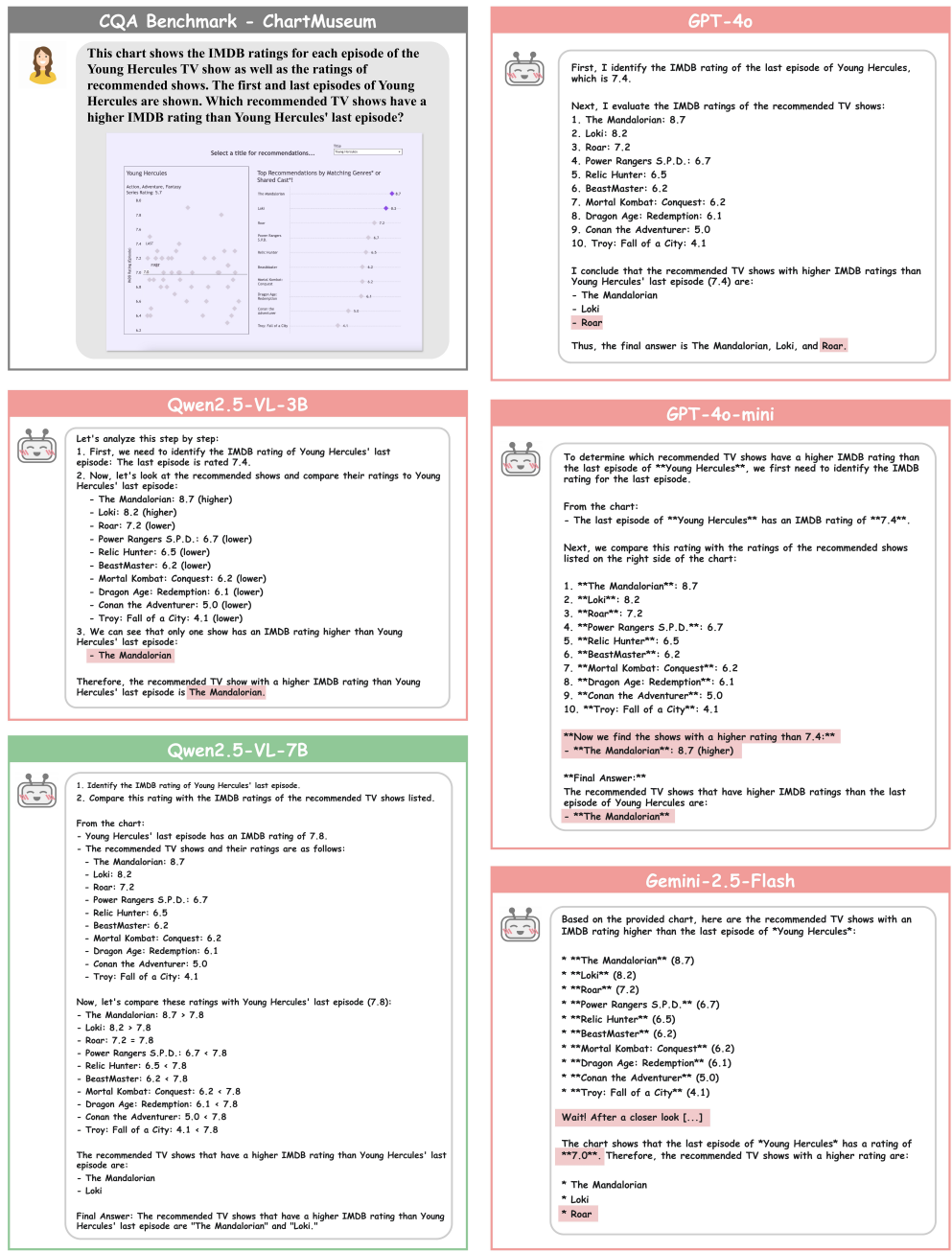}
    \caption{\textbf{Examples of Preliminary Exploration on ChartMuseum.} We conduct case study on ChartMuseum samples by prompting the model to generate CoT reasoning and final answer. CoT reasoning prompt is omitted in the input for clarify.}
    \label{fig:appendix_preliminary_evidence_2_chartmuseum}
\end{figure}

To investigate the underlying causes of failures in chart understanding, we employ five MLLMs, including three \textit{close-source} models (GPT-4.1-mini~\citep{openai2025-gpt-4.1-mini}, GPT-4o~\citep{openai2024gpt4o}, Gemini-2.5-Flask~\citep{google2025gemini2.5flash}) and two \textit{open-source} models (Qwen2.5-VL-3B and Qwen2.5-VL-7B~\citep{qwen2.5_report}) to identify the root causes of their failures.
Specifically, we analyze their CQA outputs case-by-case on different CQA benchmarks (\S\ref{subsec:experiment-evaluation}), noticing several key patterns in MLLMs' CQA failures:

\begin{itemize}
    \item \textbf{Reasoning Accuracy \& Consistency:} While prompting MLLMs to do CoT reasoning can guide them toward correct answers in some cases, we still observe notable visual reasoning failures. For example, in \Fref{fig:appendix_preliminary_evidence_1_charxiv}, Qwen2.5-VL-3B misaligns line colors with their corresponding labels at the beginning, which propagates this misperception through subsequent reasoning and results in an incorrect answer. On the other hand, in \Fref{fig:appendix_preliminary_evidence_2_chartmuseum}, GPT-4o fails to excluded ``\textit{Loki}'' despite having correctly identified it in earlier steps, unveiling the inconsistency in its evolution of reasoning. Another form of reasoning inconsistency emerges in recursive self-correction, where it may occur repeatedly throughout the model’s reasoning process, ultimately producing inconsistent or divergent answers (\textit{e.g.}, Gemini-2.5-Flash in \Fref{fig:appendix_preliminary_evidence_2_chartmuseum}).
    \item \textbf{Visual Grounding Accuracy:} MLLMs exhibit significant challenges in precisely capture visual details from the chart images. For example, in \Fref{fig:appendix_preliminary_evidence_1_charxiv}, GPT-4o inaccurately estimates the $W_H$ value of the red ``fi'' point as approximately $0.105$, while the true value is significantly less than $0.1$, residing just above $0.0$.
    \item \textbf{Visual Reasoning Effectiveness:} In complex reasoning tasks requiring the integration of multiple visual regions and reasoning steps, MLLMs often struggle to effectively link visual attention with logical reasoning. For instance (\Fref{fig:appendix_preliminary_evidence_2_chartmuseum}), although GPT-4o and GPT-4o-mini both perceive accurately in their initial perception, they exhibit distinct failures in subsequent reasoning: GPT-4o incorrectly includes ``\textit{Roar}'' while GPT-4o-mini fails to incorporate ``\textit{Loki}''.
\end{itemize}

\subsection{Preliminary Validation on \oursdata}
\label{subsec:preliminary_validation}

\begin{figure}[H]
    \centering
    \includegraphics[width=1.0\textwidth]{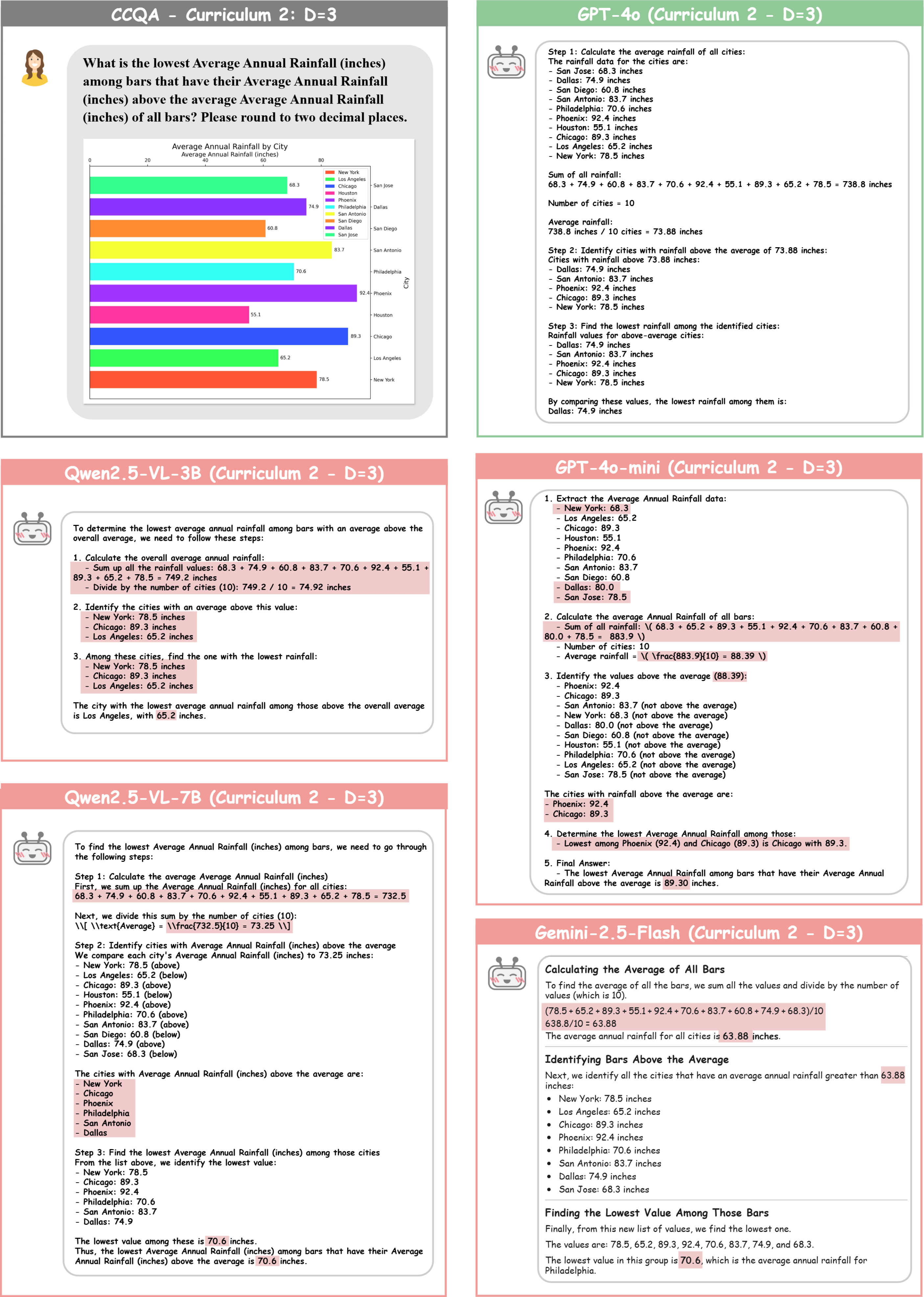}
    \caption{\textbf{Examples of Validation On Curriculum Level 2 ($D \geq 3$).} We validate our \oursdata on \textit{Tier 3} (\Sref{appendix:five fine-grained curriculum tiers}), i.e., \textit{Curriculum Level 2} ($D \geq 3$), by prompting the model to generate CoT reasoning and final answer. CoT reasoning prompt is omitted in the input for clarify.}
    \label{fig:appendix_preliminary_validation3}
\end{figure}

\begin{figure}[H]
    \centering
    \includegraphics[width=1.0\textwidth]{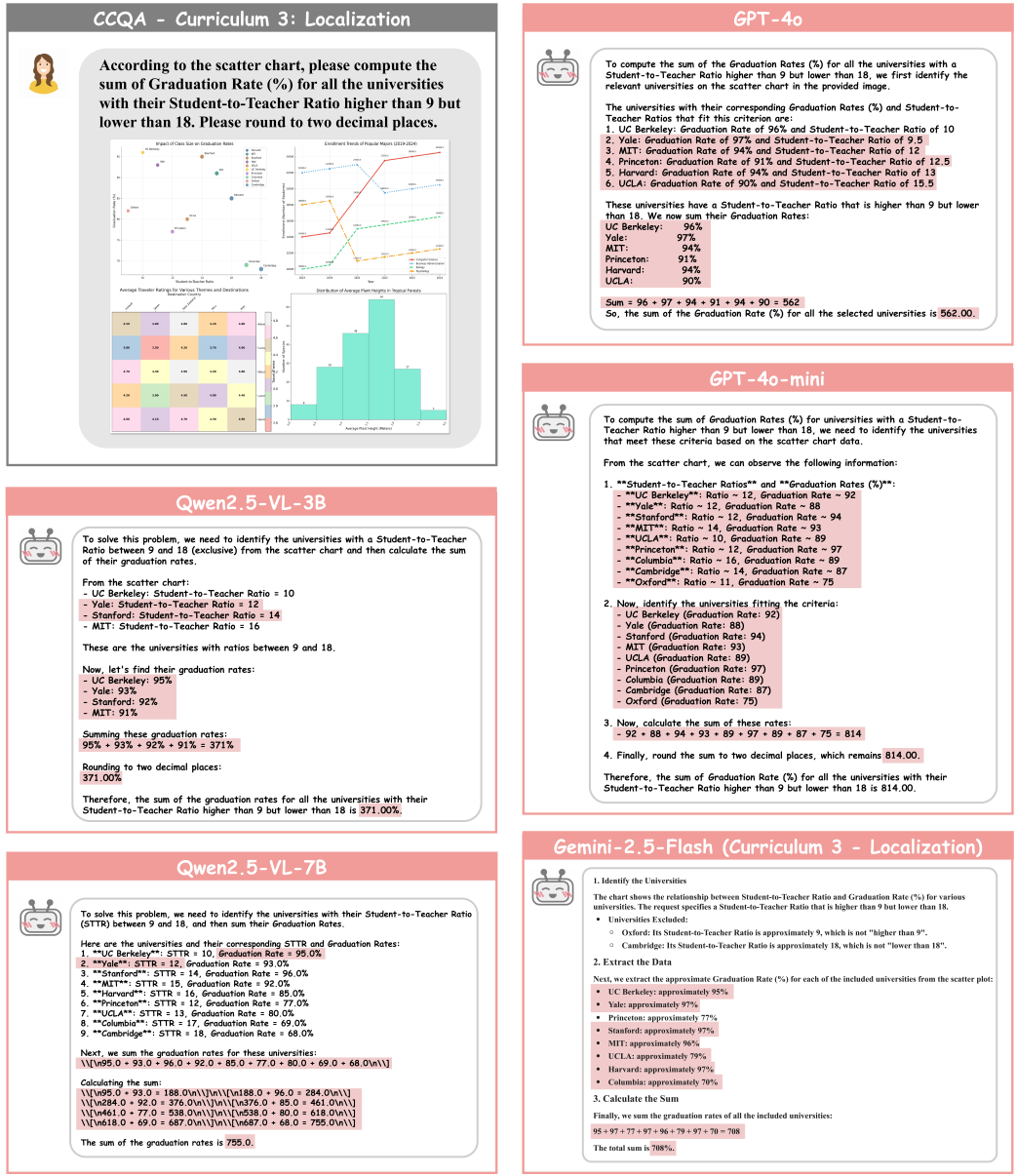}
    \caption{\textbf{Examples of Validation On Curriculum Level 3 ($D \geq 3$).} We validate our \oursdata on \textit{Tier 4} (\Sref{appendix:five fine-grained curriculum tiers}), i.e., \textit{Curriculum Level 3} ($D \geq 3$), by prompting the model to generate CoT reasoning and final answer. CoT reasoning prompt is omitted in the input for clarify.}
    \label{fig:appendix_preliminary_validation4}
\end{figure}

\begin{figure}[H]
    \centering
    \includegraphics[width=1.0\textwidth]{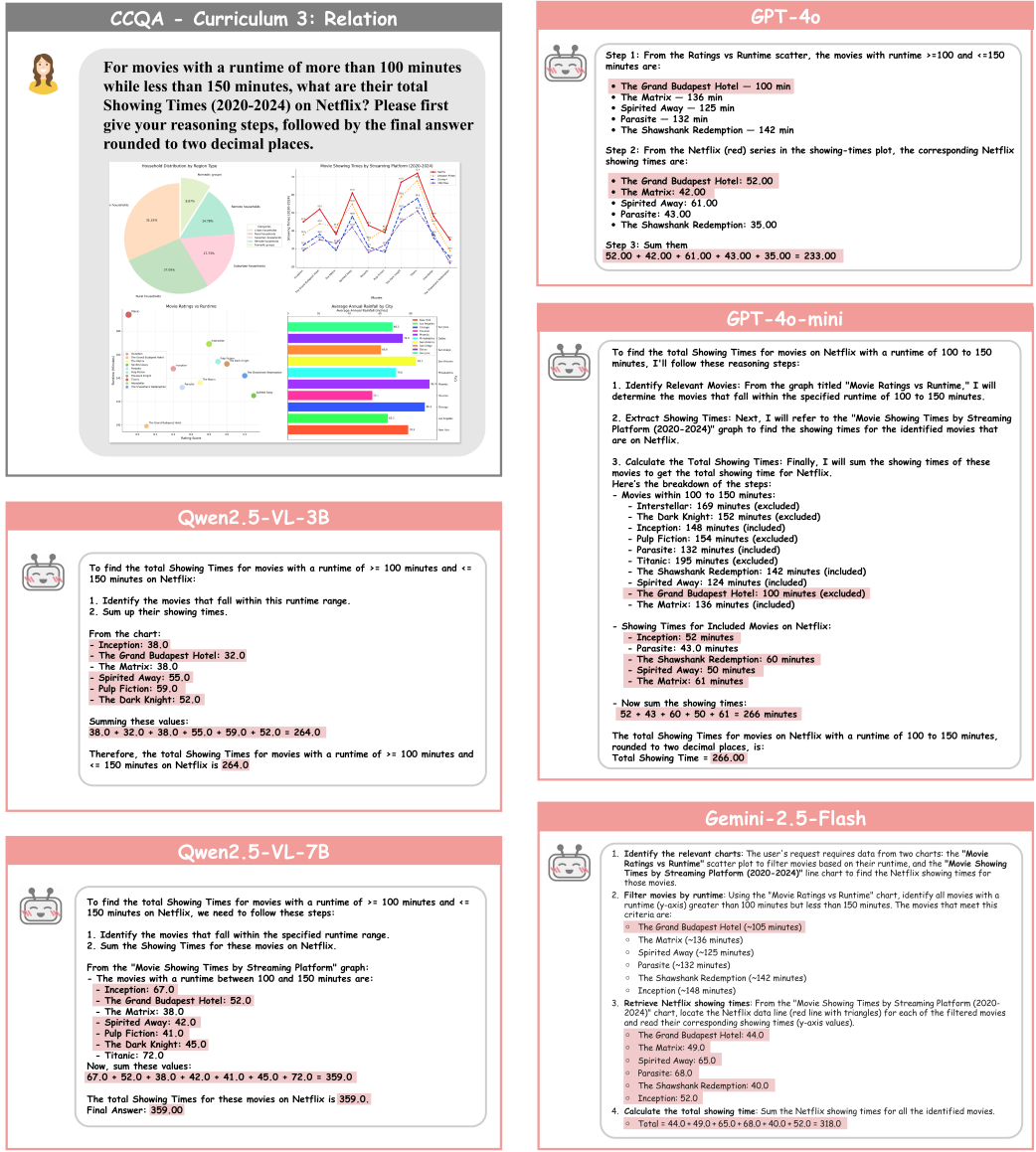}
    \caption{\textbf{Examples of Validation On Curriculum Level 3 ($D \geq 3$).} We validate our \oursdata on \textit{Tier 5} (\Sref{appendix:five fine-grained curriculum tiers}), i.e., \textit{Curriculum Level 3} ($D \geq 3$), by prompting the model to generate CoT reasoning and final answer. CoT reasoning prompt is omitted in the input for clarify.}
    \label{fig:appendix_preliminary_validation5}
\end{figure}

In validating our proposed curriculum learning benchmark, \oursdata (\S\ref{sec:dataset_construction}), we employ the same five MLLMs as our preliminary exploration (\S\ref{subsec:preliminary_exploration}), examining case studies on five fine-grained difficulty tiers (\S\ref{appendix:five fine-grained curriculum tiers}) of the three curriculum levels (\S\ref{subsec:curriculum_construction}):

\begin{itemize}
    \item \textbf{Tier 1: Single-Plot Reasoning ($D=1$).} The example in \Fref{fig:appendix_preliminary_validation1} is a \textit{Statistics-Mean} query (\Tref{tab:cqa_operators}) that MLLMs often fail to correctly answer. Among all three faiure cases, Qwen2.5-VL-3B and GPT-4o-mini fail at accurately perceive the numbers from the chart image, while Qwen2.5-VL-7B encounters calculation errors despite correct visual understanding.

    \item \textbf{Tier 2: Single-Plot Reasoning ($D=2$).} The example in \Fref{fig:appendix_preliminary_validation2} defines a specific \textit{Subset} (\Tref{tab:cqa_operators}) that poses significant obstacles for most MLLMs in arriving at the correct answer. Except for Gemini-2.5-Flask, all other four models struggle to accurately identify the required chart components.

    \item \textbf{Tier 3: Single-Plot Reasoning ($D \geq 3$).} The example in \Fref{fig:appendix_preliminary_validation3} involves multiple operators (\Tref{tab:cqa_operators}) that unfolds notable challenges for most MLLMs in both visual reasoning and mathematical calculation. Except for GPT-4o, all other four models struggle to accurately calculate the mean, leading to wrong final answers.

    \item \textbf{Tier 4: Multi-Plot Reasoning (Localization).} The example in \Fref{fig:appendix_preliminary_validation4} involves both \textit{Localization} and \textit{Statistics-Sum} (\Tref{tab:cqa_operators}). All five models fail to correctly perceive the visual information, resulting in wrong intermediate reasoning and incorrect final answer.

    \item \textbf{Tier 5: Multi-Plot Reasoning (Relation).} The example in \Fref{fig:appendix_preliminary_validation5} additionally pay attention to the \textit{Relation} (\Tref{tab:cqa_operators}) of different subplots. Likewise, the incorrect visual reasoning eventually lead to wrong final answers for all five models.
\end{itemize}

The diverging trends of failure cases (\textit{increase}) and success cases (\textit{decrease}) (\Fref{fig:appendix_preliminary_validation1}-\ref{fig:appendix_preliminary_validation5}) demonstrate the effectiveness of our curriculum design for progressively increased complexity tiers, meanwhile highlighting the challenges that persist in chart understanding across all reasoning depths and task complexity levels.

%%%%%%%%%%  Preliminary Exploration on CQA Challenges  %%%%%%%%%%

%%%%%%%%%%  Preliminary Validation on CCQA  %%%%%%%%%%
\begin{figure}[H]
    \centering
    \includegraphics[width=1.0\textwidth]{assets/c1_d1.pdf}
    \caption{\textbf{Examples of Validation On Curriculum Level 1 ($D=1$).} We validate our \oursdata on \textit{Tier 1} (\Sref{appendix:five fine-grained curriculum tiers}), i.e., \textit{Curriculum Level 1} ($D=1$), by prompting the model to generate CoT reasoning and final answer. CoT reasoning prompt is omitted in the input for clarify.}
    \label{fig:appendix_preliminary_validation1}
\end{figure}

\begin{figure}[H]
    \centering
    \includegraphics[width=1.0\textwidth]{assets/c2_d2.pdf}
    \caption{\textbf{Examples of Validation On Curriculum Level 2 ($D \geq 2$).} We validate our \oursdata on \textit{Tier 2} (\Sref{appendix:five fine-grained curriculum tiers}), i.e., \textit{Curriculum Level 2} ($D \geq 2$), by prompting the model to generate CoT reasoning and final answer. CoT reasoning prompt is omitted in the input for clarify.}
    \label{fig:appendix_preliminary_validation2}
\end{figure}

\clearpage
\section{Related Work}

% \subsection{Chain-of-Thought Reasoning}
\paragraph{Chain-of-Thought Reasoning.}
CoT reasoning has emerged as foundations for enhancing the interpretability and performance of large language models (LLMs). Even prompting LLMs to perform CoT reasoning before answering can improve performance \citep{cot_prompting_2022, cot_prompting_2023}.
CoT reasoning is particularly beneficial for MLLMs in complex visual reasoning tasks where attentions are interrelated to both visual and textual features \citep{cot_dpo_vlm_2024, visual_cot_2024, cot_vla_2025, cogcom_2024}.
Recent work enhances CoT via rationale-augmented training \citep{cot_dpo_vlm_2024} and feedback-based refinement \citep{measure_improve_cot_2024}, as well as incorporating simple visual operations into reasoning \citep{cogcom_2024}.
However, most approaches remain largely text-centric or loosely grounded, failing to tightly integrate visual evidence into each reasoning step.

\paragraph{Visual Grounding.}
Visual grounding aligns language with visual regions, enabling spatially-aware localization of entities, attributes, and relationships. 
Through \textit{explicit} visual localizations \citep{Gou2024NavigatingTD} or \textit{implicit} coordinate-free grounding \citep{Kang2025YourLVA}, recent advances in segmentation \citep{carion2025sam3segmentconcepts} and GUI grounding \citep{Cheng2024SeeClickHGA,Wu2025GUIActorCV} significantly improve text-visual alignment. 
Nevertheless, grounding is typically regarded as a standalone objective or terminal prediction, rather than an integral component of the reasoning process.
In contrast, our work leverages visual grounding as an intermediate and iterative medium within the reasoning chain, where visual changes dynamically inform subsequent reasoning steps and vice versa, shifting visual grounding from a passive alignment objective to an active component in interleaved multimodal reasoning.

\paragraph{Multimodal Chart Understanding.}
CQA represents a specialized task that requires accurate understanding of structured visual representations and complex reasoning over visual and textual elements.
Recent benchmarks have focused on real-world chart complexity and diversity, such as ChartQA \citep{chartqa_2022}, ChartQA-Pro \citep{chartqapro-2025}, ChartMuseum \citep{chartmuseum_2025}, CharXiv \citep{charxiv_2024}, etc.
Prior work addresses CQA via structured pipelines \citep{simplot_2024}, compositional reasoning \citep{gotcqa_2024}, improved chart representation \citep{chartformer_2024}, and synthetic data generation \citep{synthesize_2024}. However, these methods often decouple perception from reasoning or rely on static representations, limiting adaptive visual dynamics throughout reasoning.

\section{Dataset Construction}
\label{appendix: dataset construction}

\textbf{Chart Metadata.}
Extending our introduction to \oursdata (\S\ref{sec:dataset_construction}), we elaborate on the seven types of charts, the domain categories of the source plotting data, and fundamental operators that support multi-layer nested functions, as summarized in Tables~\ref{tab:data_construct}-\ref{tab:cqa_operators}.

\textbf{Data Augmentation.}
To effectively support curriculum learning with meta-learning insights (\S\ref{appendix:subsec:meta-learning}), we design a comprehensive set of chart-specific data augmentation strategies implemented through chart rendering functions. These augmentations introduce controlled variability in both the structural layout and visual presentation of charts, thereby enhancing model generalization across diverse chart types. Specifically, we consider the following transformations: (1) chart rotation at different angles (e.g., 0°, 30°, 45°, 60°, 75°, 90°); (2) orientation adjustments between vertical and horizontal layouts; (3) axis placement variations, such as shifting the $x$- and/or y-axis among left, right, top, and bottom; (4) color setting across chart elements; (5) legend positioning (top, bottom, center, left, or right) and visibility; (6) label positions and visibili(ty such as axis labels, tick labels, and numeric annotations); and other augmentation strategies tailored for specific types of charts (e.g., `explode' settings for pie charts, marker styles for scatter plots, etc.)
Collectively, these augmentation strategies form a systematic approach for generating richly diverse chart appearances using a small set of metadata, ensuring robustness and adaptability of models trained under meta-learning supported curriculum learning.

% Define custom colors
\definecolor{headergreen}{RGB}{137, 187, 129}
\definecolor{lightgreen}{RGB}{236, 244, 235}
\definecolor{mediumgray}{RGB}{230, 230, 230}

% First table - Description section
\begin{table}[H]

\vspace{10pt}

\centering
\renewcommand{\arraystretch}{1.3}
\begin{tabular}{>{\raggedright\arraybackslash}p{0.25\textwidth} >{\raggedright\arraybackslash}p{0.7\textwidth}}

\toprule
\rowcolor{headergreen}
\textcolor{white}{\textbf{Category}} & \textcolor{white}{\textbf{Description}} \\
\midrule
\rowcolor{white}
\textbf{Chart Types} & 

\begin{minipage}[c]{\linewidth}
\vspace{2pt}
\centering
% --- Row 1: 4 figures ---
\begin{tabular}{cccc}
Bar & Histogram & Scatter & Line \\
\includegraphics[width=2.0cm,height=1.5cm]{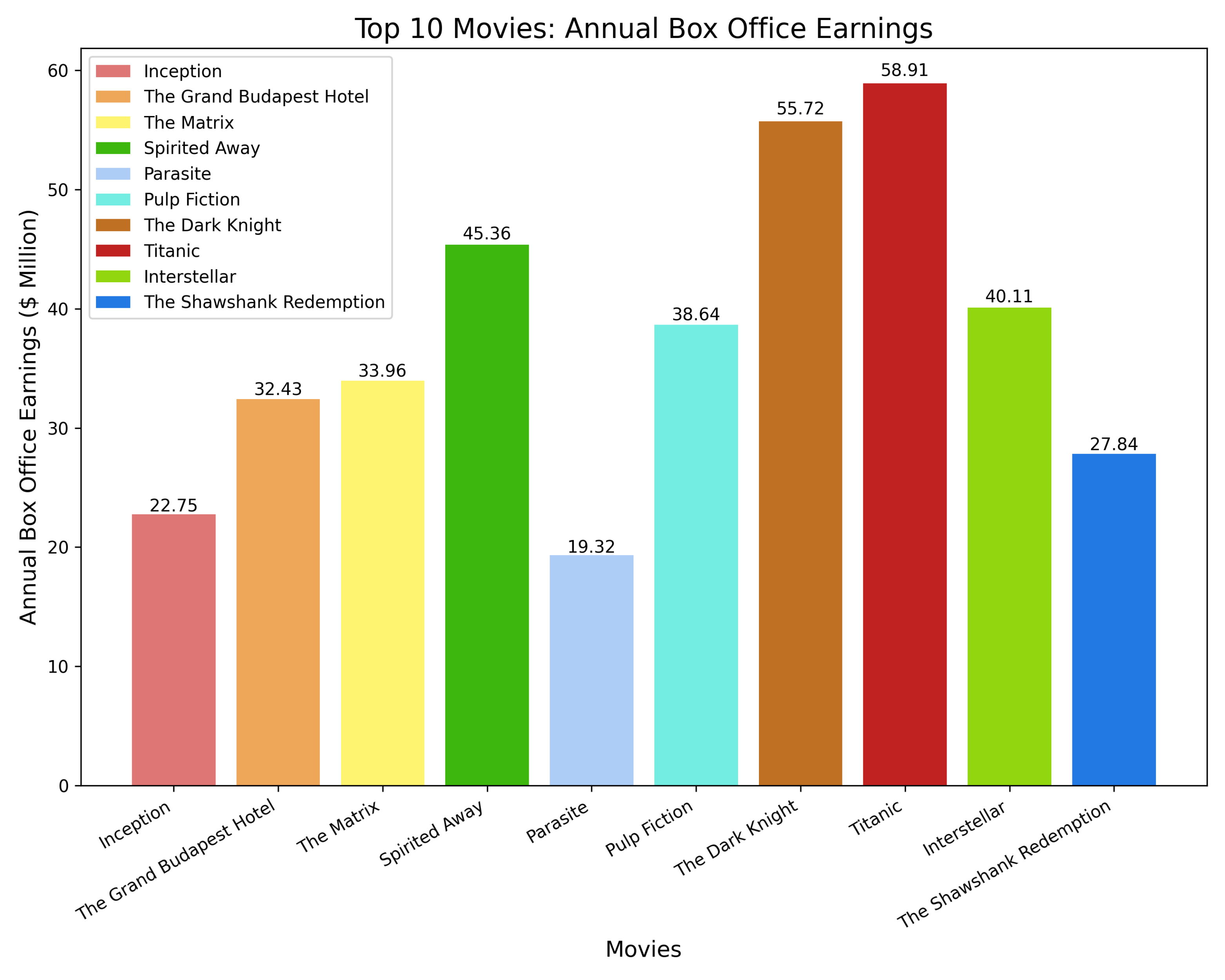} &
\includegraphics[width=2.0cm,height=1.5cm]{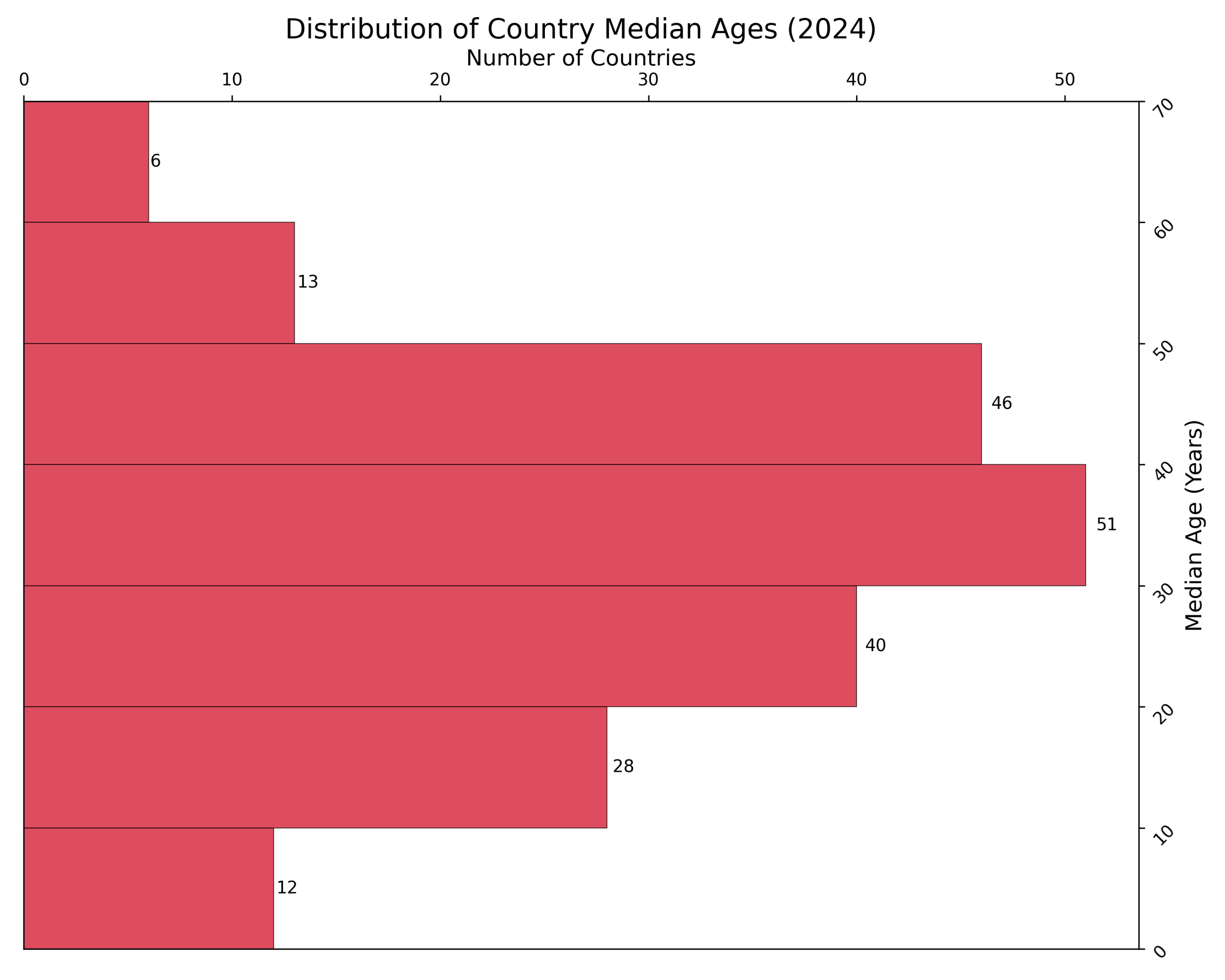} & \includegraphics[width=2.0cm,height=1.5cm]{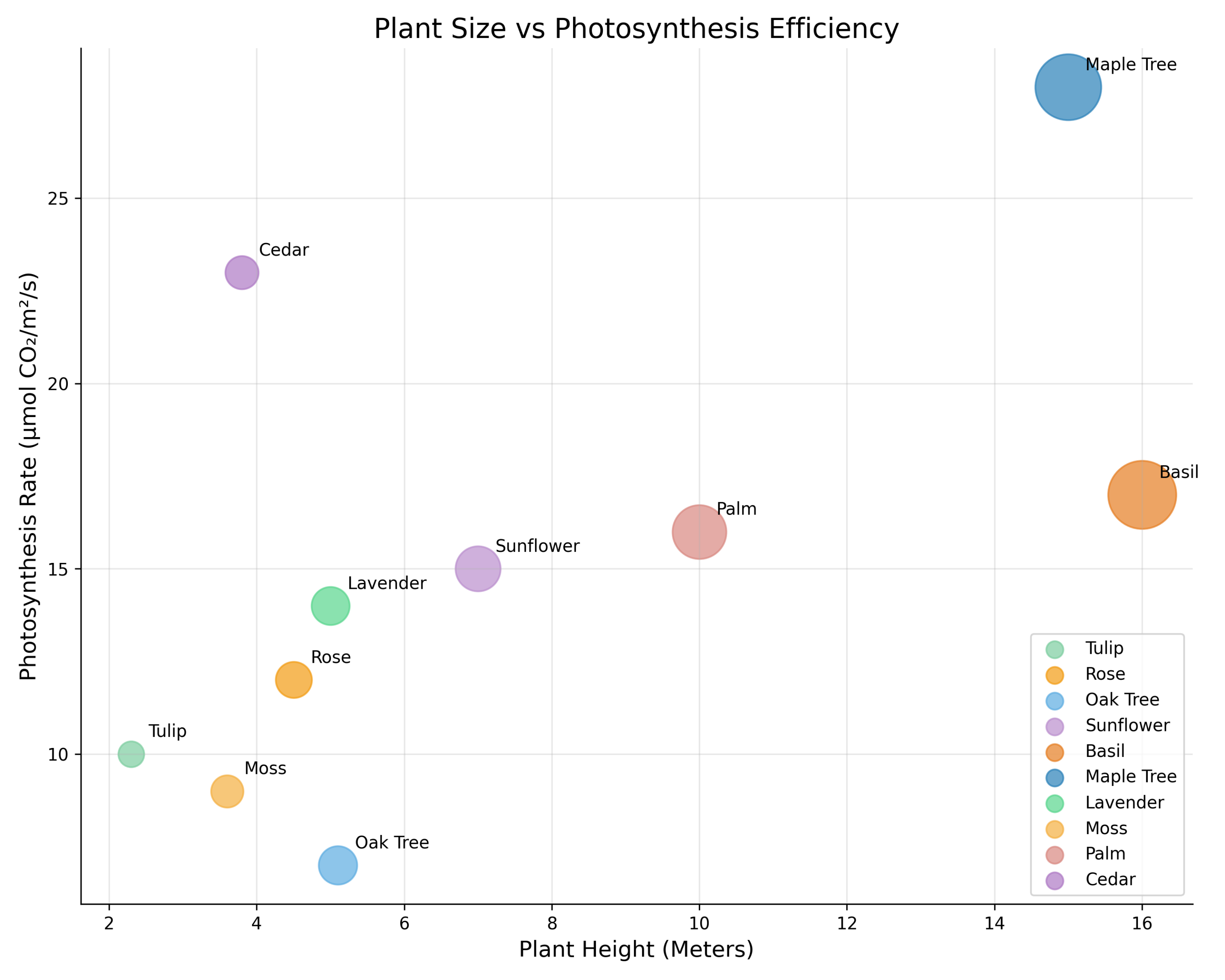} &
\includegraphics[width=2.0cm,height=1.5cm]{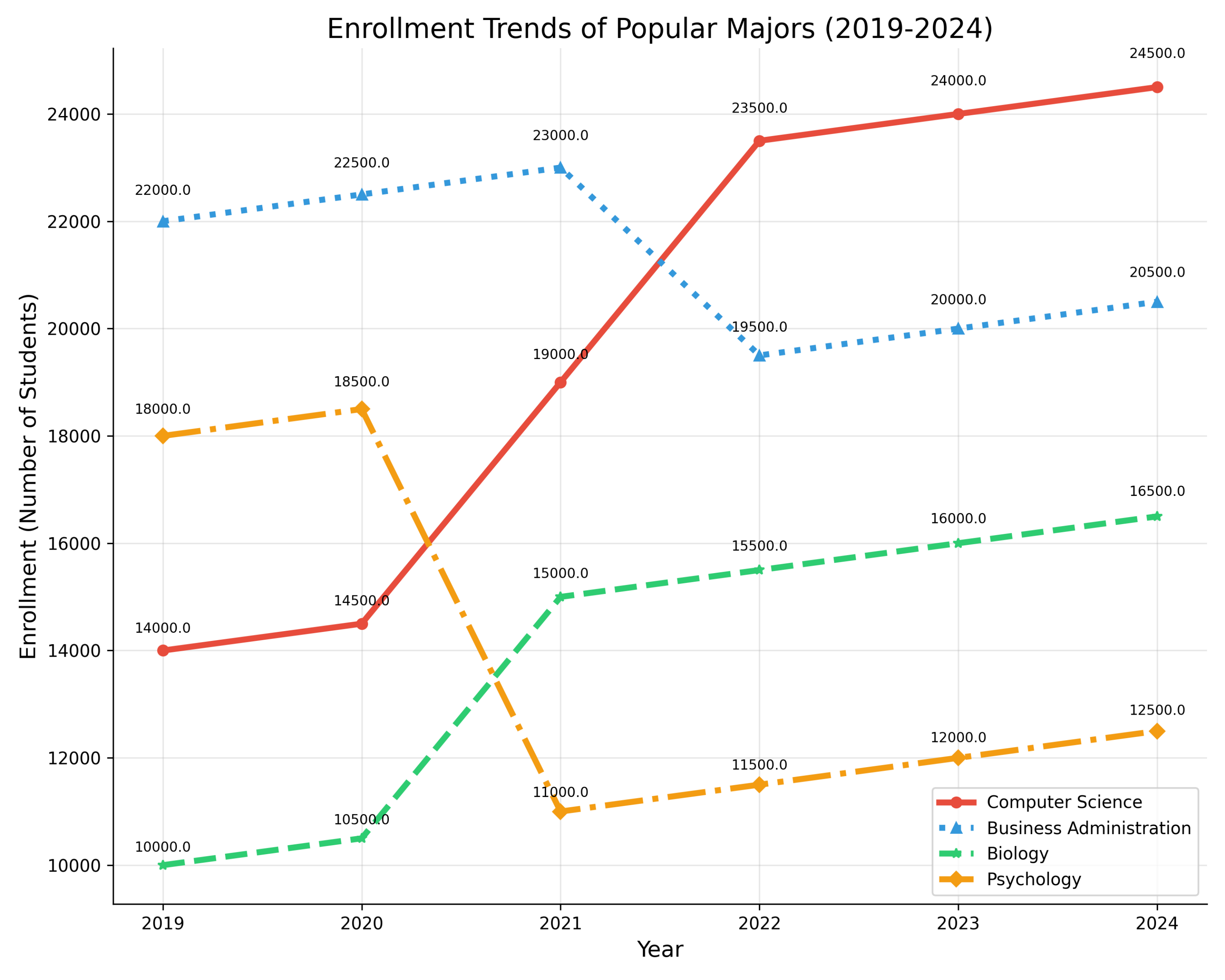} \\
\end{tabular}

\vspace{0.4cm}

% --- Row 2: 3 figures ---
\begin{tabular}{ccc}
Heatmap & Pie & Radar \\
\includegraphics[width=2.0cm,height=1.5cm]{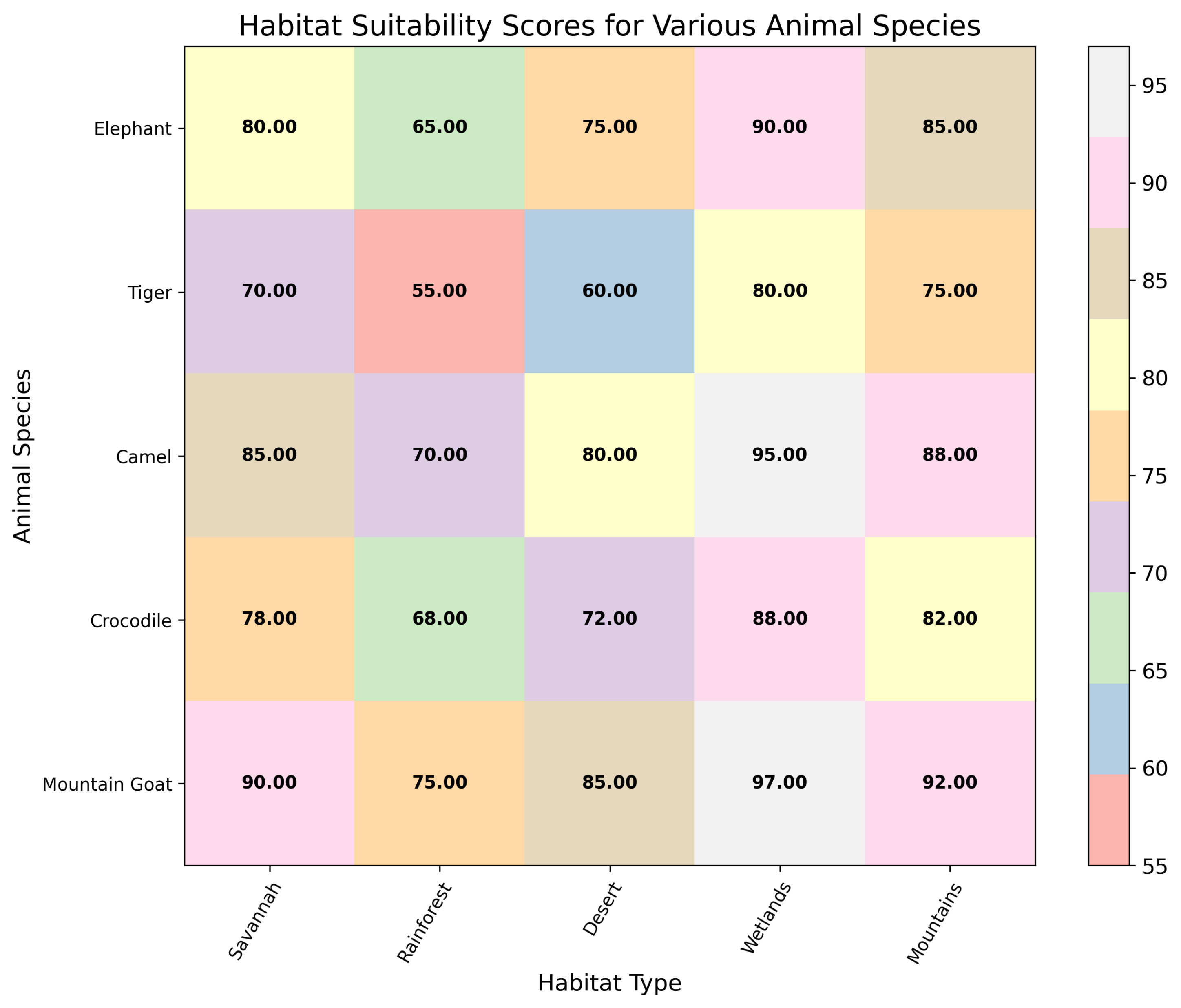} &
\includegraphics[width=2.0cm,height=1.5cm]{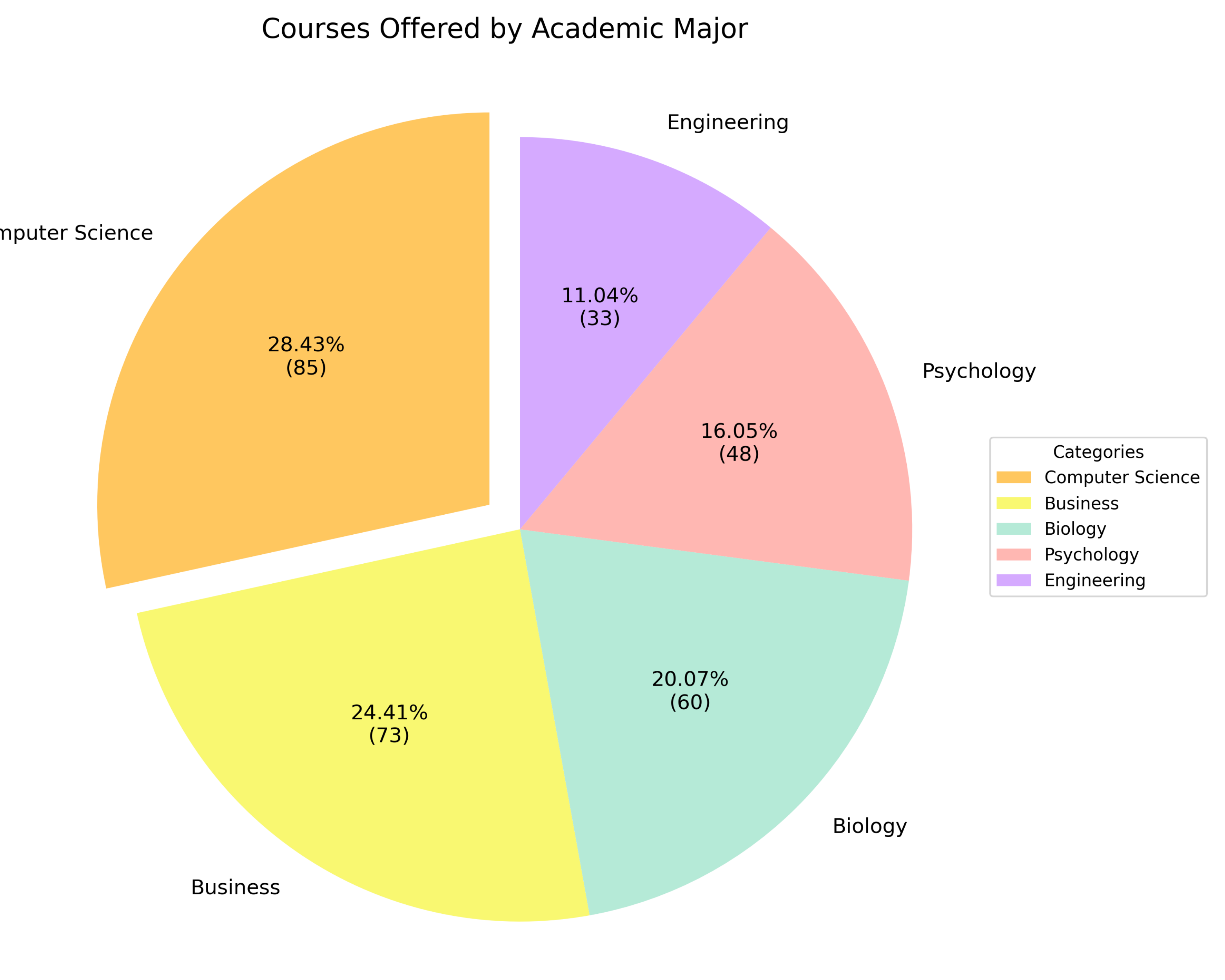} &
\includegraphics[width=2.0cm,height=1.5cm]{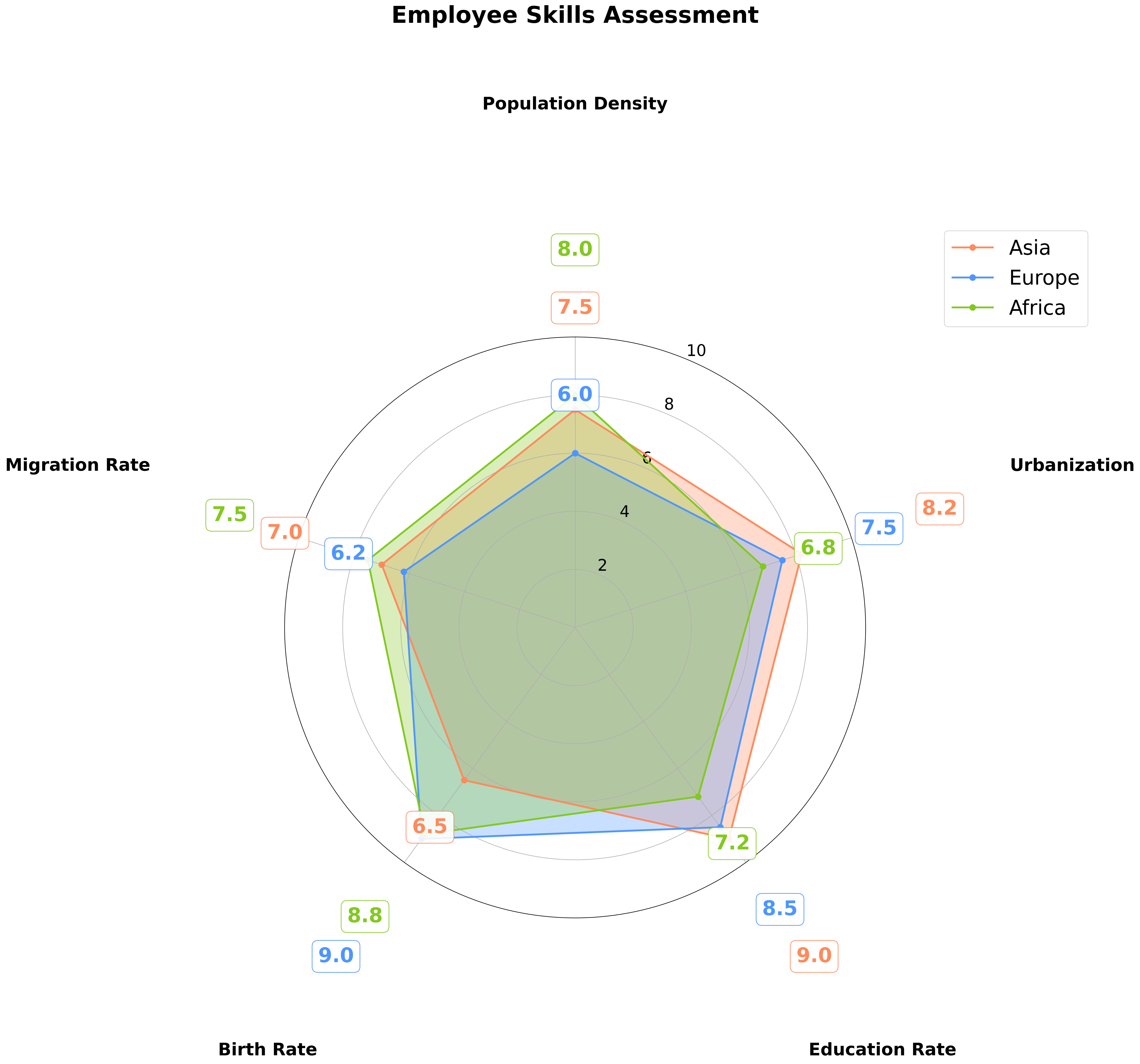} \\
\end{tabular}
\vspace{2pt}
\end{minipage} \\

% Bar Chart, Histogram, Scatter Plot, Line Chart, Heatmap, Pie Chart, Radar Chart \\
\rowcolor{lightgreen}
\textbf{Domain Categories} & 
\begin{minipage}[c]{11cm}
\vspace{2pt}
\begin{enumerate}[leftmargin=0.5cm, itemsep=1pt, parsep=0pt]
\footnotesize
\item Media \& Entertainment
\item Geography \& Demography
\item Education \& Academia
\item Business \& Industry
\item Major \& Course
\item Animal \& Zoology
\item Plant \& Botany
\item Biology \& Chemistry
\item Food \& Nutrition
\item Space \& Astronomy
\item Sale \& Merchandise
\item Market \& Economy
\item Sports \& Athletics
\item Computing \& Technology
\item Health \& Medicine
\item Energy \& Environment
\item Travel \& Expedition
\item Arts \& Culture
\item Communication \& Collaboration
\item Language \& Linguistics
\item History \& Archaeology
\item Weather \& Climate
\item Transportation \& Infrastructure
\item Psychology \& Personality
\item Materials \& Engineering
\item Philanthropy \& Charity
\item Fashion \& Apparel
\item Parenting \& Child Development
\item Architecture \& Urban Planning
\item Gaming \& Recreation
\end{enumerate}
\vspace{2pt}
\end{minipage} \\
\bottomrule
\end{tabular}

\caption{\textbf{Metadata of Chart Plotting.} We employ seven commonly used chart types across 30 different domain categories to construct the meta images for \oursdata.}
\label{tab:data_construct}

\end{table}

% Second table - CQA Task Types
\begin{table}[H]

\vspace{5pt}

\centering
\renewcommand{\arraystretch}{1.3}
\begin{tabular}{>{\raggedright\arraybackslash}p{0.25\textwidth} >{\raggedright\arraybackslash}p{0.7\textwidth}}

\toprule
\rowcolor{headergreen}
\textcolor{white}{\textbf{Operator}} & \textcolor{white}{\textbf{Description}} \\
\midrule
\rowcolor{white}
\textbf{Read} & Read or estimate the value of chart component(s) that meet given requirement(s) \\
\rowcolor{lightgreen}
\textbf{Statistics - Sum/Mean/Median} & Calculate the sum/mean/median of a group of chart components that meet given requirement(s) \\
\rowcolor{white}
\textbf{Statistics - Count} & Count the number of chart components that meet given requirement(s) \\
\rowcolor{lightgreen}
\textbf{Extrema - Value - Min/Max} & Calculate the minimum/maximum value (which may be combined with nested functions, \textit{e.g.,} the minimum mean value of two groups of chart components) of chart components that meet given requirement(s) \\
\rowcolor{white}
\textbf{Extrema - Position} & Localize chart component(s) that meet given requirement(s), \textit{e.g.,} the leftmost bar in the bar chart \\
\rowcolor{lightgreen}
\textbf{Sort - Ascending/Descending} & Sort a group of chart components that meet given requirement(s) \\
\rowcolor{white}
\textbf{Compare - Value/Diff/Position} & Compare the value/difference/position of two groups of chart components based on the given requirement(s) \\
\rowcolor{lightgreen}
\textbf{Filter} & Filter chart component(s) based on the given requirement(s) \\
\rowcolor{white}
\textbf{Threshold} & Identify chart component(s) based on the given threshold condition(s) \\
\rowcolor{lightgreen}
\textbf{Subset} & Identify the subset of chart component(s) that satisfy the specified requirement(s) \\
\rowcolor{white}
\textbf{Localization} & Localize specific chart components and/or subplots \\
\rowcolor{lightgreen}
\textbf{Relation} & Understand relations between or among different chart components and/or subplots \\
\bottomrule
\end{tabular}

\caption{\textbf{Foundational Operators.} Our \oursdata incorporates 12 basic operators to query different aspects of chart components, facilitating comprehensive understanding of each chart elements.}
\label{tab:cqa_operators}

\vspace{20pt}

\end{table}

\vspace{10pt}
\begin{figure}[H]
    \centering
    \includegraphics[width=1.0\textwidth]{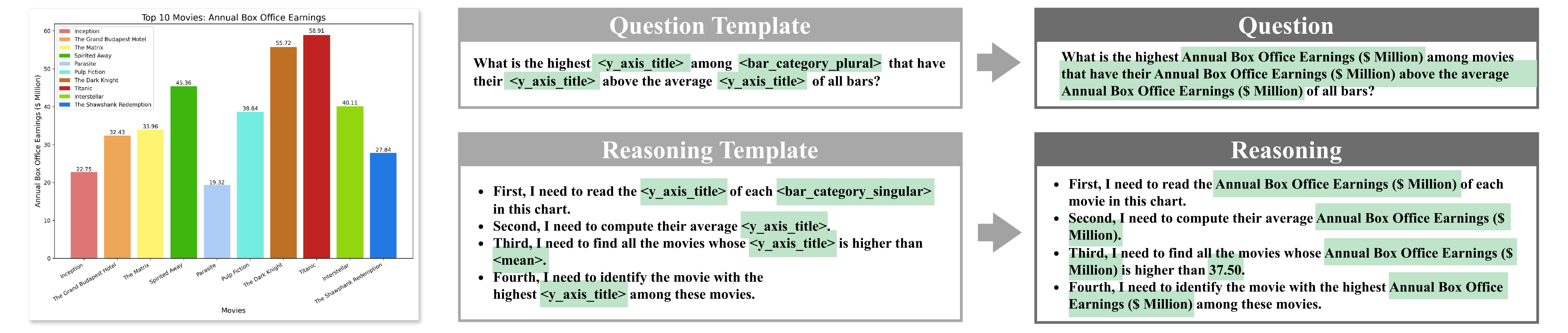}
    \caption{\textbf{From Template To CQA Data.} A template-based data generation example that illustrates how question and reasoning templates are converted to CQA data based on the chart data.}
    \label{fig:template_example}
\end{figure}

\subsection{Data Structure}
\label{appendix:subsec:data_structure}

\begin{figure}[H]
    \centering
    \includegraphics[width=1.0\textwidth]{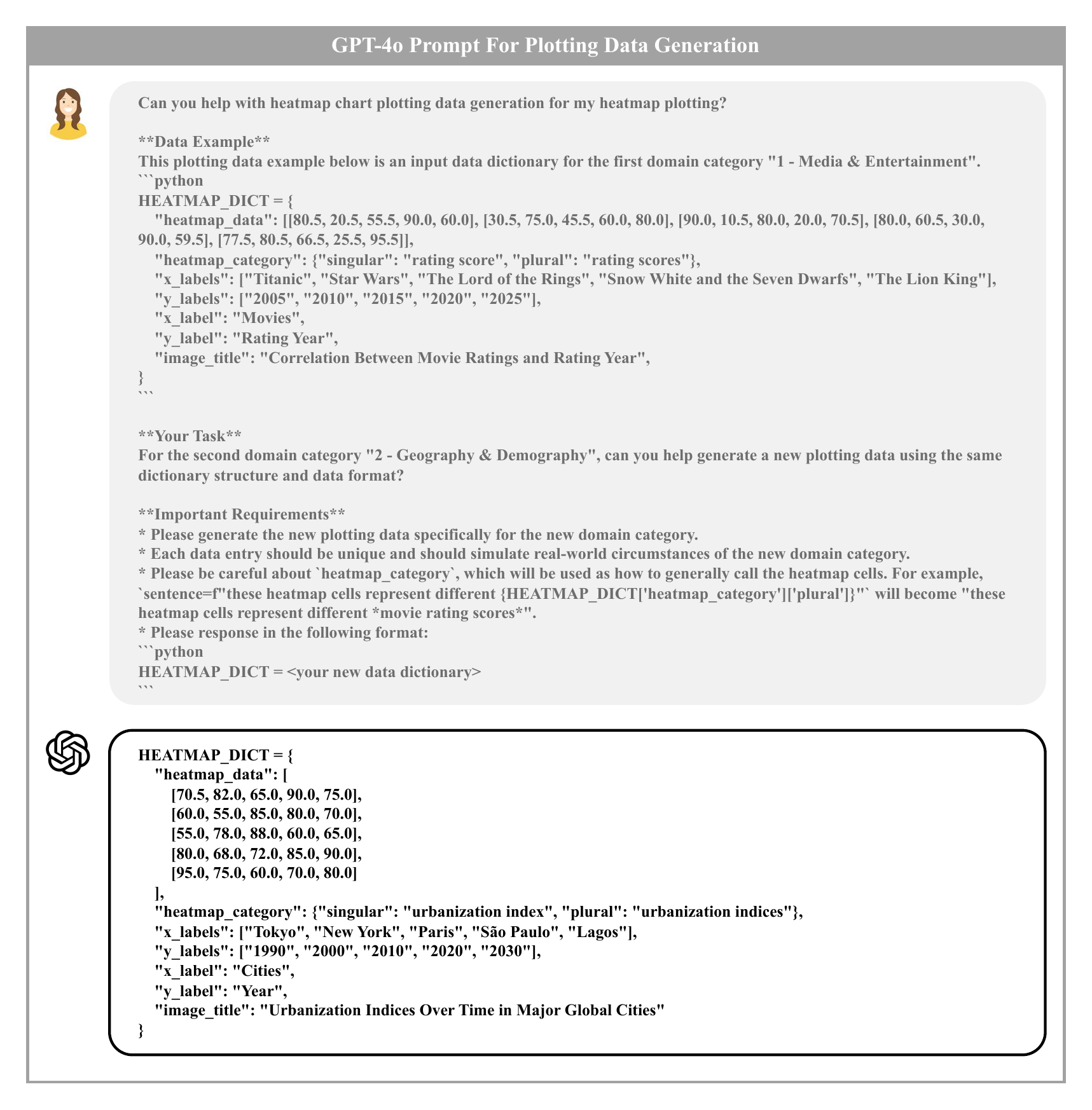}
    \vspace{-20pt}
    \caption{\textbf{Example of Plotting Data Generation Prompt.} We prompt GPT-4o to generate the source plotting data, which will be used as the input for chart drawing functions. This example is for the heatmap plotting data generation of the second domain category (\Tref{tab:data_construct}).}
    \label{fig:appendix_gpt4o_plotting_data_generation}
\end{figure}

Our \oursdata (\S\ref{sec:dataset_construction}) encompass seven basic chart types, including \textit{bar chart, histogram, scatter plot, line chart, heatmap, pie chart, and radar chart} (\Tref{tab:data_construct}).
Only the chart plotting data (\textit{i.e.}, the value and label of each chart component, along with the axis and image titles) are generated by GPT-4o (see an example in~\Fref{fig:appendix_gpt4o_plotting_data_generation}).
We construct the CQA data of each chart type and curriculum level across 30 domain categories (\Tref{tab:data_construct}).
The number of samples for each chart type is influenced by chart features (e.g., scatter plots depend on both X- and Y-axis features, whereas heatmaps depend on cell values and labels), CQA types (\Tref{tab:data_construct}, \ref{tab:cqa_operators}), and properties of the source plotting data (e.g., the number of bars, scatter points, or cells; variations in label angles; etc.).
To ensure high data quality, we implement template-based CQA generation that guarantees not only the diversity of CQA tasks but also the accuracy and reliability of intermediate reasoning, visual grounding, and final answers.

Specifically, all question-answer pairs in \oursdata, together with their corresponding reasoning steps and dynamic visual grounding coordinates, are generated using human-defined templates and functions (\S\ref{subsec:dataset_principles}).
An example is shown in \Fref{fig:template_example} to illustrate the template-based CQA data generation process.

\vspace{10pt}

% Moved from section 4.2 to here
\subsection{Multi-Level Curriculum}
\label{subsec:curriculum_construction}

Implementing curriculum learning to progressively increase reasoning difficulty across three distinct levels (\Fref{fig:dataset_construction}), each level targets at specific aspects of visual reasoning development (\Tref{tab:cqa_operators}): \textit{read}, \textit{statistics}, \textit{extrema}, \textit{sorting}, \textit{comparison}, \textit{filtering}, \textit{thresholding}, \textit{subset constraints}, \textit{localization}, and \textit{relation}.
The knowledge transfer between levels contributes to the increase of task complexity (\S\ref{appendix: dataset construction}).

\paragraph{\textsc{Level 1}: Foundational Single-Operation Reasoning.}

\textsc{Level 1} establishes fundamental chart understanding capabilities (\textit{reasoning depth}: $D_1=1$) with single-operation reasoning processes on single-plot charts:

\vspace{-3.6mm}
\begin{align}
\quad & \textbf{\texttt{SINGLE}}(x) = f(x) \label{eq:level1_operation}
\end{align}
\vspace{-5.6mm}

where $f(\cdot)$ represents basic operations such as direct value reading, simple arithmetic, and elementary comparisons within a single-plot chart.
Hereby, the \textbf{\textit{reasoning depth}} of a CQA data sample, denoted as $D_l$, is defined as the number of nested operations for curriculum level $l$.

Accordingly, the template structure of \textsc{Level-1} CQA data follow the format:

\vspace{-3.6mm}
\begin{equation}
\textsc{Template}_1 = \{Q, \{R_t, B_t^*\}_{t=1}^{T_1}, A\}
\label{eq:level1_template}
\end{equation}
where $Q$ is the question, $T_1 \geq 1$ is the number of reasoning steps, $R_t$ is the $t$-th reasoning step, $B_t^*$ is the corresponding ground-truth visual grounding with reasoning depth $D_1 = 1$, and $A$ is the final answer.

\paragraph{\textsc{Level 2:} Multi-Operation Reasoning.}

\textsc{Level 2} introduces compositional reasoning (\textit{reasoning depth}: $D_2>1$) through nested operations on single charts:

\vspace{-3.6mm}
\begin{align}
\quad & \textbf{\texttt{SINGLE}}(x) = F(f(x)) \label{eq:level2_operation}
\end{align}
\vspace{-5.6mm}

where $F(\cdot)$ represents composite operations applied to $f(x)$, \textit{e.g.,} $F(f(x))=h(g(f(x)))$.
This level requires models to perform sequential reasoning and visual grounding where each step builds upon previous computations. Consequently, \textsc{Level 2} templates extend to multi-step reasoning:

\vspace{-3.6mm}
\begin{equation}
\textsc{Template}_2 = \{Q, \{R_t, B_t^*\}_{t=1}^{T_2}, A\}
\label{eq:level2_template}
\end{equation}
where $T_2 \geq 2$ is the number of reasoning steps, each reasoning step $R_t$ progresses through nested operations with reasoning depth $D_2 \geq 2$, and $B_t^*$ is the corresponding ground-truth visual grounding for the $t$-th step.

\paragraph{\textsc{Level 3}: Complex Multi-Chart Reasoning}

\textsc{Level 3} represents the most challenging scenarios that involve complex reasoning ($D_3 \geq 2$) across multiple subplots and chart types:

\vspace{-3.6mm}
\begin{align}
\quad & \textbf{\texttt{MULTI}}(x) = \textbf{\texttt{MULTI}}(\textbf{\texttt{SINGLE}}_i(x)) \label{eq:level3_operation}
\end{align}
\vspace{-5.6mm}

where $i$ conforms to $1 \leq i \leq n$ and $1 \leq n \leq \textit{subplot\_num}$. \textsc{Level 3} templates thereby incorporate multi-step and cross-chart dependencies:

\vspace{-3.6mm}
\begin{equation}
\textsc{Template}_3 = \{Q, \{R_t, B_t^*, C_t\}_{t=1}^{T_3}, A\}
\label{eq:level3_template}
\end{equation}
where $T_3 \geq 3$ is the number of reasoning steps, each reasoning step $R_t$ progresses through complex nested operations with reasoning depth $D_3 \geq 3$, $B_t^*$ is the corresponding ground-truth visual grounding, and $C_t$ indicates the chart index for the $t$-th reasoning step. Specifically, our multi-plot reasoning incorporates both localization $(n-1)$ and relation $(n>1)$ operations across multiple charts.

\subsection{Fine-Grained Curriculum Tiers}
\label{appendix:five fine-grained curriculum tiers}

We construct our three-level curriculum dataset (\Fref{fig:dataset_construction}) based on reasoning depth and chart complexity (\S\ref{subsec:curriculum_construction}).
According to their fine-grained problem-solving difficulty, we categorize them into five curriculum tiers:

\begin{itemize}
    \item \textbf{Tier 1: Curriculum Level 1 ($D=1$).} In \textit{Tier 1}, all CQA data correspond to queries about single-plot chart image input. Reasoning is limited to one depth level, \textit{i.e.}, single-function reasoning (\Eref{eq:level1_operation}, $D_1 \geq 1$), to derive the final answer. 
    
    \item \textbf{Tier 2: Curriculum Level 2 ($D=2$).} In \textit{Tier 2}, all CQA data correspond to queries about single-plot chart image input. Reasoning requires two depth levels, \textit{i.e.}, constructed through two nested functions (\Eref{eq:level2_operation}, $D_2 \geq 2$), to derive the final answer.

    \item \textbf{Tier 3: Curriculum Level 2 ($D \geq 3$).} In \textit{Tier 3}, all CQA data correspond to queries about single-plot chart image input, with reasoning that involves three or more depth levels, \textit{i.e.}, constructed through three or more nested functions (\Eref{eq:level2_operation}, $D_2 \geq 3$), to derive the final answer.

    \item \textbf{Tier 4: Curriculum Level 3 - Localization ($D \geq 3$).} In \textit{Tier 4}, all CQA data correspond to queries about multi-plot chart image input, with reasoning that involves three or more depth levels, \textit{i.e.}, constructed through three or more nested functions (\Eref{eq:level3_operation}, $D_3 \geq 3$), to derive the final answer. While different from single-plot charts, multi-plot CQA tasks in \textit{Tier 4} involves the precise localization of target subplot(s) that directly yield the answer.

    \item \textbf{Tier 5: Curriculum Level 3 - Relation ($D \geq 3$).} CQA tasks in \textit{Tier 5} are similar to the constitution of \textit{Tier 4}, corresponding to queries about multi-plot charts with reasoning that involves three or more depth levels, \textit{i.e.}, constructed through three or more nested functions (\Eref{eq:level3_operation}, $D_3 \geq 3$). The key distinction is that, while \textit{Tier 4} emphasizes precise localization of target subplot(s), \textit{Tier 5} additionally demands the modeling of relations across the identified subplots.
\end{itemize}

\subsection{Meta-Learning Supported Curriculum Learning}
\label{appendix:subsec:meta-learning}

Our curriculum learning design (\S\ref{sec:methodology}) is reinforced through meta-learning, which provides a principled way to structure both data and task complexity. Specifically, we leverage meta-learning through the following aspects:

\begin{enumerate}[leftmargin=*]
    \item \textbf{Domain diversity as meta-tasks.} We construct \oursdata using 30 domain categories (\Tref{tab:data_construct}), where each category contributes one source plotting data, and thus one chart image. This structured diversity provides a wide range of meta-tasks that expose MLLMs to domain-generalizable visual reasoning.

    \item \textbf{Chart-type variability as meta-structures.} We employ 7 fundamental chart types (\Tref{tab:data_construct}) to visualize the 30 domain datasets. Multiplying 30 plotting datasets by 7 chart types yields 210 unique chart-structure metadata, based on which the entire dataset is systematically constructed. This ensures that each domain is represented across diverse chart structures, promoting cross-task adaptation.  

    \item \textbf{Operator set as meta-functions.} To support multi-layer nested reasoning, we define 12 fundamental operators (\Tref{tab:cqa_operators}). These operators serve as compositional primitives for constructing multi-level CQA tasks. By progressively increasing the depth of nesting, we control CQA difficulty level, thereby enabling MLLMs to gradually acquire higher-order reasoning capabilities.  

    \item \textbf{Decomposition as learning scaffolds.} Following decomposition insights (\S\ref{appendix:subsec:preliminary_on_method_motivation}), we disentangle each task into \textit{visual decomposition} (low-level chart components) and \textit{reasoning decomposition} (singular operations across nested functions). This scaffolding allows MLLMs to incrementally learn fine-grained visual perception and step-wise logical reasoning, supporting the high-level composition of accurate chart understanding.  

    \item \textbf{Meta-learning for transferability.} Beyond dataset construction, our design leverages meta-learning to encourage transferability across chart types, domains, and reasoning depths. By repeatedly exposing MLLMs to varied meta-tasks with systematically controlled complexity, we enable them to acquire generalizable strategies rather than overfitting to particular chart types or reasoning templates. Our meta-learning implementation strengthens the robustness of curriculum learning by aligning it with principles of adaptation and generalization.  
\end{enumerate}  

Together, these design principles ensure that our curriculum learning is not only systematic but also meta-learnable, allowing MLLMs to progressively integrate visual and reasoning competencies across tasks of increasing complexity.

\subsection{Quality Control}
\label{appendix:subsec:dataset_construction_quality_control}

To ensure high data fidelity for both training and evaluation, we enforce rigorous quality control throughout the construction of \oursdata:

First, we implement the meta-learning-driven construction pipeline in which each chart type is defined by human-authored meta-functions. These meta-functions explicitly encode the data generation logic and reasoning structure, with configurable chart attributes such as the number of chart elements, spatial arrangements, structural relationships, shapes and color settings, etc. This design enables precise and controllable synthesis of visual reasoning samples while preserving data diversity across multiple chart visual dimensions, such as colors, shapes, sizes, positions, etc. Prior to inclusion in the dataset, all generated instances are pre-executed to verify functional correctness and internal consistency.

Second, we perform manual validation of the visualization metadata, including ground-truth visual grounding masks and the corresponding step-by-step reasoning annotations associated with each metadata instance. In addition to manually verifying the 30 \textit{meta images} for each type of charts, we randomly sample 10 visual reasoning instances of each category (\textit{i.e.}, $10 \times 30 \times 7$ in total) to validate accurate vision-reasoning alignment (Tab.~\ref{tab:human_validation}). This process ensures that every visual grounding step accurately corresponds in multimodal spaces, and that the resulting reasoning chains are coherent, interpretable, and faithful to the underlying data.

Critically, we restrict the role of LLM (GPT-4o) to generating randomized numerical parameters within predefined domain categories for chart construction, which will only be used as inputs in plotting functions as visual grounded reasoning chain templates. That is, these parameters are deterministically incorporated into plotting functions and template-based reasoning chains, pre-computed in predefined equations prior to data instantiation. As a result, the use of LLMs is ensured to exhibit no effect on the correctness of ground-truth answers, thereby preserving the overall reliability and integrity of the dataset.

\begin{table}[h]
\centering
\small

\vspace{16pt}
\renewcommand{\arraystretch}{1.5}
\resizebox{\textwidth}{!}{%
\begin{tabular}{l | ccccccc}
\toprule
\textbf{Human Validation} & \textbf{Bar} & \textbf{Histogram} & \textbf{Scatter} & \textbf{Line} & \textbf{Heatmap} & \textbf{Pie} & \textbf{Radar} \\
\midrule
Meta Image   & 30 & 30 & 30 & 30 & 30 & 30 & 30 \\
CQA Instance & 10 & 10 & 10 & 10 & 10 & 10 & 10 \\
\bottomrule
\end{tabular}%
}

\caption{\textbf{Human Validation on Data Quality}}
\label{tab:human_validation}

\end{table}

\section{Visual Grounding Strategies}
\label{appendix: 3 visual grounding strategies}

We propose three visual grounding strategies --- \textit{applied}, \textit{boxed}, and \textit{cropped} (\Fref{fig:methodology}, \Tref{tab:grounding_comparison}) --- to enable \textit{dynamic} visual focus navigation throughout the evolution of multi-step reasoning. All three strategies follow the same \modeRVA process where the model generates reasoning steps accompanied by grounded bounding box coordinates, while the ``\textit{dynamic}" nature refers to how the visual focus adaptively changes as the train of thoughts progresses.
Each strategy implements a distinct grounding mechanism for directing the model's visual attention to corresponding image regions of focus while maintaining coherent reasoning flow. On the other hand, these strategies also represent different trade-offs between visual clarity, computational efficiency, and reasoning precision (\Tref{tab:grounding_comparison}).

\begin{table}[!ht]
\centering
\small

\vspace{20pt}

\resizebox{\textwidth}{!}{%
\begin{tabular}{l|c|c|c|c|c|c|c}
\toprule

\textbf{Method} & \textbf{\shortstack{Low \\ Computation}} & \textbf{\shortstack{High \\ Precision}} & \textbf{\shortstack{Full \\ Context}} & \textbf{\shortstack{No \\ Occlusion}} & \textbf{\shortstack{Multi- \\ Region}} & \textbf{\shortstack{Easy \\ Integration}} & \textbf{\shortstack{Easy \\ Comprehension}} \\

\midrule

\textbf{\textit{applied}} & \cmark & \xmark & \cmark & \xmark & \cmark & \cmark & \cmark \\
\midrule
\textbf{\textit{boxed}} & \cmark & \xmark & \cmark & \cmark & \cmark & \cmark & \cmark \\
\midrule
\textbf{\textit{cropped}} & \xmark & \cmark & \cmark & \cmark & \cmark & \xmark & \xmark \\
\bottomrule
\end{tabular}%
}

\caption{\textbf{Comparative Analysis of Visual Grounding Strategies.} We evaluate each grounding strategy across seven key dimensions that are significant for effective visual reasoning. \cmark~indicates the strategy possesses the advantage, while \xmark~indicates limitation. \textit{Boxed} grounding achieves the best overall balance, \textit{applied} grounding provides clear interpretability and highlighting with moderate trade-offs, and \textit{cropped} grounding maximizes precision at the cost of computational efficiency and understanding straightforwardness.}
\label{tab:grounding_comparison}

\vspace{10pt}

\end{table}

\subsection{\textbf{Applied}: Grounding Through Dynamic Visual Focus Highlighting}
\label{appendix: visual grounding strategy - applied}

\paragraph{Method.}
The \textit{applied} grounding strategy directly underlines the predicted regions of focus through semi-transparent yellow highlighting overlays. As the reasoning progresses, the yellow highlighting adaptively shifts to emphasize different visually focused regions mirroring each reasoning step, while preserving full visual context.

Specifically, our \textit{applied} grounding strategy implements visual grounding through semi-transparent highlighting overlays that mask visual focuses:

\vspace{-2.5mm}
\begin{equation}
I'_{vis,t} = I_{orig} \odot (1 - \kappa \cdot M_{focus,t}) + \kappa \cdot H_{yellow} \odot M_{focus,t}
\label{eq:applied_grounding}
\end{equation}
where $I_{orig}$ is the original input image, $M_{focus,t}$ is the binary mask derived from the list of bounding boxes $\{B_{t,i}\}_{i=1}^{N_t}$ where each $B_{t,i} = [x_{min}, y_{min}, x_{max}, y_{max}]$ and $N_t$ is the number of bounding boxes at reasoning step $t$, $H_{yellow}$ is the highlight color (\textit{i.e.,} yellow), $\kappa$ controls transparency, and $\odot$ denotes element-wise multiplication.

\begin{figure}[H]
    \vspace{50pt}
    
    \centering
    \includegraphics[width=1.0\textwidth]{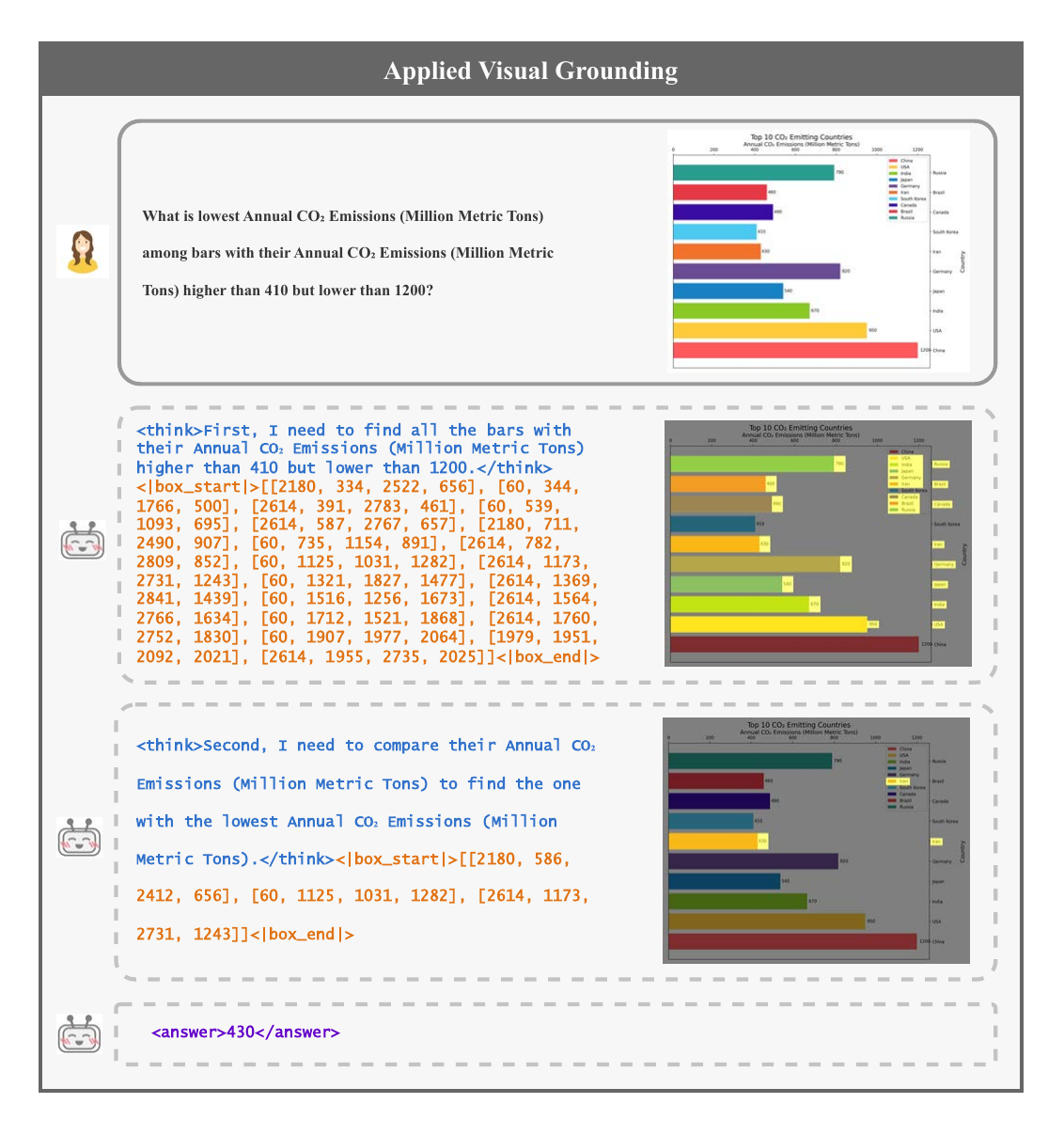}
    
    \caption{\textbf{Example of Applied Visual Grounding.} The \textit{applied} visual grounding method directly accentuates the regions of focus through semi-transparent yellow highlighting overlays.}
    \label{fig:A1_applied_examples}

    \vspace{30pt}
\end{figure}

\paragraph{Examples.}
\Fref{fig:A1_applied_examples} shows an example for multi-step CoT reasoning with \textit{applied} visual grounding.

\subsection{\textbf{Boxed}: Grounding Through Dynamic Visual Box Guides}
\label{appendix: visual grounding strategy - boxed}

\paragraph{Method.}
The \textit{boxed} grounding strategy straightforwardly guides visual attention by adding red rectangular borders to the focus regions. These red boxes dynamically relocate and resize along with the evolution of the reasoning chain, emphasizing current regions of focus through explicit visual boundaries.

Particularly, the border guides are generated by drawing rectangular outlines at the specified coordinates:

\vspace{-2.5mm}
\begin{equation}
I'_{vis,t} = \textsc{Box}(I_{orig}, \{B_{t,i}\}_{i=1}^{N_t}, C_{red}, \tau)
\label{eq:boxed_grounding}
\end{equation}
where $I_{orig}$ is the original input image, $\{B_{t,i}\}_{i=1}^{N_t}$ is the list of bounding boxes at reasoning step $t$ where each $B_{t,i} = [x_{min}, y_{min}, x_{max}, y_{max}]$, $\textsc{Box}(\cdot)$ draws colored rectangular border lines for each specified regions of focus, $C_{red}$ is the border line color (\textit{i.e.,} red), and $\tau$ is the thickness of border lines.

\begin{figure}[H]
    \centering
    \includegraphics[width=1.0\textwidth]{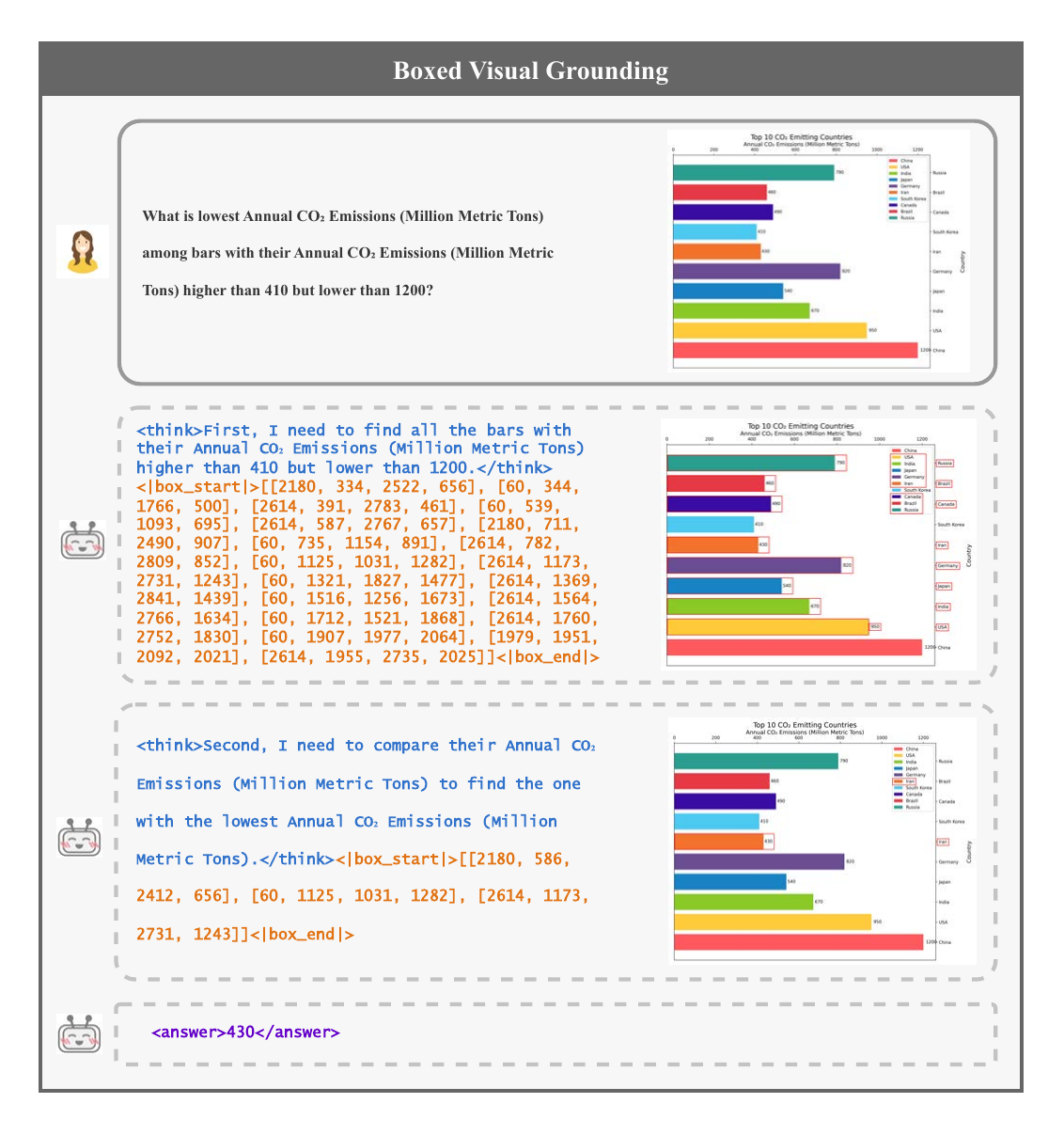}
    \caption{\textbf{Example of Boxed Visual Grounding.} The \textit{boxed} visual grounding method directly accentuates the regions of focus through semi-transparent yellow highlighting overlays.}
    \label{fig:A2_boxed_examples}
\end{figure}

\paragraph{Examples.}
\Fref{fig:A2_boxed_examples} shows an example for multi-step CoT reasoning with \textit{boxed} visual grounding.

\subsection{\textbf{Cropped}: Grounding Through Dynamic Visual Focus Zooming}
\label{appendix: visual grounding strategy - cropped}

\paragraph{Method.}
The \textit{cropped} grounding strategy localizes corresponding regions of focus by zooming in, extracting and presenting the focused sub-regions as separate zoomed images alongside the full chart. As reasoning evolves, different cropped regions are dynamically generated and presented, enabling detailed examination of the specific components relevant to each reasoning step.

Therefore, the cropping operation extracts sub-regions using array indexing based on the bounding box coordinates:

\vspace{-2.5mm}
\begin{align}
I'_{vis,t} &= \{I_{orig}, \{\textsc{Crop}(I_{orig}, B_{t,i})\}_{i=1}^{N_t}\} \label{eq:cropped_grounding} \\
\text{where } \textsc{Crop}(I_{orig}, B_{t,i}) &= I_{orig}[y_{min}:y_{max}, x_{min}:x_{max}] \label{eq:cropped_grounding_crop}
\end{align}
Here, $I_{orig}$ is the original chart image, $\{B_{t,i}\}_{i=1}^{N_t}$ is the list of bounding boxes at reasoning step $t$, and the model processes both the full context $I_{orig}$ and multiple zoomed crops $\{\textsc{Crop}(I_{orig}, B_{t,i})\}_{i=1}^{N_t}$ simultaneously.

\begin{figure}[H]
    \centering
    \includegraphics[width=1.0\textwidth]{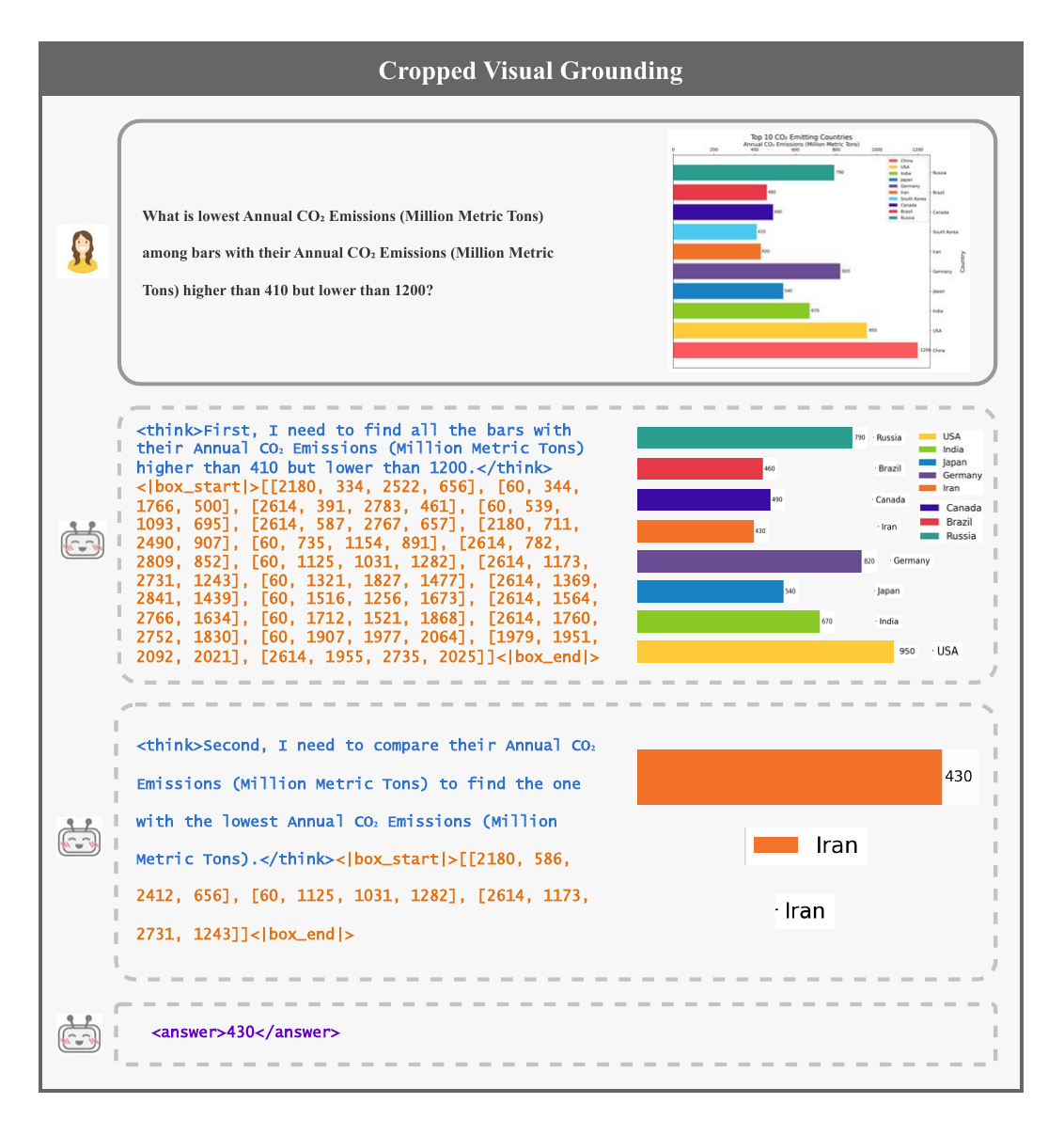}
    
    \vspace{-10pt}
    
    \caption{\textbf{Example of Cropped Visual Grounding.} The \textit{cropped} visual grounding method directly accentuates the regions of focus through semi-transparent yellow highlighting overlays.}
    \label{fig:A3_cropped_examples}
\end{figure}

\paragraph{Examples.}
\Fref{fig:A3_cropped_examples} shows an example for multi-step CoT reasoning with \textit{cropped} visual grounding.


\vspace{0pt}

\section{Generation Mode}
\label{appendix:generation_mode_examples}

% \xuehang{add examples for different generation modes}

\subsection{Mode \textbf{\modeA}}

\begin{figure}[H]
    \centering
    \includegraphics[width=1.0\textwidth]{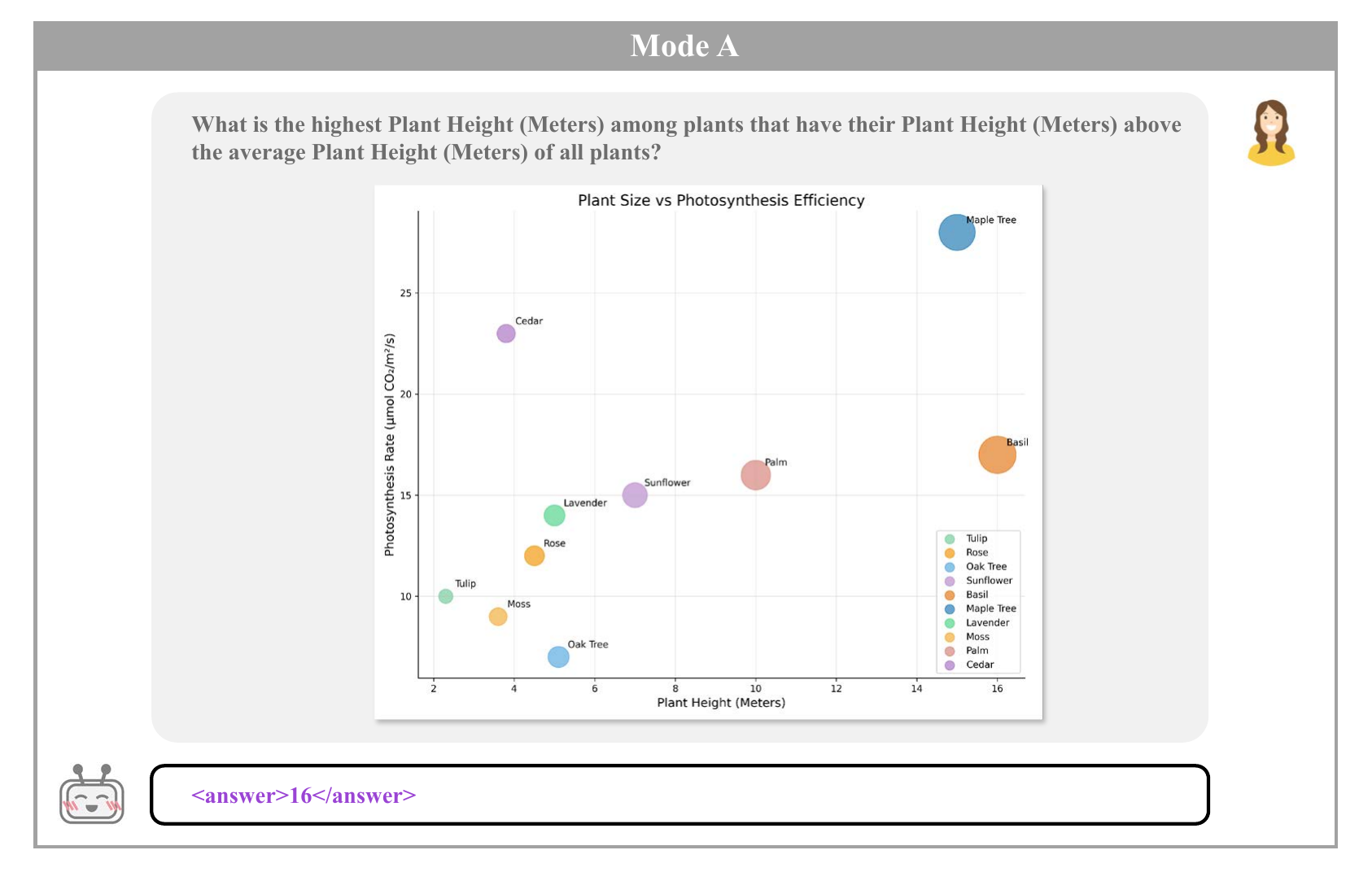}
    \vspace{-20pt}
    \caption{\textbf{Example of Generation Mode \modeA.} A CQA example resolved through generation mode \textbf{\modeA}.}
    \label{fig:generation_mode_A}
\end{figure}

\Fref{fig:generation_mode_A} shows an example of generation mode \textbf{\modeA}, where the model directly outputs the \textit{\textcolor{answer}{answer}} without intermediate reasoning and visual grounding.

\subsection{Mode \textbf{\modeVA}}

\Fref{fig:generation_mode_VA} shows an example of generation mode \textbf{\modeVA}, through which the model first generates its intermediate \textit{\textcolor{vision}{visual grounding}} via \textit{applied} grounding method, followed by its final \textit{\textcolor{answer}{answer}}.
For clarity, the input instructions for \textbf{\modeVA} generation are omitted in the figure.
To save space, the user’s intermediate responses are shown as smaller images on the right of each model response, corresponding to the model’s response on the left in each turn of the multi-step interaction.

\subsection{Mode \textbf{\modeRA}}

\begin{figure}[H]
    \centering
    \includegraphics[width=1.0\textwidth]{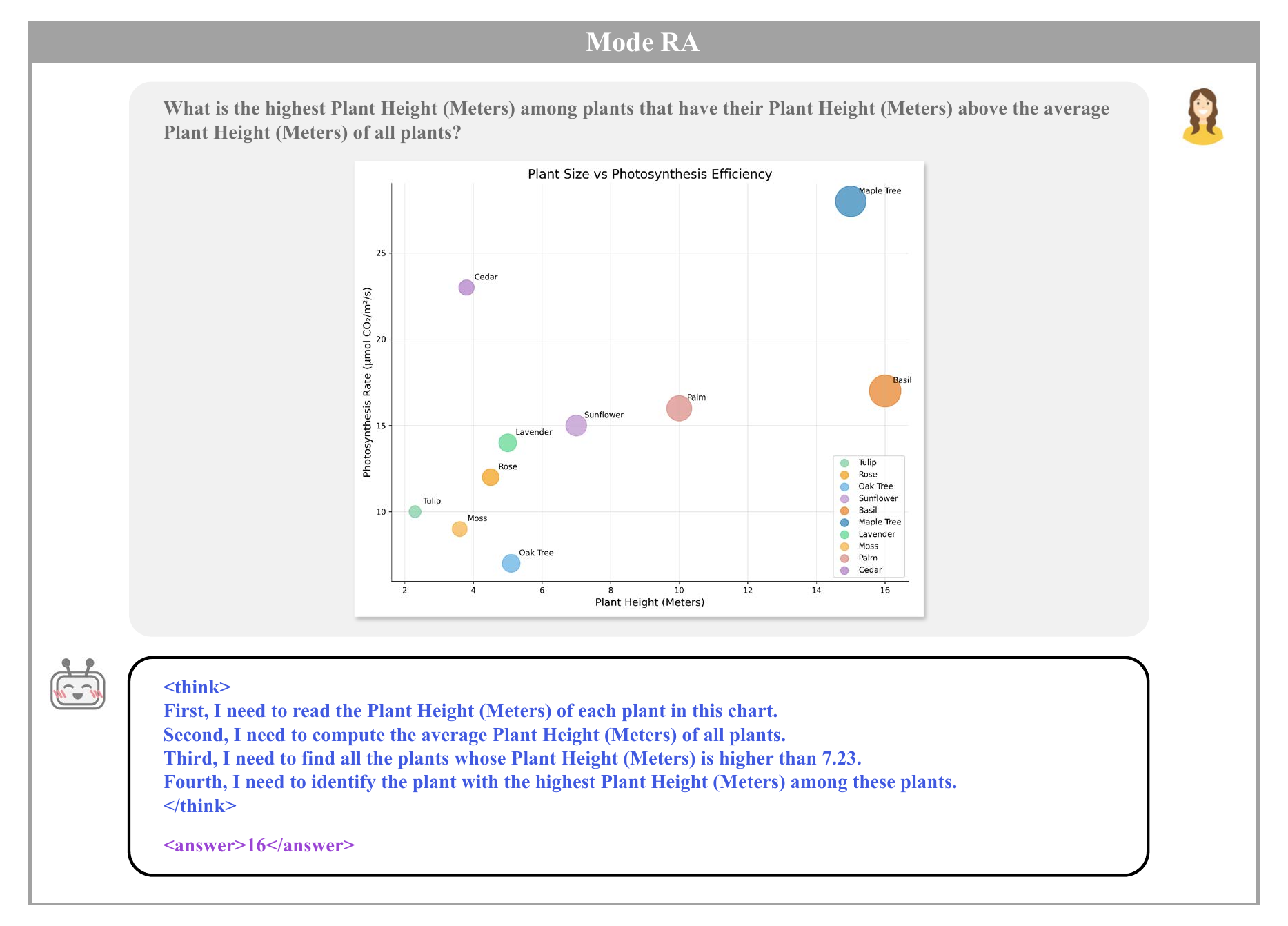}
    \vspace{-20pt}
    \caption{\textbf{Example of Generation Mode \modeRA.} A CQA example resolved through generation mode \textbf{\modeRA}.}
    \label{fig:generation_mode_RA}
\end{figure}

\Fref{fig:generation_mode_RA} shows an example of generation mode \textbf{\modeRA}, through which the model first generates its intermediate \textit{\textcolor{reason}{CoT reasoning}}, followed by its final \textit{\textcolor{answer}{answer}}.
The model is prompted to produce its CoT reasoning and final answer in a single-turn manner.
For clarity, the input instructions of CoT reasoning and answering for \textbf{\modeRA} mode are omitted in the figure.

\subsection{Mode \textbf{\modeRVA}}

\begin{figure}[H]
    \centering
    \includegraphics[width=1.0\textwidth]{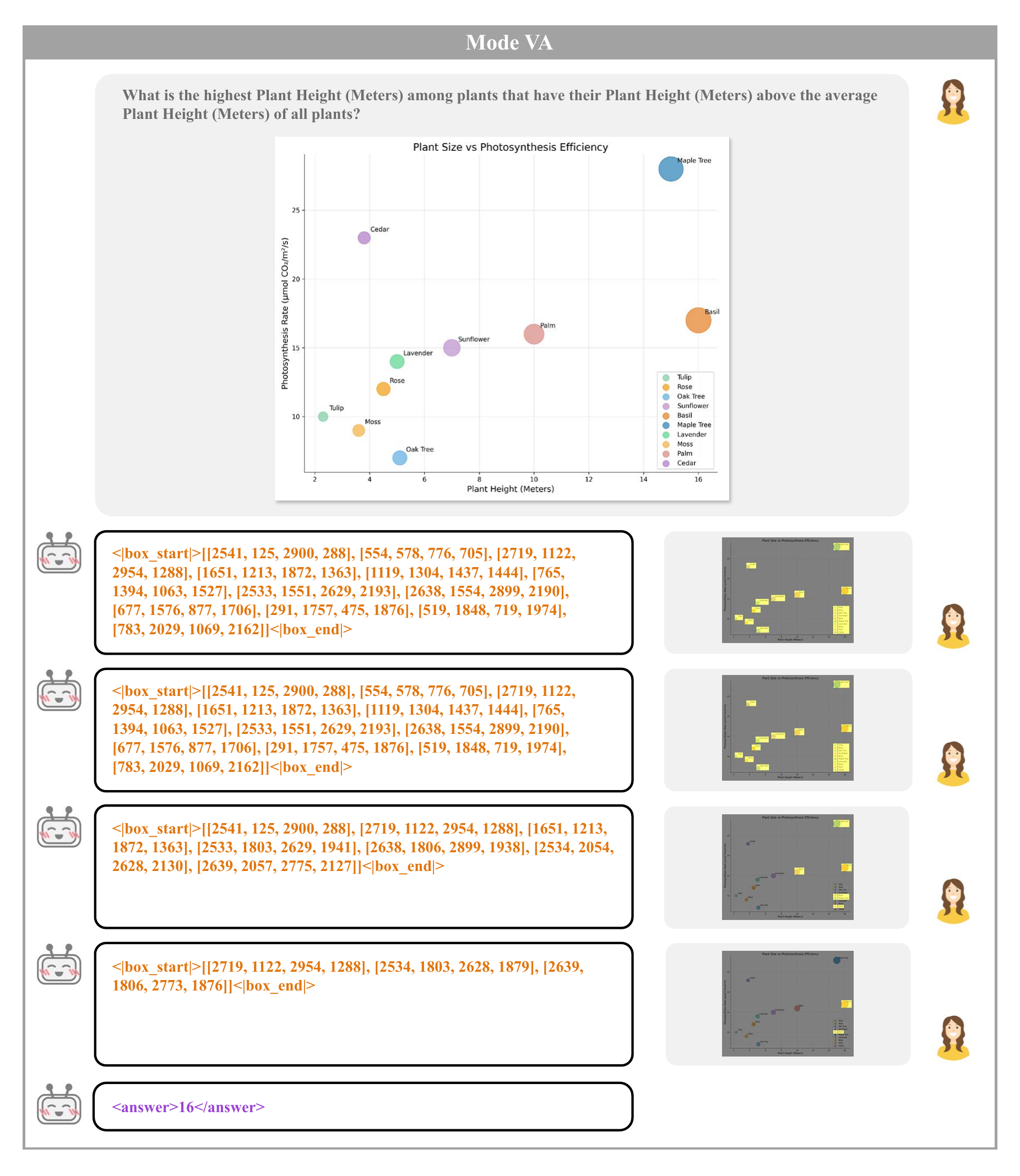}
    \vspace{-20pt}
    \caption{\textbf{Example of Generation Mode \modeVA.} A CQA example resolved through generation mode \textbf{\modeVA}.}
    \label{fig:generation_mode_VA}
\end{figure}

\begin{figure}[H]
    \centering
    \includegraphics[width=1.0\textwidth]{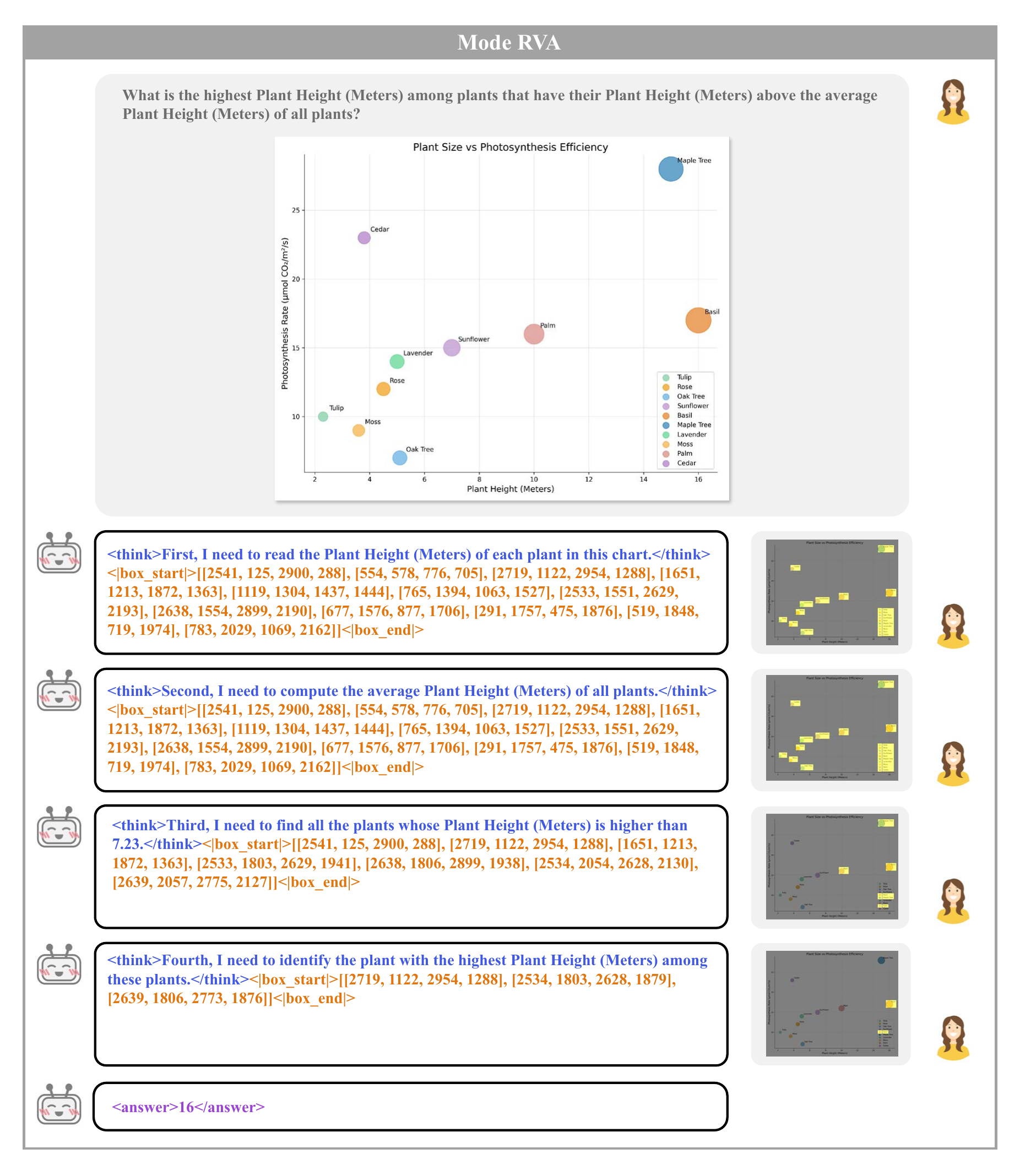}
    \vspace{-20pt}
    \caption{\textbf{Example of Generation Mode \modeRVA.} A CQA example resolved through generation mode \textbf{\modeRVA}.}
    \label{fig:generation_mode_RVA}
\end{figure}

\Fref{fig:generation_mode_RVA} shows an example of generation mode \textbf{\modeRVA}, where the model first produces its intermediate \textit{\textcolor{reason}{reasoning}} with \textit{\textcolor{vision}{visual grounding}}, followed by its final \textit{\textcolor{answer}{answer}}.
Similar to \Fref{fig:generation_mode_VA}, for clarity, the input instructions for \textbf{\modeRVA} mode generation are omitted in the figure.
To save space, the user’s intermediate responses are shown as smaller images on the right of each model response, corresponding to the model’s response on the left in each turn of the multi-step \modeRVA reasoning.

\section{Evaluation Metrics}
\label{appendix:sec:eval_metrics}

To elaborate more details in \S\ref{subsec:eval_evaluation_metrics}, our evaluation incorporates multiple complementary metrics to assess different aspects of model performance.

\subsection{Evaluation of Answers}
\label{appendix:subsec:eval_metric:answer}

In pursuit of accurate evaluation on multimodal datasets that contain both multi-choice and free-form responses, we compute answer accuracy by comparing model outputs with their corresponding ground-truth answers.
Aiming for more comprehensive assessment, we employ two complementary evaluation approaches: MLLM-as-judge for semantic understanding and rule-based evaluation for systematic accuracy measurement.
The overall accuracy score for each dataset is calculated as the mean accuracy across all test samples.

\begin{figure}[H]
    \centering
    \includegraphics[width=0.9\textwidth]{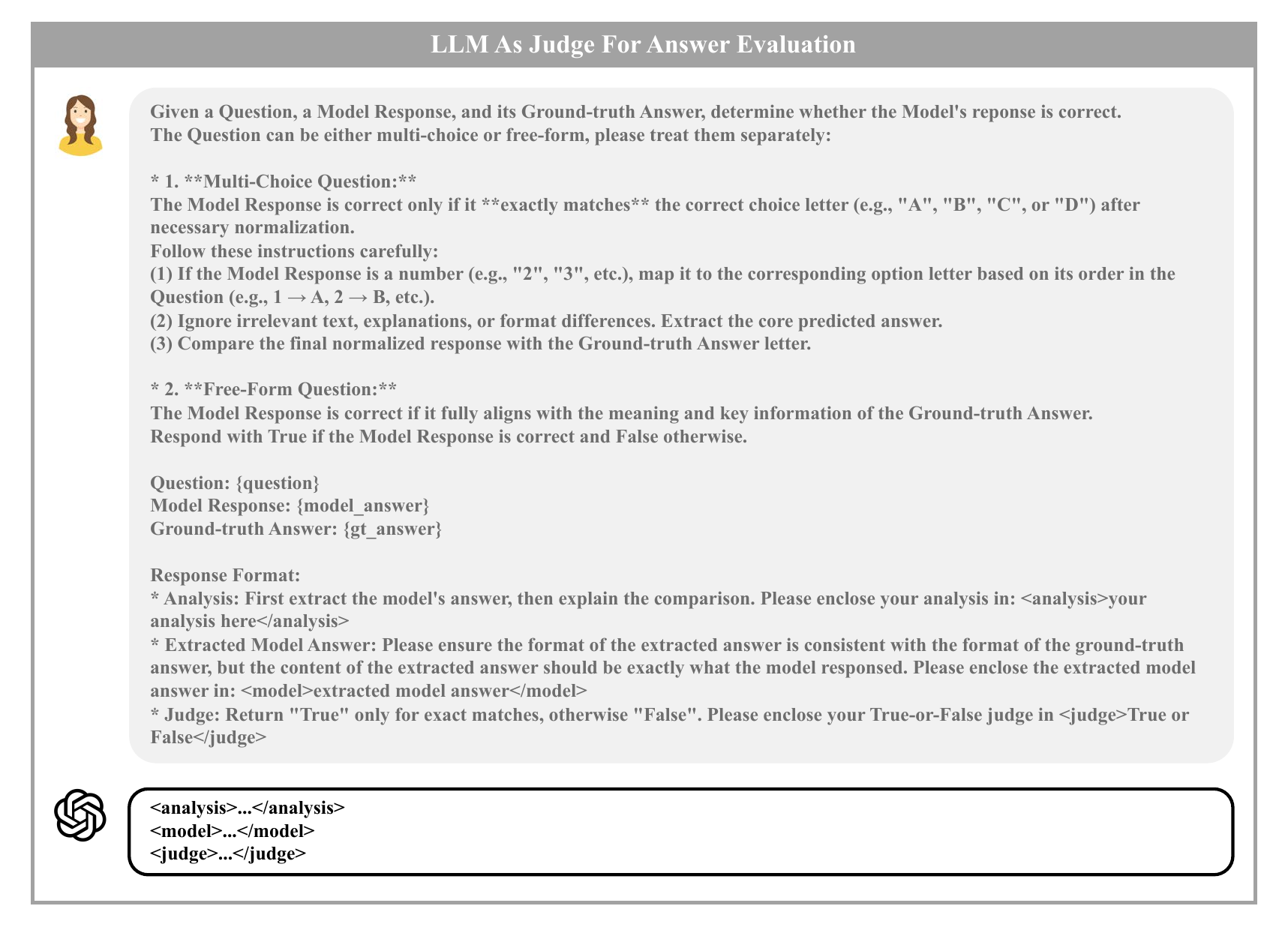}
    \vspace{-10pt}
    \caption{\textbf{LLM-As-Judge For Answer Evaluation.} We employ GPT-4.1-mini as the judge to assess model answer accuracy using the prompt shown in this figure.}
    \label{fig:appendix_llm_as_judge_answer_evaluation}
\end{figure}

\paragraph{MLLM-Based Answer Evaluation.}
For MLLM-based answer evaluation, we employ GPT-4.1-mini as the \textit{judge}, guided by the prompt shown in \Fref{fig:appendix_llm_as_judge_answer_evaluation}.
Each model response undergoes MLLM-as-judge evaluation to extract the model answer content, ensuring consistent comparison with ground truth.
The \textit{judge} performs a \textit{True}-or-\textit{False} assessment by evaluating whether the model response semantically matches the ground truth, accounting for variations in phrasing and presentation while maintaining semantic equivalence.

\paragraph{Rule-Based Answer Evaluation.}
To mitigate potential biases introduced by using MLLMs as judges (\S\ref{subsec:eval_evaluation_metrics}), we complement the MLLM-as-judge approach with a systematic rule-based evaluation (Algorithm~\ref{algorithm:rule_based_answer_eval}).
This rule-based method assesses answer accuracy through predefined parsing and judgment rules, incorporating both strict and relaxed error tolerance through four range criteria:
\textit{absolute accuracy} ($acc\textsc{@}0.0$) and three progressively relaxed thresholds ($acc\textsc{@}0.05$, $acc\textsc{@}0.1$, $acc\textsc{@}0.2$).

\vspace{20pt}
\begin{algorithm}[H]
\small
\caption{Rule-Based Answer Evaluation with Tolerance Ranges}
\label{algorithm:rule_based_answer_eval}
\begin{algorithmic}[1]
\REQUIRE Ground truth answer $gt$, predicted answer $pred$, choices $C$ (optional), tolerance ranges $R = \{0.0, 0.05, 0.1, 0.2\}$
\ENSURE Accuracy scores $acc@r$ for each $r \in R$

\STATE $gt \leftarrow$ \textsc{Clean}$(gt)$, $pred \leftarrow$ \textsc{Clean}$(pred)$
\STATE $answer\_type \leftarrow$ \textsc{DetectType}$(gt)$

\IF{$gt = \emptyset$ \AND $pred = \emptyset$}
    \RETURN $acc@r = 1.0$ for all $r \in R$
\ENDIF

\IF{$answer\_type = $ "multi-choice"}
    \STATE $gt\_list \leftarrow$ \textsc{ParseChoices}$(gt, C)$
    \STATE $pred\_list \leftarrow$ \textsc{ParseChoices}$(pred, C)$
    \IF{$gt\_list = pred\_list$}
        \RETURN $acc@r = 1.0$ for all $r \in R$
    \ELSE
        \STATE $match\_rate \leftarrow \frac{|gt\_list \cap pred\_list|}{|gt\_list|}$
        \FOR{$r \in R$}
            \STATE $acc@r \leftarrow \mathbf{1}[match\_rate \geq r]$
        \ENDFOR
    \ENDIF

\ELSIF{$answer\_type \in $ \{int, float\}}
    \STATE $gt\_num \leftarrow$ \textsc{ExtractNumber}$(gt)$
    \STATE $pred\_num \leftarrow$ \textsc{Extract}$(pred)$
    \STATE $acc@0.0 \leftarrow$ \textsc{ExtractNumber}$(pred\_num, gt\_num)$
    \FOR{$r \in \{0.05, 0.1, 0.2\}$}
        \STATE $lower \leftarrow gt\_num \times (1-r)$
        \STATE $upper \leftarrow gt\_num \times (1+r)$
        \STATE $acc@r \leftarrow \mathbf{1}[lower \leq pred\_num \leq upper]$
    \ENDFOR

\ELSE
    \STATE $exact\_match \leftarrow$ \textsc{GradeAnswer}$(pred.lower(), gt.lower())$
    \STATE $substring\_match \leftarrow \mathbf{1}[|gt| > 5 \text{ and } gt.lower() \in pred.lower()]$
    \STATE $acc@r \leftarrow max(exact\_match, substring\_match)$ for all $r \in R$
\ENDIF

\RETURN $acc@r$ for all $r \in R$
\end{algorithmic}
\end{algorithm}

\subsection{Evaluation of Reasoning}
\label{appendix:subsec:eval_metric:reasoning}
\begin{figure}[H]
    \centering
    \includegraphics[width=0.9\textwidth]{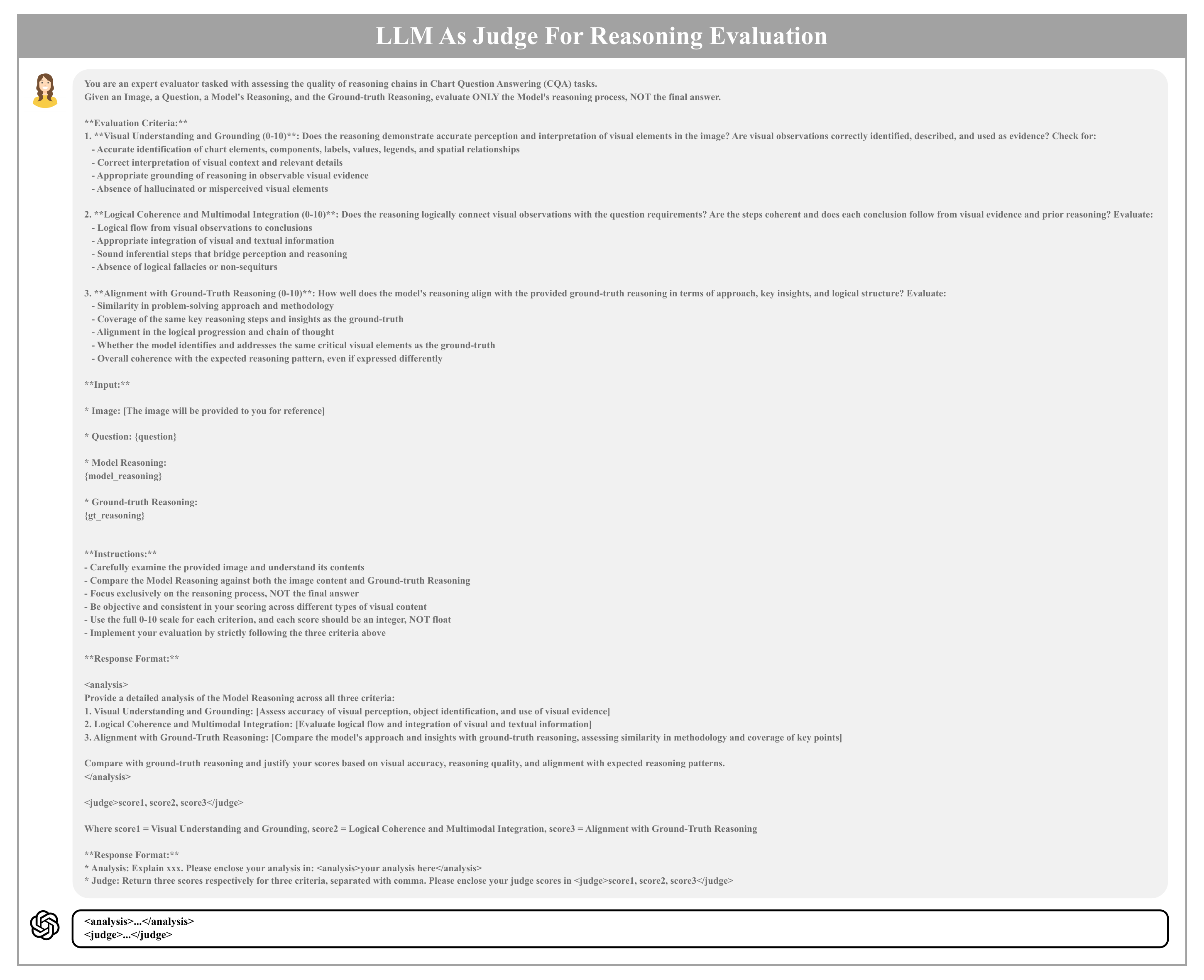}
    \vspace{-10pt}
    \caption{\textbf{LLM-as-Judge For Reasoning Evaluation.} We employ GPT-4.1-mini as the judge to evaluate model reasoning using the prompt shown in this figure. Prompt is restricted to smaller sizes to save space.}
    \label{fig:appendix_llm_as_judge_reasoning_evaluation}
\end{figure}

To comprehensively evaluate model reasoning, we implement both micro- and macro-level assessments (\S\ref{subsec:eval_evaluation_metrics}).
Our micro-level evaluation relies on five metrics (\Eref{eq:rouge} -~\ref{eq:cosine}), providing the semantic similarity assessment of model reasoning. The final score ($acc\textsc{@}mic$) is the average across all five metrics.
At the macro level ($acc\textsc{@}mac$), we leverage GPT-4.1-mini as the \textit{judge}, which rates the quality of model reasoning on a $0-10$ scale based on three criteria:
(1) \textit{visual understanding and grounding},
(2) \textit{logical coherence and multimodal integration}, and
(3) \textit{alignment with ground-truth reasoning}.
While micro-level evaluation focuses on fine-grained similarity between ground-truth and model reasoning, macro evaluation provides a holistic judgment of reasoning quality through MLLM-as-judge (\Fref{fig:appendix_llm_as_judge_reasoning_evaluation}).

\vspace{10pt}

\vspace{10pt}

\paragraph{Micro-Evaluation: Reasoning Similarity.}
We employ five metrics to measure the semantic similarity between ground-truth and model reasoning, including \textsc{ROUGE-L} (\Eref{eq:rouge}), \textsc{BLEU} (\Eref{eq:bleu}), \textsc{METEOR} (\Eref{eq:meteor}), \textsc{BERTScore} (\Eref{eq:bertscore}), and \textsc{Cosine Similarity} (\Eref{eq:cosine}).

%%%%%%%%%%  ROUGE & BLEU  %%%%%%%%%%
\vspace{-3.6mm}
\begin{align}
\textsc{ROUGE} &= \textsc{ROUGE-L} = \frac{2 \cdot P_{\textit{lcs}} \cdot R_{\textit{lcs}}}{P_{\textit{lcs}} + R_{\textit{lcs}}} \label{eq:rouge} \\
\textsc{BLEU} &= \textsc{BLEU-4} = \textsc{BP} \cdot \exp\left(\sum_{n=1}^{4} w_n \log p_n\right) \label{eq:bleu}
\end{align}
\vspace{-3.6mm}

where $P_{\textit{lcs}}$ and $R_{\textit{lcs}}$ are precision and recall of longest common subsequences.

%%%%%%%%%%  F1  %%%%%%%%%%
% \begin{equation}
% \textsc{F1} = \frac{2 \cdot P_r \cdot R_r}{P_r + R_r}
% \label{eq:f1}
% \end{equation}

%%%%%%%%%%  BERT Score  %%%%%%%%%%
\begin{equation}
\textsc{BERTScore} = \frac{1}{|\mathcal{S}_p|} \sum_{s_i \in \mathcal{S}_p} \max_{s_j \in \mathcal{S}_g} \frac{\mathbf{v}_i^T \mathbf{v}_j}{|\mathbf{v}_i| |\mathbf{v}_j|}
\label{eq:bertscore}
\end{equation}
where $\mathcal{S}_p$ and $\mathcal{S}_g$ are the sets of predicted and ground-truth reasoning tokens respectively, and $\mathbf{v}_i, \mathbf{v}_j$ are their corresponding BERT contextual embeddings.

%%%%%%%%%%  METEOR  %%%%%%%%%%
\begin{equation}
\textsc{METEOR} = \frac{(1 + \eta_1) \cdot P_r \cdot R_r}{\eta_1 \cdot P_r + R_r} \cdot \left(1 - \eta_2 \cdot \left(\frac{c}{u_m}\right)^{\eta_3}\right)
\label{eq:meteor}
\end{equation}

%%%%%%%%%%  Cosine Similarity  %%%%%%%%%%
\begin{equation}
\textsc{COSINE} = \frac{\mathbf{e}_p \cdot \mathbf{e}_g}{|\mathbf{e}_p| |\mathbf{e}_g|}
\label{eq:cosine}
\end{equation}

where $P_r$ and $R_r$ are precision and recall of reasoning tokens, $u_m$ is the number of matched unigrams, $c$ is the number of chunks, and we define $\eta_1 = 0.9$, $\eta_2 = 0.5$, and $\eta_3 = 3$ as hyperparameters controlling the weight of recall, penalty magnitude, and penalty sharpness, respectively; and $\mathbf{e}_p$ and $\mathbf{e}_g$ are the embedding vectors of predicted and ground-truth reasoning steps respectively.

\paragraph{Macro-Evaluation: Reasoning Quality.}
The quality of model reasoning is evaluated through MLLM-as-judge assessment.
Specifically, we employ GPT-4.1-mini as the \textit{judge}, guided by the prompt shown in \Fref{fig:appendix_llm_as_judge_reasoning_evaluation}, to assign a quality score on a $0-10$ scale based on three criteria, including \textit{visual understanding and grounding}, \textit{logical coherence and multimodal integration}, and \textit{alignment with ground-truth reasoning}.
The final quality score for each dataset is calculated as the mean score across all test samples.

\begin{itemize}
    \item \textbf{Criterion 1: Visual Understanding and Grounding.} Reasoning accuracy in identifying, interpreting, and grounding reasoning in visual elements, meanwhile without introducing hallucinated details
    
    \item \textbf{Criterion 2: Logical Coherence and Multimodal Integration.} Logical progression throughout the entire reasoning chain, with appropriate integration of multimodal information.
    
    \item \textbf{Criterion 3: Alignment with Ground-Truth Reasoning.}
    Consistency with ground-truth reasoning, especially in terms of problem-solving approach, key insights, and logical structure, even if expressed differently.
\end{itemize}

\subsection{Evaluation of Visual Grounding}
\label{appendix:subsec:eval_metric:visual_grounding}

As introduced in (\S\ref{subsec:eval_evaluation_metrics}), we employ two IoU variants, \textsc{cIoU} (\Eref{eq:ciou}) and \textsc{gIoU} (\Eref{eq:giou}), as the primary evaluation metrics for visual grounding assessment.

%%%%%%%%%%  cIoU & gIoU  %%%%%%%%%%
\vspace{-3.6mm}
\begin{align}
\textsc{CIoU} &= \textsc{IoU} - \frac{\rho^2(c_p, c_g)}{d^2} \label{eq:ciou} \\
\textsc{GIoU} &= \textsc{IoU} - \frac{|A_c - A_u|}{A_c} \label{eq:giou}
\end{align}
\vspace{-3.6mm}

where $\rho^2(c_p, c_g)$ is the squared distance between predicted and ground-truth centroids, $d$ is the diagonal of the enclosing box, $A_c$ is the enclosing area, and $A_u$ is the union area.

\subsection{Evaluation Mode}
\label{appendix:subsec:evaluation_mode}

Furnishing models with the capabilities to reason through dynamic visual grounding, we employ different generation modes (\S\ref{subsec:exp_setup}) to support comparable evaluations:

\vspace{-5pt}
\begin{itemize}
    \item \textbf{Mode \modeA}: \textit{Answer-Only} mode where MLLMs are prompted to directly generate the final answer.
    \item \textbf{Mode \modeRA:} \textit{Reason-Answer} mode where MLLMs first go through the intermediate reasoning process, followed by the final answer.
    \item \textbf{Mode \modeVA:} \textit{Vision-Answer} mode where MLLMs first generate their visual grounding coordinates, followed by the final answer.
    \item \textbf{Mode \modeRVA:} \textit{Reason-Vision-Answer} mode where MLLMs first go through the reasoning process with dynamic visual grounding, and then generate the final answer.
\end{itemize}

\section{In-Depth Analysis}
\label{appendix:sec:in_depth_analysis}

\subsection{Implementation Details.}
\label{appendix:subsec:implementation_details}
We train each model for \textsc{3} epochs with an initial learning rate $lr=1e-4$ using \texttt{cosine} scheduler.
The ratio of training and validation is set to \texttt{train:val=9:1}.
Employing two NVIDIA 80G H100 GPUs, our model training is powered by LoRA for memory efficiency.
For hyperparameter settings, \textit{Stage I} supervises visual grounding only ($\lambda_V = 1.0$, $\lambda_R = 0.0$, $\lambda_A = 0.0$), and \textit{Stage II} supervises reasoning, grounding, and answer jointly (Eq.~\ref{equation:stage2_loss}).
Implementing different generation modes, \textcolor{reason}{\textit{reasoning} (\modeR)} is enclosed within \texttt{\textless think\textgreater \textless /think\textgreater}, 
\textcolor{vision}{\textit{visual grounding} (\modeV)} is enclosed within \texttt{\textless|box\_start|\textgreater \textless|box\_end|\textgreater} for explicit visual grounding while represented as \texttt{\textless GROUND\textgreater} in implicit visual ground (\S\ref{appendix:subsec:explicit_vs_implicit}), 
and the \textcolor{answer}{\textit{final answer} (\modeA)} is enclosed within \texttt{\textless answer\textgreater \textless /answer\textgreater}.

\subsection{Grounding Method \& Computation Cost}
\label{appendix:subsec:computation_overhead}

Employing zoom-in visual grounding, the \textit{cropped} grounding method requires substantially larger memory at the same resolution.
To mitigate this cost, we reduce the training resolution of \textit{cropped} grounding to $128 \times 128$, thereby maintaining a comparable computational overhead.
Despite the resolution degradation, reasoning with \textbf{cropped} visual grounding achieves notably higher accuracy than the baseline (up to $4.72\%$ improvement on \oursdata) and performs competitively with the other two grounding methods (\Tref{tab:main_result_on_ours_data}). These results highlight the effectiveness of zoom-in visual enhancement, albeit at the expense of increased computational cost when aiming for higher performance.

\begin{table}[htbp]

\vspace{10pt}

\centering
\renewcommand{\arraystretch}{1.5}
\begin{tabularx}{\textwidth}{>{\centering\arraybackslash}p{0.12\textwidth}|>{\centering\arraybackslash}p{0.12\textwidth}>{\centering\arraybackslash}p{0.06\textwidth}>{\centering\arraybackslash}p{0.06\textwidth}>{\centering\arraybackslash}p{0.12\textwidth}>{\centering\arraybackslash}p{0.18\textwidth}>{\centering\arraybackslash}p{0.12\textwidth}}
\toprule
\textbf{Method} & \textbf{Resolution} & \textbf{$D_{max}$} & \textbf{$T_{max}$} & \textbf{SFT} & \textbf{RL} & \textbf{Inference} \\
\midrule
\textbf{Applied} & $448 \times 448$ & $4$ & $5$ & $2 \times 80GB$ & $\geq 8 \times 80GB$ & $1 \times 80GB$ \\
\textbf{Boxed} & $448 \times 448$ & $4$ & $5$ & $2 \times 80GB$ & $\geq 8 \times 80GB$ & $1 \times 80GB$ \\
\textbf{Cropped} & $128 \times 128$ & $4$ & $5$ & $2 \times 80GB$ & $\geq 8 \times 80GB$ & $1 \times 80GB$ \\
\bottomrule
\end{tabularx}

\caption{\textbf{The Computation \& Configuration Of Different Grounding Method.} This table summarizes the visual computation requirements and parameter configuration.}
\label{tab:grounding_computation_and_config}

\end{table}

\subsection{The Role of Visual Grounding: From Extrinsic Assistance To Intrinsic Abilities}

A critical finding from our experiments reveals the fundamental distinction between the utility of multi-step visual reasoning during training versus inference (\Tref{tab:main_results_RVA}). 
While incorporating visual grounding in the training process significantly enhances models' intrinsic visual reasoning capabilities, directly applying the same multi-step approach during inference can paradoxically degrade performance due to error accumulation (\Fref{appendix:subsec:failure_cases} \& \S\ref{appendix:inference_in_RAV_mode}).

\textbf{Training Benefits of Visual Grounding.}
Our curriculum learning approach with visual grounding supervision effectively teaches models to develop stronger intrinsic representations for chart understanding.
By learning to align reasoning steps with visual focuses during training, models internalize the ability to focus on relevant image components, leading to improved performance even when generating direct answers without explicit visual grounding steps.

\textbf{Inference Challenges with Multi-Turn Visual Grounding.}
On the other hand, when models are required to explicitly generate visual grounding coordinates during inference (i.e., \textit{Mode} \modeRVA), performance degrades in comparison with direct answer generation (\textit{Mode} \modeA) and reasoning without grounding (\textit{Mode} \modeRA). This degradation stems from two primary factors: 
(1) \textit{Cumulative grounding errors}: Inaccurate bounding box predictions in early reasoning steps propagate and compound errors in subsequent steps;
(2) \textit{Reasoning-grounding misalignment}: Discrepancies between intended visual focus and actual predicted coordinates lead to reasoning based on incorrect visual regions.

\textbf{Power of Intrinsic Visual Reasoning Capabilities.}
Results in \Tref{tab:main_results_RVA} demonstrate that visual grounding serves as an effective \textit{training signal} rather than an \textit{inference mechanism}. Our curriculum learning with visual supervision enables models to learn better intrinsic visual-textual alignments, which manifest as improved performance in direct answer generation \Tref{tab:main_result_on_ours_data}.
However, explicitly requiring visual grounding during inference introduces additional complexity and error sources that outweigh the potential benefits.
Nevertheless, compared with baselines, our finetuned models manage to achieve remarkably higher performance in not only \textit{Mode} \modeA, but also \textit{Modes} \modeRA, \modeVA, and \modeRVA.
% \xuehang{add numbers from the table}

\subsection{Visual Grounded Reasoning via Reinforcement Learning}
\label{appendix:subsec:rl_vs_sft}

To further validate the generality of \ours, we extend our two-stage training framework to reinforcement learning (RL), reformulating training objectives as specialized reward signals that directly incentivize accurate visual grounding, faithful reasoning, and correct answer generation.

\paragraph{Stage I: Visual Grounding.}
In \textit{Stage I}, we define visual grounding reward $\mathcal{R}_V$ that evaluates the geometric quality of the predicted visual focus $V_t$ against the ground-truth focus region $V_t^*$ using CIoU (\Eref{eq:ciou}) and GIoU (\Eref{eq:giou}):

\begin{equation}
    \mathcal{R}_V(V_t, V_t^*) = \lambda_{\mathrm{CIoU}}\,\mathcal{R}_{\mathrm{CIoU}}(V_t, V_t^*) + \lambda_{\mathrm{GIoU}}\,\mathcal{R}_{\mathrm{GIoU}}(V_t, V_t^*),
    \label{equation:rl_grounding_reward}
\end{equation}

where $\lambda_{\mathrm{CIoU}} = \lambda_{\mathrm{GIoU}} = 0.5$. As such, \textit{Stage I} training objective becomes:
\begin{equation}
    \mathcal{R}^{(S1)} = \sum_{t=1}^{T} \mathcal{R}_V(V_t, V_t^*).
    \label{equation:rl_stage1}
\end{equation}

\paragraph{Stage II: Interleaved Visual Reasoning Reward.}
In \textit{Stage II}, the training objective extends to the full interleaved visual grounded reasoning chain with a composite reward:

\begin{equation}
    \mathcal{R}^{(S2)} = \lambda_R \sum_{t=1}^{T} \mathcal{R}_R(R_t, R_t^*) + \lambda_V \sum_{t=1}^{T} \mathcal{R}_V(V_t, V_t^*) + \lambda_A\,\mathcal{R}_A(A, A^*) + \lambda_F\,\mathcal{R}_F,
    \label{equation:rl_stage2}
\end{equation}

where $\lambda_R$, $\lambda_V$, $\lambda_A$, and $\lambda_F$ are weighting coefficients balancing the four reward components. Each component is defined as follows:

\textbf{Reasoning reward} $\mathcal{R}_R(R_t, R_t^*)$ measures the semantic similarity between the predicted reasoning step $R_t$ and its ground-truth counterpart $R_t^*$ as a rubric-based aggregate of \textsc{ROUGE-L} (\Eref{eq:rouge}), \textsc{BERTScore} (\Eref{eq:bertscore}), and \textsc{Cosine Similarity} (\Eref{eq:cosine}):

\vspace{-3.6mm}
\begin{align}
\textsc{Similarity} = \lambda_\textit{rouge} \cdot \textsc{rouge} + \lambda_\textit{bert} \cdot \textsc{bert} + \lambda_\textit{cosine} \cdot \textsc{cosine}
\label{eq:rl_reason_reward}
\end{align}
\vspace{-3.6mm}

where $\lambda_\textit{rouge}$, $\lambda_\textit{bert}$, and $\lambda_\textit{cosine}$ are  weighting coefficients respectively set to 0.3, 0.4, 0.3 during \textit{Stage II} training.

\textbf{Grounding reward} $\mathcal{R}_V(V_t, V_t^*)$ evaluates geometric alignment of the predicted visual focus at each reasoning step, as defined in \Eref{equation:rl_grounding_reward}.

\textbf{Answer reward} $\mathcal{R}_A(A, A^*)$ is a binary exact-match signal that returns $1$ if the predicted final answer $A$ matches the ground-truth $A^*$ and $0$ otherwise.

\textbf{Format reward} $\mathcal{R}_F$ is a binary compliance signal that returns $1$ if the model output adheres to the required structured format, ensuring output parsability throughout training.

% ------ RL Results ------

\begin{table}[t!]

\vspace{-28pt}

\centering
\resizebox{\textwidth}{!}{%
\begin{tabular}{c | c c c c c | c c c c c | c c c c c}
\toprule
\multirow{2}{*}{\textbf{Model}} & \multicolumn{5}{c|}{\textbf{Level 1}} & \multicolumn{5}{c|}{\textbf{Level 2}} & \multicolumn{5}{c}{\textbf{Level 3}} \\
\cmidrule(lr){2-6} \cmidrule(lr){7-11} \cmidrule(lr){12-16}
& $\boldsymbol{\textsc{@}M}$ & $\boldsymbol{\textsc{@}0.0}$ & $\boldsymbol{\textsc{@}0.05}$ & $\boldsymbol{\textsc{@}0.1}$ & $\boldsymbol{\textsc{@}0.2}$ & $\boldsymbol{\textsc{@}M}$ & $\boldsymbol{\textsc{@}0.0}$ & $\boldsymbol{\textsc{@}0.05}$ & $\boldsymbol{\textsc{@}0.1}$ & $\boldsymbol{\textsc{@}0.2}$ & $\boldsymbol{\textsc{@}M}$ & $\boldsymbol{\textsc{@}0.0}$ & $\boldsymbol{\textsc{@}0.05}$ & $\boldsymbol{\textsc{@}0.1}$ & $\boldsymbol{\textsc{@}0.2}$ \\

\midrule

\rowcolor{baseline}
\multicolumn{16}{c}{\textbf{Baseline}} \\

\midrule

\textbf{Baseline} & 45.25 & 43.52 & 51.54 & 56.25 & 61.54 & 22.75 & 22.86 & 31.14 & 37.21 & 45.36 & 16.18 & 16.00 & 21.89 & 25.82 & 31.71 \\

\midrule

\rowcolor{finetuned}
\multicolumn{16}{c}{\textbf{Ours (RL)}} \\

\midrule

\textbf{Stage II} & \gaincell{45.25}{53.64} & \gaincell{43.52}{51.71} & \gaincell{51.54}{57.43} & \gaincell{56.25}{61.36} & \gaincell{61.54}{65.00} & \gaincell{22.75}{24.61} & \gaincell{22.86}{25.39} & \gaincell{31.14}{35.21} & \gaincell{37.21}{41.04} & \gaincell{45.36}{51.57} & \gaincell{16.18}{16.21} & \gaincell{16.00}{18.86} & \gaincell{21.89}{25.25} & \gaincell{25.82}{29.32} & \gaincell{31.71}{33.96} \\

\addlinespace

\textbf{Stage I+II} & \gaincell{45.25}{55.14} & \gaincell{43.52}{53.23} & \gaincell{51.54}{61.37} & \gaincell{56.25}{67.38} & \gaincell{61.54}{70.54} & \gaincell{22.75}{26.25} & \gaincell{22.86}{28.46} & \gaincell{31.14}{40.35} & \gaincell{37.21}{48.05} & \gaincell{45.36}{57.94} & \gaincell{16.18}{19.07} & \gaincell{16.00}{22.53} & \gaincell{21.89}{27.43} & \gaincell{25.82}{31.43} & \gaincell{31.71}{35.58} \\

\midrule

\rowcolor{finetuned}
\multicolumn{16}{c}{\textbf{Ours (SFT)}} \\

\midrule

\textbf{Stage II} & \gaincell{45.25}{54.21} & \gaincell{43.52}{51.21} & \gaincell{51.54}{59.50} & \gaincell{56.25}{65.00} & \gaincell{61.54}{68.50} & \gaincell{22.75}{25.86} & \gaincell{22.86}{27.14} & \gaincell{31.14}{39.18} & \gaincell{37.21}{47.25} & \gaincell{45.36}{55.18} & \gaincell{16.18}{16.86} & \gaincell{16.00}{17.32} & \gaincell{21.89}{23.93} & \gaincell{25.82}{28.61} & \gaincell{31.71}{33.11} \\

\addlinespace

\textbf{Stage I+II} & \gaincell{45.25}{60.71} & \gaincell{43.52}{57.21} & \gaincell{51.54}{63.71} & \gaincell{56.25}{69.57} & \gaincell{61.54}{73.79} & \gaincell{22.75}{30.11} & \gaincell{22.86}{29.61} & \gaincell{31.14}{41.82} & \gaincell{37.21}{48.87} & \gaincell{45.36}{57.46} & \gaincell{16.18}{17.98} & \gaincell{16.00}{18.21} & \gaincell{21.89}{24.25} & \gaincell{25.82}{29.07} & \gaincell{31.71}{34.50} \\

\midrule

\rowcolor{finetuned}
\multicolumn{16}{c}{\textbf{Ours (SFT + RL)}} \\

\midrule

\textbf{Stage II} & \gaincell{45.25}{55.86} & \gaincell{43.52}{53.64} & \gaincell{51.54}{60.79} & \gaincell{56.25}{66.43} & \gaincell{61.54}{69.29} & \gaincell{22.75}{27.96} & \gaincell{22.86}{28.79} & \gaincell{31.14}{40.75} & \gaincell{37.21}{48.64} & \gaincell{45.36}{57.25} & \gaincell{16.18}{20.64} & \gaincell{16.00}{21.39} & \gaincell{21.89}{26.79} & \gaincell{25.82}{30.25} & \gaincell{31.71}{34.93} \\

\addlinespace

\textbf{Stage I+II} & \gaincell{45.25}{62.29} & \gaincell{43.52}{59.79} & \gaincell{51.54}{65.86} & \gaincell{56.25}{72.21} & \gaincell{61.54}{73.79} & \gaincell{22.75}{29.36} & \gaincell{22.86}{29.74} & \gaincell{31.14}{42.57} & \gaincell{37.21}{50.39} & \gaincell{45.36}{59.39} & \gaincell{16.18}{22.43} & \gaincell{16.00}{23.39} & \gaincell{21.89}{29.25} & \gaincell{25.82}{32.43} & \gaincell{31.71}{36.89} \\

% \addlinespace

\bottomrule
\end{tabular}%
}

\caption{\textbf{\ours with Reinforcement Learning on \oursdata.} Employing Qwen2.5-VL-3B as the base model, we evaluate the performance of its baseline and finetuned versions on \oursdata across three curriculum levels via five accuracy evaluation metrics $\boldsymbol{acc\textsc{@}}X$ where $X$ is MLLM (abbreviated as $\boldsymbol{\textsc{@}}M$) or ranges (\Sref{subsec:eval_evaluation_metrics}). All settings use \ours{}$@$Applied throughout training and inference.}
\label{tab:rl_main_result_on_ours_data}

\end{table}

\subsection{Explicit vs. Implicit Visual Grounded Reasoning}
\label{appendix:subsec:explicit_vs_implicit}

We introduce two designs for integrating visual grounding into the multi-step reasoning framework: \textit{explicit} and \textit{implicit} visual grounded reasoning. Both approaches are based on our two-stage training framework \ours (Fig.~\ref{fig:methodology}), but differ fundamentally in \textit{how} the MLLM is encouraged to attend to chart regions across reasoning steps.

\subsubsection{Explicit Visual Grounding}
\label{appendix:subsubsec:explicit}

In \textit{explicit} visual grounded reasoning, the MLLM is trained end-to-end to produce multi-step reasoning with dynamically changed visual focuses. Concretely, in \textit{Stage I}, given the chart image $\mathcal{I}$ and question $Q$, the model learns to dynamically ground logical reasoning in corresponding visually focused regions by predicting bounding boxes $V_t$ for each reasoning step $R_t$ across $t$ steps of reasoning (Eqs.~\ref{equation:stage1_generate}-\ref{equation:stage1_loss}).
Each predicted bounding box is represented as a normalized coordinate tuple $[x_{\min}, y_{\min}, x_{\max}, y_{\max}]$ over the image canvas, and the grounding loss $\mathcal{L}_V$ directly supervises the model's spatial predictions $V_t$ against ground-truth focus regions $V_t^*$.
In \textit{Stage II}, the model transitions to interleaved visual grounded reasoning (\modeRVA). At each step $t$, the bounding box $V_t$ is applied to the original chart image $\mathcal{I}$ through one of our visual grounding strategies (\S\ref{appendix: 3 visual grounding strategies}) to construct the grounded visual state $\mathcal{I}_t$, which is then provided as an additional visual input at the next reasoning step to implement the $V_{t'} \rightarrow \mathcal{I}_{t'}$ mapping (Eq.~\ref{equation:stage2_interleave}).
Note that during \textit{training}, we use ground-truth $V_t^*$ as intermediate visual augmentation; while in \textit{inference}, the model continues its reasoning based on its last-step $V_t$.

Under this design, the MLLM jointly learns to directly articulate its visual focus as explicit coordinate predictions and dynamically augment its reasoning by linking each $R_t$ with corresponding visually focused regions.
The training objective in Eq.~\ref{equation:stage2_loss} jointly supervises reasoning, grounding, and final answer quality, with $\lambda_R$, $\lambda_V$, and $\lambda_A$ balancing the three components.
The key characteristic of explicit visual grounded reasoning is that the MLLM internalizes both roles on its own, inherently encouraged to dynamically shift attention alongside reasoning.

\subsubsection{Implicit Visual Grounding}
\label{appendix:subsubsec:implicit}

In implicit visual grounded reasoning design, visual grounding is performed by a dedicated \textit{grounder module} that operates alongside the MLLM, implicitly affecting MLLM reasoning. 

\paragraph{Grounder Module.}
The grounder is a lightweight grounding head $G_\phi$ that operates on the model's internal representations to produce patch-level attention maps over the chart image. Specifically, it takes as input (1) the visual patch embeddings $\mathbf{E}^{\text{v}} \in \mathbb{R}^{N_v \times d}$, extracted from the frozen vision encoder with $N_v$ patches and hidden dimension $d$, and (2) the decoder hidden state $\mathbf{h}_t \in \mathbb{R}^{d}$ at a designated $<$\texttt{GROUND}$>$ token position appended after each reasoning step $R_t$. As such, the grounder computes a patch-level attention map:

\begin{equation}
    \mathbf{a}_t = g_\phi\!\left(\mathbf{E}^{\text{v}},\, \mathbf{h}_t\right) \in \Delta^{N_v - 1}
    \label{equation:grounder}
\end{equation}

where $\Delta^{N_v-1}$ denotes the $(N_v{-}1)$-simplex (i.e., the output is a distribution over visual patches). Internally, $g_\phi$ applies self-attention over $\mathbf{E}^{\text{v}}$ to capture contextual relationships among patches, followed by separate projection networks for the visual and decoder representations, and a scaled dot-product operation to compute attention logits. The training objective for $g_\phi$ combines KL divergence (Eq.~\ref{equation:loss_kl}), binary cross-entropy (Eq.~\ref{equation:loss_bce}), Dice loss (Eq.~\ref{equation:loss_dice}), and cross-entropy (Eq.~\ref{equation:loss_ce}), all computed against ground-truth binary focus masks $M_t^* \in \{0,1\}^{N_v}$:

\begin{equation}
    \mathcal{L}_V(V_t, V_t^*) = \lambda_{\mathrm{KL}}\mathcal{L}_{\mathrm{KL}} + \lambda_{\mathrm{BCE}}\mathcal{L}_{\mathrm{BCE}} + \lambda_{\mathrm{Dice}}\mathcal{L}_{\mathrm{Dice}} + \lambda_{\mathrm{CE}}\mathcal{L}_{\mathrm{CE}}
    \label{equation:grounder_loss}
\end{equation}

where each term supervises a distinct aspect of the predicted attention map $\mathbf{a}_t$ against the ground-truth binary focus mask $M_t^*$:

\begin{align}
    \mathcal{L}_{\mathrm{KL}} &= \sum_{i=1}^{N_v} \tilde{M}_{t,i}^* \log \frac{\tilde{M}_{t,i}^*}{\mathbf{a}_{t,i} + \epsilon} \label{equation:loss_kl} \\
    \mathcal{L}_{\mathrm{BCE}} &= -\frac{1}{N_v}\sum_{i=1}^{N_v} \left[ M_{t,i}^* \log(\mathbf{a}_{t,i} + \epsilon) + (1 - M_{t,i}^*)\log(1 - \mathbf{a}_{t,i} + \epsilon) \right] \label{equation:loss_bce} \\
    \mathcal{L}_{\mathrm{Dice}} &= 1 - \frac{2\sum_{i=1}^{N_v} \mathbf{a}_{t,i}\, M_{t,i}^*}{\sum_{i=1}^{N_v} \mathbf{a}_{t,i} + \sum_{i=1}^{N_v} M_{t,i}^* + \epsilon} \label{equation:loss_dice} \\
    \mathcal{L}_{\mathrm{CE}} &= -\log\, \mathbf{a}_{t,\, \arg\max_i M_{t,i}^*} \label{equation:loss_ce}
\end{align}

where $\tilde{M}_t^* = M_t^* / (\sum_i M_{t,i}^* + \epsilon)$ is the row-normalized ground-truth mask treated as a target probability distribution for the KL term, $\epsilon$ is a small constant for numerical stability, and $\mathbf{a}_{t,i}$ denotes the predicted attention weight for the $i$-th visual patch. $\mathcal{L}_{\mathrm{KL}}$ penalizes distributional divergence between the predicted attention map and the normalized ground-truth mask; $\mathcal{L}_{\mathrm{BCE}}$ enforces patch-level binary classification independently across all patches; $\mathcal{L}_{\mathrm{Dice}}$ optimizes region overlap between the predicted and ground-truth focus regions; and $\mathcal{L}_{\mathrm{CE}}$ treats the patch with the highest ground-truth activation as the target class, encouraging the model to concentrate probability mass on the most salient region. The loss weights $\lambda_{\mathrm{KL}}$, $\lambda_{\mathrm{BCE}}$, $\lambda_{\mathrm{Dice}}$, and $\lambda_{\mathrm{CE}}$ are set to $0.25$ uniformly in our experiments.

\paragraph{Stage I: Grounder Training.}
In \textit{Stage I}, the MLLM parameters $\theta$ are kept frozen while the grounder $G_\phi$ is trained to predict accurate visual focuses for each reasoning step. The decoder hidden state $\mathbf{h}_t$ at $<$\texttt{GROUND}$>$ position is extracted and passed to $G_\phi$ together with $\mathbf{E}^{\text{v}}$. The training objective follows \Eref{equation:stage1_loss}, with $\mathcal{L}_V$ computed as in \Eref{equation:grounder_loss}.

\paragraph{Stage II: MLLM Training with Visual Grounded Feedback.}
In \textit{Stage II}, the grounder $G_\phi$ is frozen and the MLLM parameters $\theta$ are finetuned on the full interleaved reasoning objective.
The model is trained to produce reasoning chains of the form:

\begin{equation}
    \underbrace{R_1}_{\text{step 1}}<\texttt{GROUND}>\;\underbrace{R_2}_{\text{step 2}}<\texttt{GROUND}>\;\cdots\;\underbrace{R_t}_{\text{step }t}<\texttt{GROUND}>\;\underbrace{A}_{\text{answer}}<\texttt{GROUND}>
    \label{equation:implicit_format}
\end{equation}

where $<$\texttt{GROUND}$>$ tokens serve as anchors at which the grounder produces attention maps $\{\mathbf{a}_t\}_{t=1}^{T}$, which are used to construct grounded visual states $\{\mathcal{I}_t\}_{t=1}^{T}$ through our visual grounding strategies (\S\ref{appendix: 3 visual grounding strategies}). However, as the MLLM generates reasoning steps $R_t$ without explicit coordinate predictions, the visual grounding signal remains implicit. As such, the model is encouraged to internally shift its internal attention focus at each reasoning step, revealing less effective visual augmentation than \textit{explicit} visual grounded reasoning (\S\ref{subsec:ablation_results}).
This performance gap also highlights differences between visual grounded reasoning tasks that leverage learn-to-focus dynamics to enhance reasoning and visual grounding tasks aimed at accurately localizing visual focus.

% Preliminary Test - Applied

\begin{wrapfigure}[14]{r}{0.6\textwidth}
    \small
    \centering
    \includegraphics[width=0.6\textwidth]{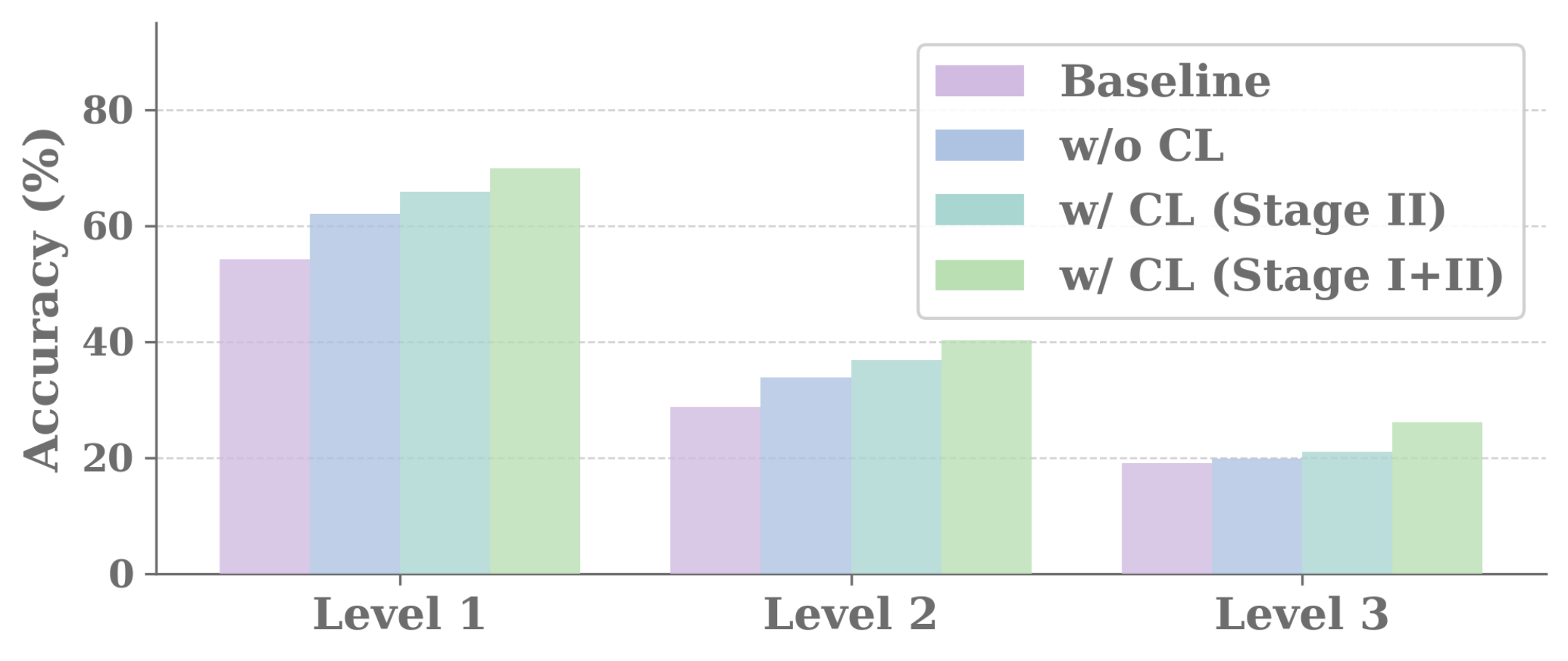}
    \vspace{-21pt}
    \caption{\textbf{Effects of Curriculum Learning.}
    Using Qwen2.5-VL-7B as the base model, we compare its performance across different training settings.}
    \label{fig:ablation_curriculum_learning}
\end{wrapfigure}

\setlength{\intextsep}{0pt}

\subsection{Curriculum Learning: Effectiveness of CL in Chart Understanding}
\label{appendix:subsec:ablation_curriculum_learning}

To validate the effectiveness of curriculum learning, we further compare \ours with the baseline model (\textit{Qwen2.5-VL-7B}) in both untrained and non-curriculum training settings. As shown in \Fref{fig:ablation_curriculum_learning}, although training without curriculum learning improves performance on \oursdata, it consistently underperforms as compared to \ours trained with either \textit{Stage I} or \textit{Stage I+II}. This gap becomes more pronounced as task complexity increases (i.e., levels $1 \rightarrow 3$), with diminishing performance gains observed in the non-curriculum setting.
These results highlight the effectiveness of curriculum learning in guiding the model to learn from fundamental chart elements and gradually develop intrinsic visual reasoning capabilities. This structured learning process not only improves the model’s adaptability to more complex chart understanding tasks but also enhances its generalization to out-of-domain multimodal reasoning.

% Preliminary Test - Applied

\begin{wrapfigure}[15]{r}{0.6\textwidth}
    \small
    \centering
    \includegraphics[width=0.6\textwidth]{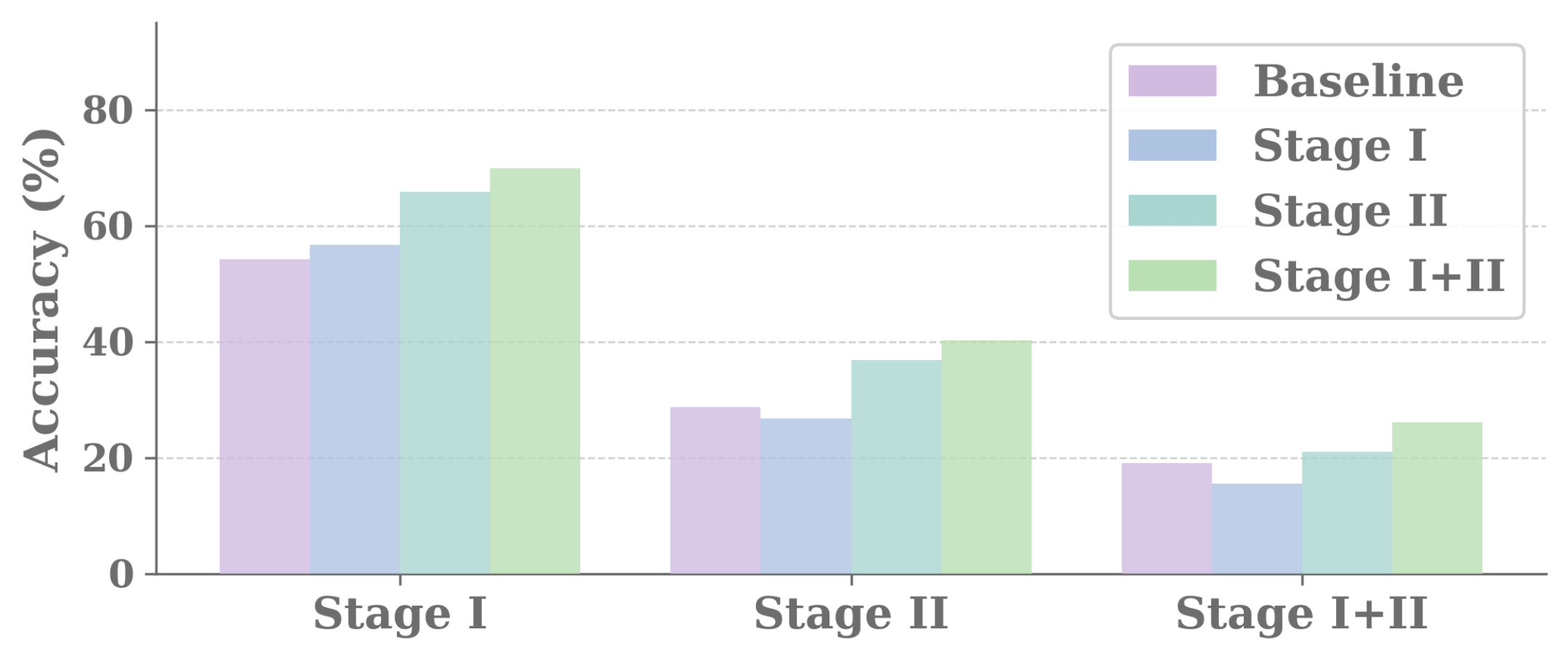}
    % \vspace{-16pt}
    \caption{\textbf{Strength of Curriculum Training.}
    Using Qwen2.5-VL-7B as the base model, we compare its performance across different training settings.}
    \label{fig:ablation_curv_stages}
\end{wrapfigure}

\setlength{\intextsep}{0pt}

\subsection{Two-Stage Learning: Strength of Training Curriculum}
\label{appendix:subsec:ablation_on_curv_stages}

In addition to data curriculum (\S\ref{appendix:subsec:ablation_curriculum_learning}), \ours also implements the two-stage curriculum training paradigm from internalizing to enhancing MLLM visual reasoning capabilities (\S\ref{sec:methodology}).
In \textit{Stage I}, MLLM learns vision-reasoning alignment through decoupled visual grounding, establishing explicit associations between visual evidence and reasoning logics. In \textit{Stage II}, MLLM is further trained to perform interleaved visual reasoning, leveraging the grounding capabilities acquired in \textit{Stage I} to support step-by-step reasoning over visual inputs.
To validate the effectiveness of this design, we compare the performance among Qwen2.5-VL-7B, \ours{}@\textit{Stage I}, \ours{}@\textit{Stage II}, and \ours{}@\textit{Stage I+II} (Fig.~\ref{fig:ablation_curv_stages}). Results show that \ours{}@\textit{Stage I+II} consistently outperforms both single-stage variants, indicating the advantage of combining decoupled alignment with interleaved reasoning via two-stage curriculum training.

In contrast, \ours{}@\textit{Stage I} alone improves performance on \textit{level} 1 simple CQA while degrades on harder single-chart and multi-chart understanding tasks. This suggests that decoupled grounding, while effective for learning basic visual-reasoning associations, is insufficient for supporting compositional reasoning that requires integrating multiple pieces of visual evidence. Conversely, \ours{}@\textit{Stage II} alone fails to fully realize these benefits, implying that interleaved reasoning without well-established grounding leads to weaker visual-reasoning alignment.

Collectively, these findings highlight that the strength of \ours lies in its staged curriculum training design: \textit{Stage I} builds foundational visual grounding ability, while \textit{Stage II} enables the model to operationalize this grounding ability into coherent, step-by-step visual reasoning. Their combination results in more robust and generalizable visual reasoning capacity across varying task complexities.

\subsection{Foundational Learning: Trade-off between Robustness \& Adaptability}
\label{subsec:importance_of_foundational_learning}

\Fref{fig:complexity_effects} reveals a critical finding that strongly validates our curriculum learning design (\S\ref{sec:dataset_construction}).
While training on \textit{level} 1 alone provides solid foundational performance ($\uparrow$35.08\% on \textit{level} 1), progressive training on \textit{levels} 1+2 demonstrates the optimal learning accumulation, achieving the best overall performance across all difficulty \textit{levels} ($\uparrow{}15.65\%$ on \textit{level} 1, $\uparrow{}11.53\%$ on \textit{level} 2, $\uparrow{}7.10\%$ on \textit{level} 3).
However, extending training to include \textit{level} 3 unfolds a concerning trade-off between robustness and adaptability:
While notably improves complex reasoning performance ($\uparrow{}12.82\%$ on \textit{level} 3), it significantly degrades foundational reasoning abilities, only higher than baseline by $\uparrow{}5.93\%$ and $\uparrow{}0.03\%$ on \textit{levels} 1 and 2, respectively.

\begin{table}[t]
\centering
\small
\renewcommand{\arraystretch}{1.15}

\begin{tabularx}{\textwidth}{l *{6}{>{\centering\arraybackslash}X}}
\toprule
\multirow{2}{*}{\textbf{Method}}
& \multicolumn{2}{c}{\textbf{Level 1}}
& \multicolumn{2}{c}{\textbf{Level 2}}
& \multicolumn{2}{c}{\textbf{Level 3}} \\
\cmidrule(lr){2-3} \cmidrule(lr){4-5} \cmidrule(lr){6-7}
& \textit{acc@M} & \textit{acc@0.0}
& \textit{acc@M} & \textit{acc@0.0}
& \textit{acc@M} & \textit{acc@0.0} \\
\midrule
\rowcolor{gray!15} \multicolumn{7}{l}{\textit{Base Models}} \\
Qwen2.5-VL-7B          & 54.21 & 50.79 & 28.68 & 28.82 & 19.01 & 19.38 \\
Qwen3-VL-4B            & 48.07 & 47.28 & 26.86 & 26.61 & 20.68 & 20.04 \\
\midrule
\rowcolor{gray!15} \multicolumn{7}{l}{\textit{Chart Specialist}} \\
ChartGemma~\citep{chartgemma2024}        & 28.71 & 21.36 & 12.46 & \phantom{0}8.25 & \phantom{0}7.71 & \phantom{0}8.86 \\
\midrule
\rowcolor{gray!15} \multicolumn{7}{l}{\textit{Tool-Use Agents}} \\
Thyme~\citep{thyme2025}    & 56.57 & 55.57 & 24.39 & 24.96 & 18.07 & 18.54 \\
DeepEyes~\citep{deepeyes_2025} & 58.43 & 57.21 & 30.04 & 29.29 & 20.75 & 20.29 \\
\midrule
\rowcolor{gray!15} \multicolumn{7}{l}{\textit{Ours}} \\
\ours{} (Qwen3-VL-4B)   & \textbf{64.21} & \textbf{63.79} & \textbf{34.86} & \textbf{34.57} & \textbf{23.82} & \textbf{23.25} \\
\ours{} (Qwen2.5-VL-7B) & \textbf{69.86} & \textbf{66.79} & \textbf{40.21} & \textbf{38.64} & \textbf{26.11} & \textbf{24.25} \\
\bottomrule
\end{tabularx}

\caption{\textbf{Comparison Against Chart Specialist \& Tool-Use Agents on
CCQA.} We evaluate \ours against a newer base model (Qwen3-VL-4B), a
chart-specialist baseline (ChartGemma), and tool-augmented agents (Thyme,
DeepEyes) across three curriculum levels.}

\label{tab:broader_baselines}
\end{table}

\subsection{Broader Baselines: Comparison Against Chart Specialist \& Tool-Use Agents}
\label{appendix:subsec:broader_baselines_compare_against_chart_specialist_and_tool_agent}

We broaden our comparison along three axes: a newer base model \texttt{Qwen3-VL-4B}~\citep{qwen3_report}, a chart-specialist baseline \texttt{ChartGemma}~\citep{chartgemma2024}, and tool-augmented agents that invoke external visual tools \texttt{Thyme}~\citep{thyme2025} and \texttt{DeepEyes}~\citep{deepeyes_2025}. All methods are evaluated on our CCQA test set across the three curriculum levels using both \textit{acc@M} and the absolute accuracy \textit{acc@0.0} (\S\ref{sec:experiments}).
As shown in \Tref{tab:broader_baselines}, \ours is generalizable to newer backbones, improving over its base model (Qwen3-VL-4B) by up to $\uparrow$16.14\%. This validates that the improvements from \ours are not tied to a specific base model, but instead reflect a transferable strengthening of intrinsic visual-grounded reasoning capabilities.
Moreover, \ours also outperforms chart specialist and tool-augmented agents built on the same Qwen2.5-VL-7B backbone, providing further evidence to the effectiveness and robustness of \ours in the face of varying chart complexity levels.

\section{Multi-Step Reasoning With Dynamic Visual Grounding}
\label{appendix:sec:case_studies}

\begin{table}[htbp]

\centering
\resizebox{\textwidth}{!}{%
\begin{tabular}{cc|cccc|cccc|cccc}
\toprule
\multirow{5}{*}{\textbf{Model}} & \multirow{5}{*}{\textbf{Size}} & \multicolumn{12}{c}{\textbf{\oursdata}} \\
\cmidrule{3-14}
& & \multicolumn{4}{c|}{\textbf{Level 1}} & \multicolumn{4}{c|}{\textbf{Level 2}} & \multicolumn{4}{c}{\textbf{Level 3}} \\
\cmidrule{3-6}\cmidrule{7-10}\cmidrule{11-14}
& & \multicolumn{2}{c}{\cellcolor{lightblue}\textbf{Reasoning}} & \cellcolor{lightorange}\textbf{Grounding} & \cellcolor{lightpurple}\textbf{Answer} & \multicolumn{2}{c}{\cellcolor{lightblue}\textbf{Reasoning}} & \cellcolor{lightorange}\textbf{Grounding} & \cellcolor{lightpurple}\textbf{Answer} & \multicolumn{2}{c}{\cellcolor{lightblue}\textbf{Reasoning}} & \cellcolor{lightorange}\textbf{Grounding} & \cellcolor{lightpurple}\textbf{Answer} \\
\cmidrule{3-4}\cmidrule{7-8}\cmidrule{11-12}
& & \cellcolor{lightblue}$\boldsymbol{acc\textsc{@}mac}$ & \cellcolor{lightblue}$\boldsymbol{acc\textsc{@}mic}$ & \cellcolor{lightorange}\textbf{mIoU} & \cellcolor{lightpurple}$\boldsymbol{acc\textsc{@}mac}$ & \cellcolor{lightblue}$\boldsymbol{acc\textsc{@}mac}$ & \cellcolor{lightblue}$\boldsymbol{acc\textsc{@}mic}$ & \cellcolor{lightorange}\textbf{mIoU} & \cellcolor{lightpurple}$\boldsymbol{acc\textsc{@}mac}$ & \cellcolor{lightblue}$\boldsymbol{acc\textsc{@}mac}$ & \cellcolor{lightblue}$\boldsymbol{acc\textsc{@}mic}$ & \cellcolor{lightorange}\textbf{mIoU} & \cellcolor{lightpurple}$\boldsymbol{acc\textsc{@}mac}$ \\
\midrule
\multicolumn{14}{c}{\cellcolor{gray!25}\textbf{Baselines}} \\

\midrule

GPT-4o & - & \underline{53.17} & \underline{51.02} & 22.03 & \underline{50.00} & \underline{53.32} & \underline{53.60} & 13.01 & \underline{24.00} & \underline{50.20} & \underline{46.86} & 11.14 & \underline{21.50} \\
% \addlinespace
% Gemma-3 & 4B & ? & ? & ? & ? & ? & ? & ? & ? & ? & ? & ? & ? \\
% \addlinespace
% Llama-3.2-V & 11B & ? & ? & ? & ? & ? & ? & ? & ? & ? & ? & ? & ? \\
\addlinespace
\multirow{2}{*}{Qwen2.5-VL} & 3B & 44.90 & 39.07 & 37.12 & 36.00 & 38.72 & 40.08 & 29.73 & 14.00 & 32.17 & 35.47 & 27.43 & 12.00 \\
& 7B & 48.53 & 41.70 & \underline{48.17} & 45.00 & 40.49 & 40.68 & \underline{32.25} & 21.50 & 35.82 & 38.10 & \underline{32.26} & 17.50 \\
\midrule
\multicolumn{14}{c}{\cellcolor{gray!25}\textbf{Ours}} \\
\midrule

\multirow{2}{*}{\shortstack{\textbf{Applied}\\(Qwen2.5-VL)}} & 3B & \cellcolor{gain!21.80}50.50 & \cellcolor{gain!48.68}53.63 & \cellcolor{gain!35.15}47.17 & \cellcolor{gain!47.00}50.00 & \cellcolor{gain!26.21}45.79 & \cellcolor{gain!21.74}45.66 & \cellcolor{gain!37.58}40.59 & \cellcolor{gain!20.00}19.00 & \cellcolor{gain!34.34}41.95 & \cellcolor{gain!14.87}38.76 & \cellcolor{gain!26.51}34.60 & \cellcolor{gain!14.00}15.00 \\

& 7B & \cellcolor{gain!27.80}\textbf{56.13} & \cellcolor{gain!41.93}\textbf{54.01} & \cellcolor{gain!33.14}\textbf{57.55} & \cellcolor{gain!26.00}\textbf{52.00} & \cellcolor{gain!33.14}\textbf{49.87} & \cellcolor{gain!24.47}\textbf{47.17} & \cellcolor{gain!45.41}45.72 & \cellcolor{gain!9.50}23.00 & \cellcolor{gain!35.69}\textbf{46.05} & \cellcolor{gain!7.19}\textbf{38.83} & \cellcolor{gain!27.92}\textbf{39.90} & \cellcolor{gain!11.00}19.50 \\

\addlinespace

\multirow{2}{*}{\shortstack{\textbf{Boxed}\\(Qwen2.5-VL)}} & 3B & \cellcolor{gain!21.99}50.23 & \cellcolor{gain!37.49}49.90 & \cellcolor{gain!30.29}45.55 & \cellcolor{gain!35.00}46.00 & \cellcolor{gain!10.94}40.70 & \cellcolor{gain!12.98}42.74 & \cellcolor{gain!30.14}38.11 & \cellcolor{gain!6.50}14.50 & \cellcolor{gain!19.43}36.98 & \cellcolor{gain!9.50}36.97 & \cellcolor{gain!22.82}33.37 & \cellcolor{gain!9.50}13.50 \\

& 7B & \cellcolor{gain!13.22}51.27 & \cellcolor{gain!24.50}48.20 & \cellcolor{gain!13.13}50.88 & \cellcolor{gain!17.00}49.00 & \cellcolor{gain!12.44}42.97 & \cellcolor{gain!13.28}43.44 & \cellcolor{gain!37.52}43.09 & \cellcolor{gain!8.00}22.50 & \cellcolor{gain!21.53}41.33 & 37.45 & \cellcolor{gain!11.78}34.52 & \cellcolor{gain!6.50}18.00 \\

\addlinespace

\multirow{2}{*}{\shortstack{\textbf{Cropped}\\(Qwen2.5-VL)}} & 3B & 42.17 & \cellcolor{gain!27.83}46.68 & \cellcolor{gain!47.30}51.22 & \cellcolor{gain!14.00}39.00 & \cellcolor{gain!10.40}40.52 & \cellcolor{gain!12.83}42.69 & \cellcolor{gain!53.27}45.82 & \cellcolor{gain!6.50}14.50 & \cellcolor{gain!27.38}39.63 & \cellcolor{gain!6.17}35.86 & \cellcolor{gain!25.52}34.27 & \cellcolor{gain!8.00}13.00 \\

& 7B & \cellcolor{gain!10.40}50.33 & \cellcolor{gain!32.27}50.79 & \cellcolor{gain!14.48}51.33 & \cellcolor{gain!11.00}47.00 & \cellcolor{gain!18.44}44.97 & \cellcolor{gain!14.21}43.75 & \cellcolor{gain!58.55}\textbf{50.10} & \cellcolor{gain!12.50}\textbf{24.00} & \cellcolor{gain!24.83}42.43 & \cellcolor{gain!6.23}38.51 & \cellcolor{gain!25.79}39.19 & \cellcolor{gain!9.50}19.00 \\

\bottomrule
\end{tabular}%
}

\caption{\textbf{Performance Evaluation for \modeRVA Mode Inference.} Employing the set of evaluation metrics (\Sref{subsec:eval_evaluation_metrics}), we assess model reasonnig, visual grounding, and final answer, respectively.}
\label{tab:main_results_RVA}

\end{table}

\subsection{Challenges In Multi-Step Visual Grounding}
\label{appendix:inference_in_RAV_mode}

We leverage \oursdata, randomly selecting 500 samples to evaluate model inference through \modeRVA mode (\S\ref{subsec:exp_setup}).
Training MLLMs with explicit reasoning and visual grounding as intermediate outputs effectively enhances model's intrinsic visual reasoning capabilities (\S\ref{subsec:main_results}).
This step-by-step visual reasoning guides the model to decompose complex tasks into structured reasoning chains through dynamic attention grounding.
With intermediate grounding naturally supporting more coherent reasoning trajectories, this in turn enhances model's ability to establish interleaved thinking-perception correspondences.
Aligning with human visual reasoning, \textit{decomposed} reasoning chains effectively help models to develop and strengthen their intrinsic visual reasoning capabilities.

Different from learning, during inference, human visual reasoning is rather a \textit{composed} process that interleaves logical reasoning with visual comprehension, while \textit{compositing} all intermediate steps into a coherent chain of thought.
In contrast, inference in \modeRVA exposes the fragility of step-wise generation: once an intermediate step is flawed, whether by incorrect calculation or inaccurate visual comprehension, the error propagates through the chain, breaking the balance between perception and reasoning that eventually leads to incorrect final answers (\S\ref{appendix:inference_in_RAV_mode}).
Therefore, \textit{it can be an effective way of learning, while may not be as useful in inference}.

\Tref{tab:main_results_RVA} summarizes the evaluation results of \modeRVA inference. Fintuned models achieve noticeable improvements across reasoning (up to 10.23\% absolute gain), grounding (up to 9.38\% absolute gain), and answering (up to 14\% absolute gain).
Beyond these numerical results, qualitative inspection (\S\ref{appendix:subsec:success_cases}) reveals distinct behavioral patterns where training with step-by-step visual grounding encourages systematic reasoning chains with sharper object localization, showcasing stronger alignment with human-like reasoning trajectories.

Despite these improvements, however, the answering performance remains lower than that of \ours when using the same base model and grounding method (\Tref{tab:main_result_on_ours_data}).
This indicates that, while \modeRVA training can effectively enhance intrinsic visual reasoning capabilities, \modeRVA inference magnifies the vulnerability to intermediate error accumulation.

% However, inference in this way can amplify the accumulation of intermediate errors that propagate through the reasoning chain, leading to incorrect final answers (\S\ref{appendix:inference_in_RAV_mode}).
% \Tref{tab:main_results_RVA} summarizes evaluation results of \modeRVA inference.
% Despite the notable improvements across reasoning (up to 10.23\% absolute gain), grounding (up to 9.38\% absolute gain), and answering (up to 14\% absolute gain), the answering performance fall shorts of \ours using the same base model and grounding method.

\begin{figure}[H]
    \vspace{20pt}
    
    \centering
    \includegraphics[width=1.0\textwidth]{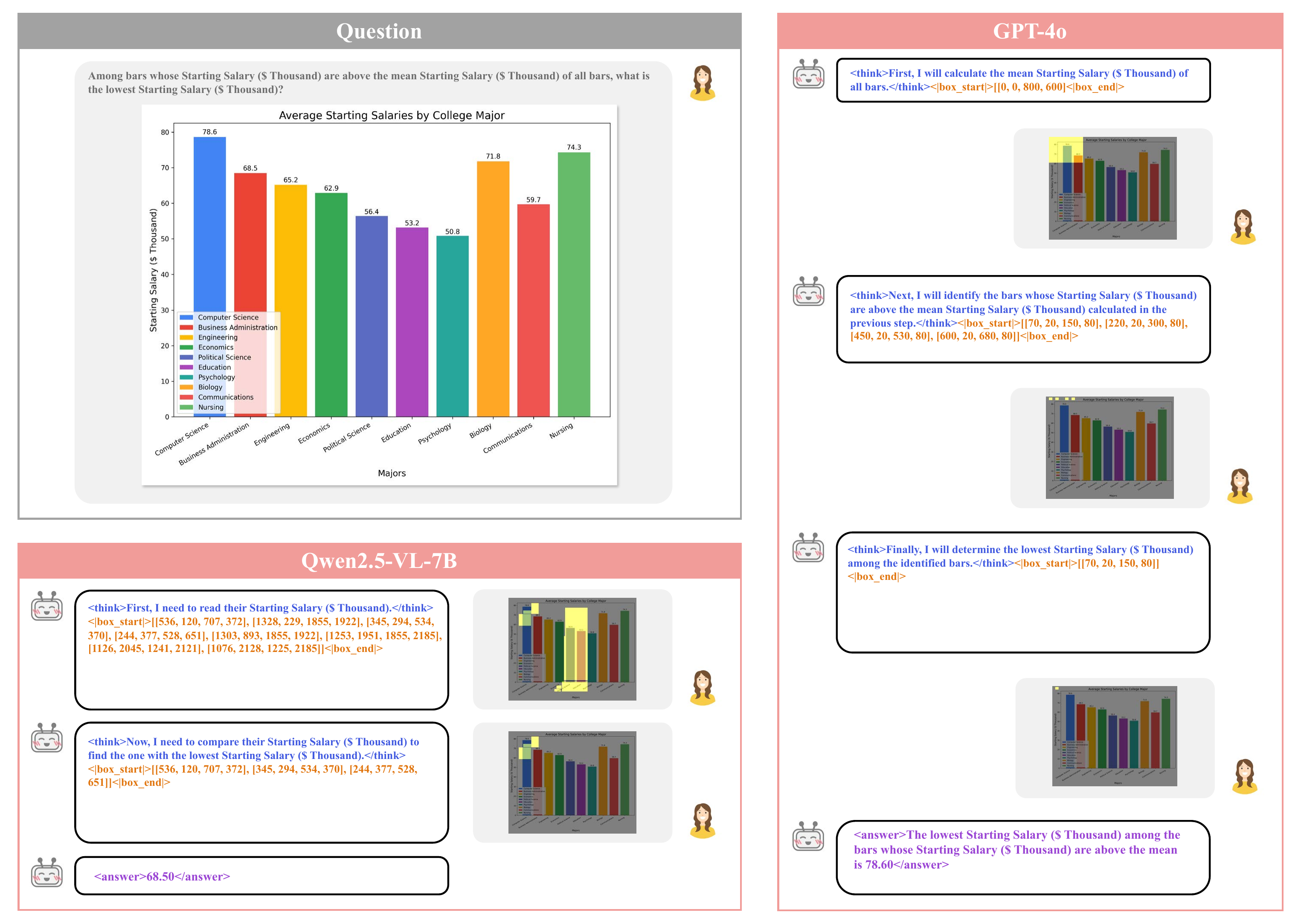}
    \vspace{-20pt}
    \caption{\textbf{Challenge of Inference in Mode \modeRVA.} Tested on GPT-4o and Qwen2.5-VL, this example illustrates the challenge MLLMs face in performing inference in \textbf{\modeRVA}.}
    \label{fig:inference_mode_RVA}
\end{figure}

Build upon our discussions above, we present the challenge for MLLMs in CQA through mode \modeRVA.
Although \modeRVA training notably enhances model's intrinsic visual reasoning capabilities, \modeRVA inference reveals critical limitations.
During \modeRVA inference, cumulative intermediate errors can cascade through the reasoning chain, ultimately compromising the model's ability to produce correct answers.
For example, \Fref{fig:inference_mode_RVA} shows a CQA example that is responded to by GPT-4o and Qwen2.5-VL-7B, respectively.
Both GPT-4o and Qwen2.5-VL-7B fail to generate the correct answer due to their disrupted reasoning chains that evolve through their incorrect visual grounding.

\subsection{Inference Failure}
\label{appendix:subsec:failure_cases}

\begin{figure}[H]
    \centering
    \includegraphics[width=1.0\textwidth]{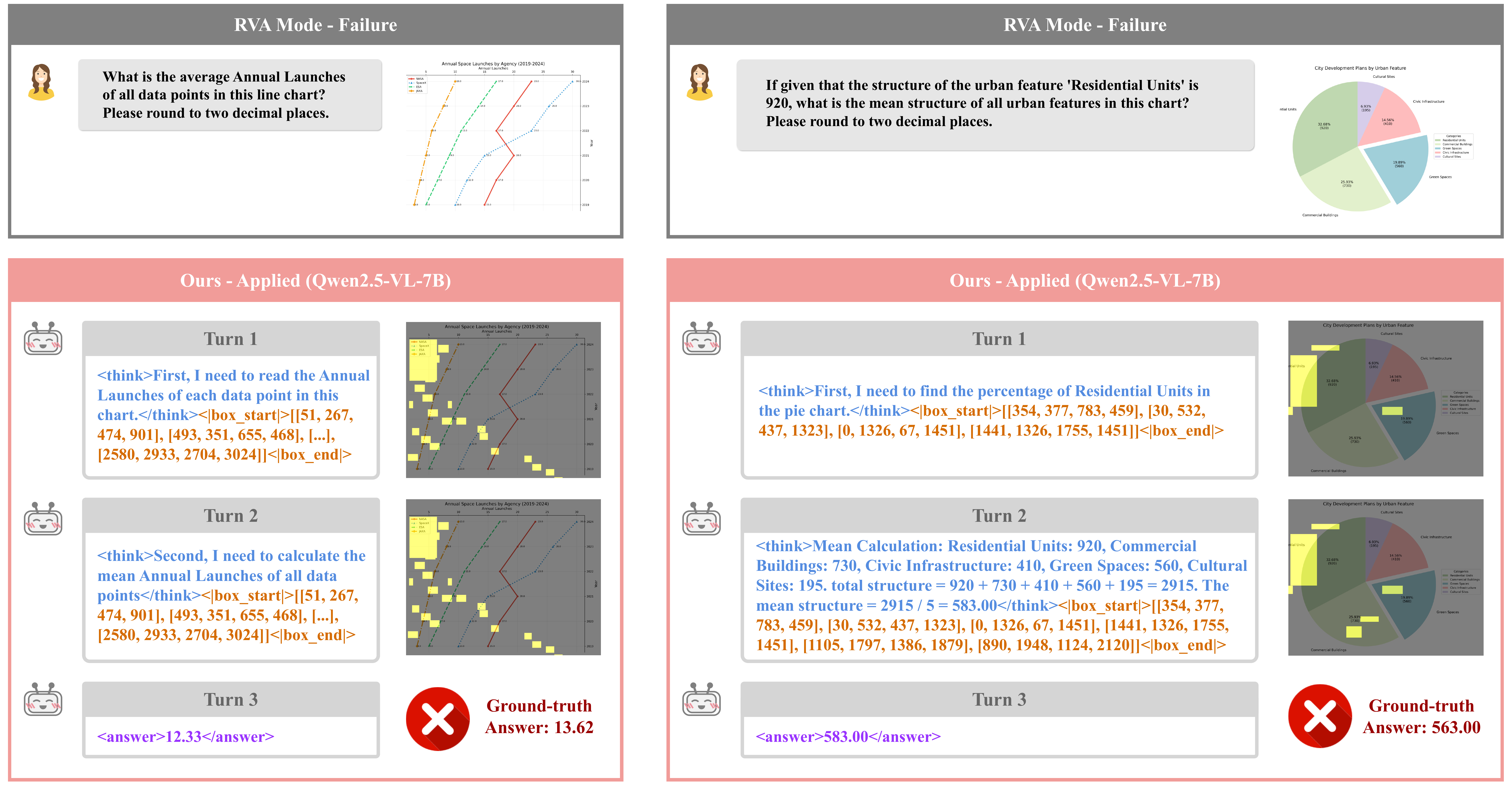}
    \caption{\textbf{Failure Examples Through Mode \modeRVA.} This figure shows two examples of \modeRVA inference on \oursdata, where the model fails to give correct answers.}
    \label{fig:case_study_modeRVA_failure}
\end{figure}

\Fref{fig:case_study_modeRVA_failure} illustrates examples of model inference failures in \modeRVA.
In both cases, the model fails to properly ground its reasoning in the chart, leading to inaccurate extraction and misinterpretation of visual information.
Arising in early reasoning steps, these visual comprehension inaccuracies can propagate through the reasoning chain, ultimately resulting in incorrect question answering.

\subsection{Inference Success}
\label{appendix:subsec:success_cases}

\paragraph{Example 1 - Mode \modeA:}

\begin{figure}[H]
    \centering

    \vspace{48pt}
    \includegraphics[width=1.0\textwidth]{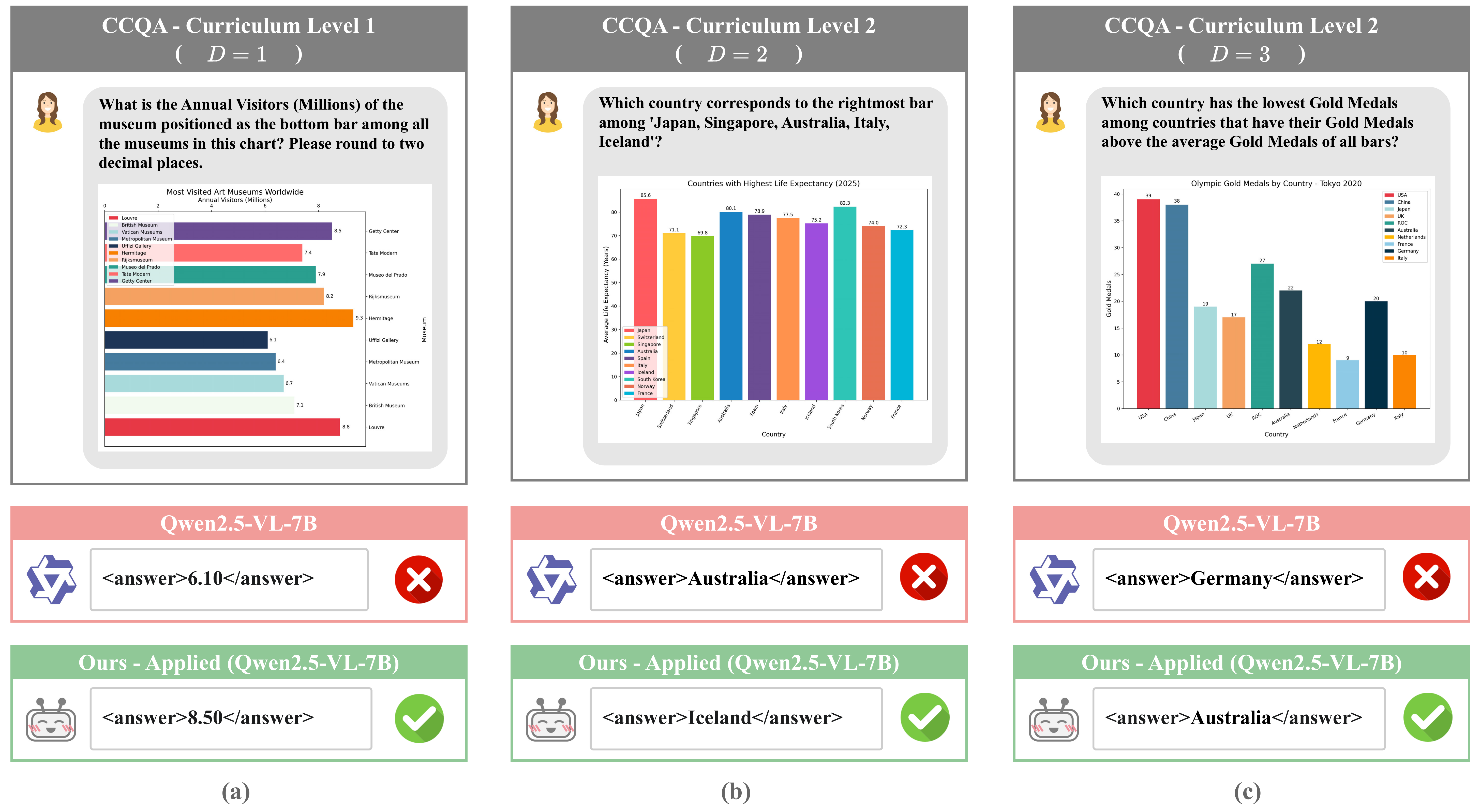}
    \caption{\textbf{Success Examples On Curriculum Levels 1-2.} This figure shows three examples on curriculum levels 1-2 of \oursdata, where the baseline Qwen2.5-VL-7B fails while our \ours using \textit{applied} grounding succeeds.}
    \label{fig:case_study_modeA_c12}

    \vspace{36pt}
    
\end{figure}

\begin{figure}[H]
    \centering
    
    \vspace{10pt}
    \includegraphics[width=1.0\textwidth]{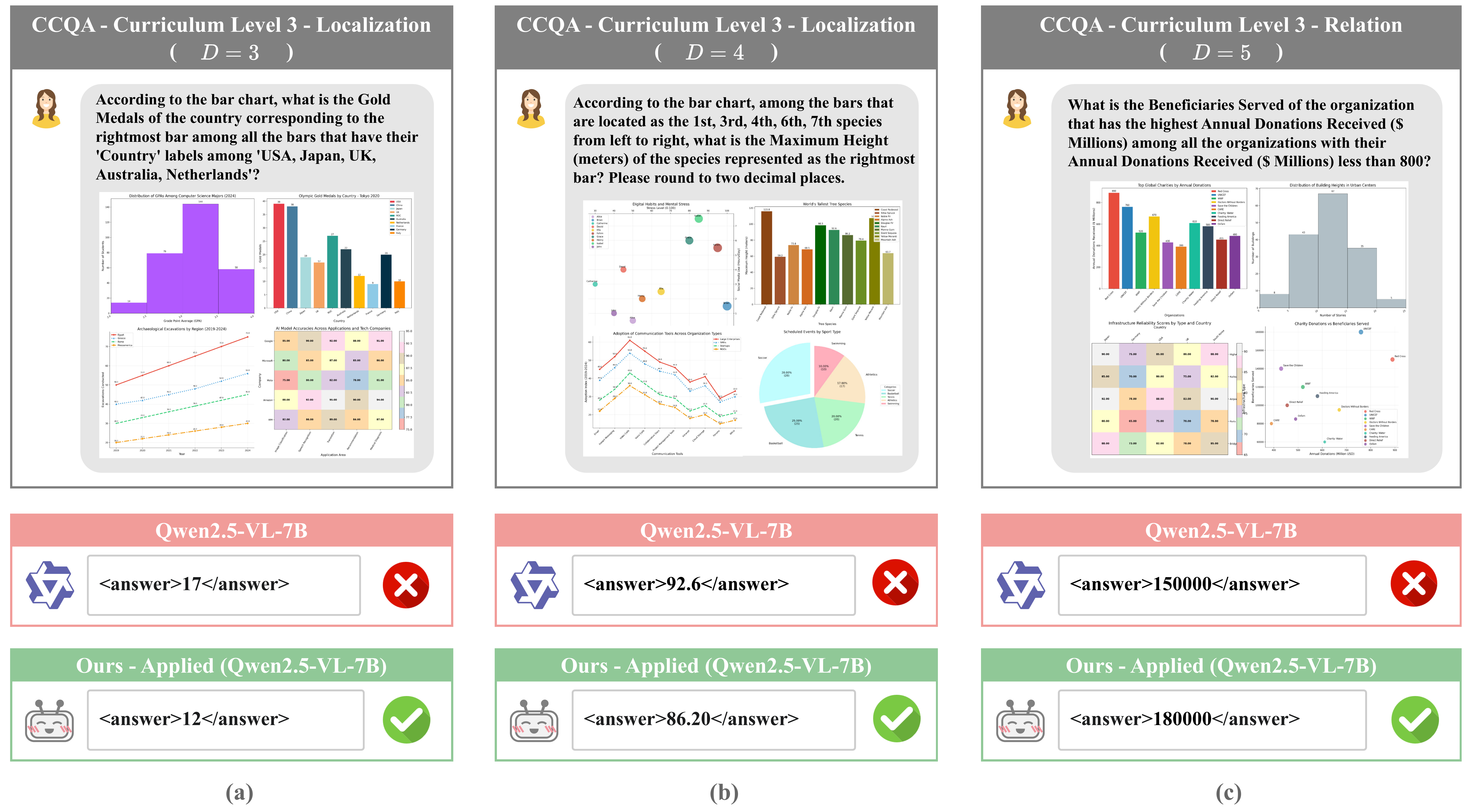}
    \caption{\textbf{Success Examples On Curriculum Level 3.} This figure shows three examples on the curriculum level 3 of \oursdata, where the baseline Qwen2.5-VL-7B fails while our \ours using \textit{applied} grounding succeeds.}
    \label{fig:case_study_modeA_c3}
\end{figure}

Figures~\ref{fig:case_study_modeA_c12} \&~\ref{fig:case_study_modeA_c3} exhibit examples on chart question answering in mode \modeA, where the baseline Qwen2.5-VL-7B fails to generate the correct answer, while \ours (Qwen2.5-VL-7B) finetuned through \textit{applied} grounding succeeds.
\Fref{fig:case_study_modeA_c12} (a) is a simple value reading problem ($D=1$), where the baseline model fails to localize the exact queried chart component.
\Fref{fig:case_study_modeA_c12} (b) consists of two nested functions ($D=2$), where the baseline model fails to localize the queried bar in the given subset of countries.
\Fref{fig:case_study_modeA_c12} (c) further enhance the CQA complexity ($D=3$), involving three nested functions across reasoning, visual grounding, and interleaved calculation that the baseline model fails to correctly response.
Different from 
\Fref{fig:case_study_modeA_c12} that query about a single chart, each CQA sample in \Fref{fig:case_study_modeA_c3} involves multiple charts that significantly complicates question answering.
The baseline model fails in \Fref{fig:case_study_modeA_c12} (a) ($D=3$) as it requires the localization of the exact chart subplot, the required subset, as well as the Y-axis value reading.
\Fref{fig:case_study_modeA_c12} (b) increases the CQA difficulty ($D=4$) by including not only accurate localization of chart components, but also extremia comparison of both bar values and spatial positions.
\Fref{fig:case_study_modeA_c12} (c) ($D=5$) presents further enhanced complexity by involving relations across different charts. This relational chart understanding making the problem solving more challenging, unveiling the significance of accurate visual reasoning in tackling complex CQA tasks.

\paragraph{Example 2 - Mode \modeRVA:}

\begin{figure}[H]
    \centering
    \includegraphics[width=1.0\textwidth]{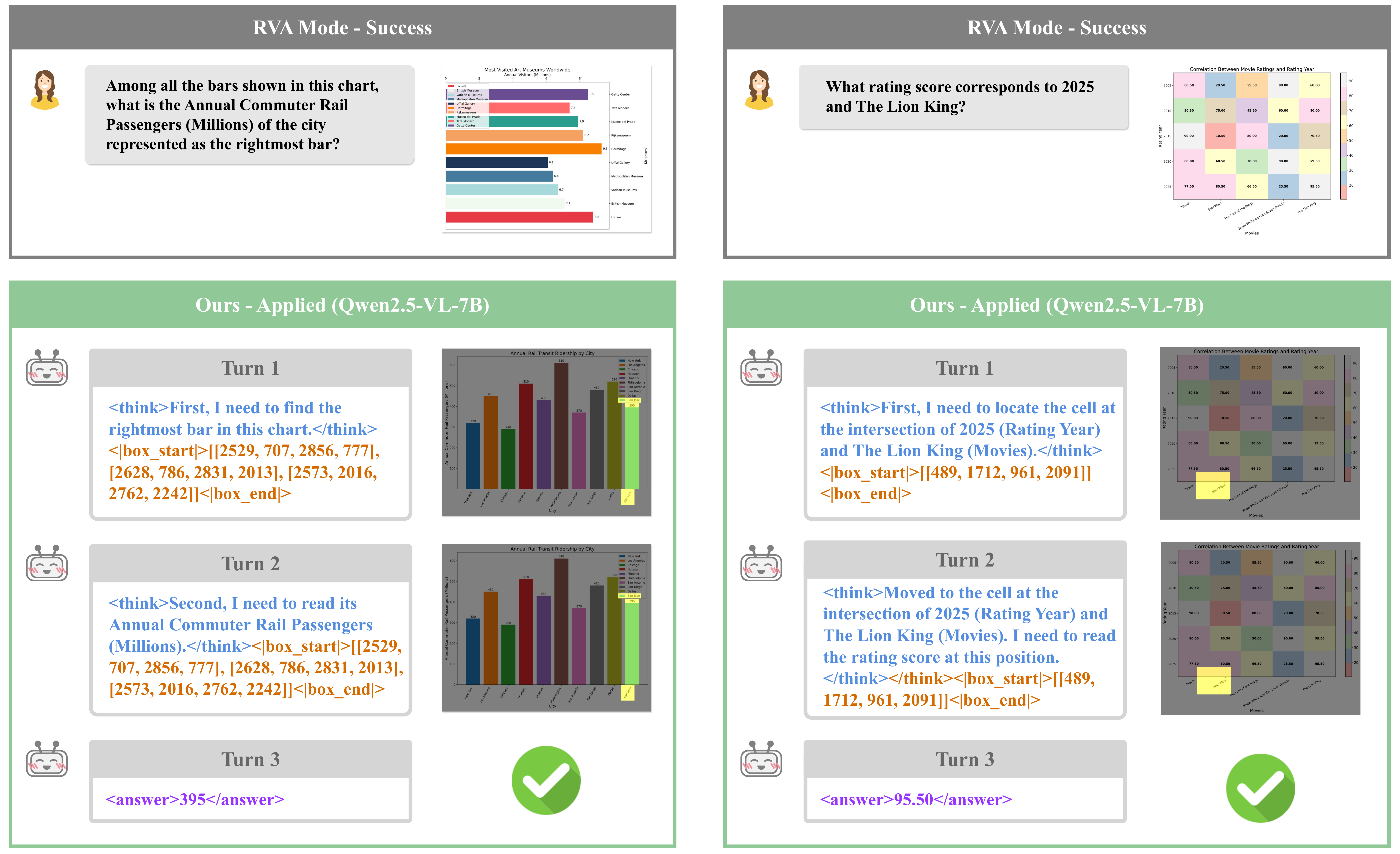}
    \caption{\textbf{Success Examples Through Mode \modeRVA.} This figure shows three examples of \modeRVA inference on \oursdata, where the baseline Qwen2.5-VL-7B fails while our \ours using \textit{applied} grounding succeeds.}
    \label{fig:case_study_modeRVA_success}
\end{figure}

\Fref{fig:case_study_modeRVA_success} presents two examples of successful inference in \modeRVA mode.
The bar chart example on the left shows reasoning with accurate visual grounding.
The heatmap example on the right shows a case where the grounding is not exact but falls close to the regions of focus, also leading to the correct answer.

\section{Limitations \& Future Work}
\label{appendix:sec:limitations}

In this work, we propose \ours (\S\ref{sec:methodology}), a curriculum learning framework that develops intrinsic visual grounded reasoning capabilities in MLLMs by reformulating chart question answering as multi-step visual grounded reasoning with dynamic spatial attention. To support model learning, we introduce \oursdata (\S\ref{sec:dataset_construction}), a three-level curriculum dataset with scalable synthetic generation across diverse chart types and reasoning patterns. Results demonstrate that \ours demonstrates notable and consistent improvements across chart understanding and out-of-domain multimodal reasoning benchmarks (\S\ref{subsec:main_results}). 
Nevertheless, we acknowledge a few limitations that we aim to investigate further in our future work. First, \oursdata encompasses seven common chart types, while less conventional visualization forms, such as Sankey diagrams, treemaps, and geographic maps, are not currently covered, potentially limiting the applicability of \oursdata to future work with more specialized training scenarios. Second, our framework and dataset focus exclusively on English-language charts and questions, which may restrict generalizability to multilingual contexts where chart-based communication is equally prevalent.
In future work, we aim to address these limitations by extending \oursdata to a broader range of chart types and multilingual settings. Beyond these, we also aim to explore agentic chart understanding, where models can better leverage external knowledge, tools, and multi-agent collaboration to complement the intrinsic visual reasoning capabilities established in this work.

\newpage

% Mode A

% Curriculum Level - 1 & 2

% Curriculum Level - 3

\vspace{30pt}
% Mode RVA - Failure

% Curriculum Level - 1 & 2

\vspace{10pt}

\vspace{30pt}
% Mode RVA - Failure

% Curriculum Level - 1 & 2

\vspace{10pt}

\end{document}